\documentclass{article}

\PassOptionsToPackage{numbers}{natbib}
 \usepackage[preprint]{neurips_2026}
\usepackage[utf8]{inputenc} % allow utf-8 input
\usepackage[T1]{fontenc}    % use 8-bit T1 fonts
\usepackage{hyperref}       % hyperlinks
\usepackage{url}            % simple URL typesetting
\usepackage{booktabs}       % professional-quality tables
\usepackage{amsfonts}       % blackboard math symbols
\usepackage{nicefrac}       % compact symbols for 1/2, etc.
\usepackage{microtype}      % microtypography
\usepackage{amsmath, amssymb}
\usepackage{graphicx}
\usepackage[table]{xcolor}
\usepackage{algorithm, algorithmic}
\usepackage{pifont}
\usepackage{titletoc}

\newcommand{\unc}[1]{{\scriptsize(#1)}}

\title{Autoregressive One-Step Generative Modeling for Dynamical System Forecasting}

\author{%
  Tianyue Yang\\
  The Center for Computational Science\\
  University College London\\
  \And
   Xiao Xue\thanks{Corresponding Author.} \\
  The Center for Computational Science\\
  University College London\\
  \texttt{x.xue@ucl.ac.uk} \\ 
}

\begin{document}

\maketitle

\begin{abstract}
Fast surrogate modeling for high-dimensional physical dynamics requires more than low short-term error: useful models must roll out efficiently while preserving the statistical structure of long trajectories. Neural operators provide inexpensive autoregressive forecasts but can drift in turbulent regimes, whereas rolling diffusion and latent generative surrogates can represent stochastic transitions at the cost of multi-step denoising, noise-schedule design, or auxiliary compression models. We propose \textbf{Me}anFlow \textbf{L}ong-term \textbf{I}nvariant \textbf{S}patiotemporal Consistency \textbf{A}utoregressive Models (\textbf{MeLISA}), a latent-free autoregressive generative surrogate built on pixel-space MeanFlow. MeLISA defines a blockwise stochastic transition kernel that generates each forecast block with a single model evaluation, avoiding latent encoders and iterative diffusion solvers at inference time. To stabilize long-horizon rollouts, MeLISA combines a \emph{Window-Consistency MeanFlow} objective that learns conditional spatiotemporal generation from partially observed temporal windows with a \emph{Time Increment Consistency} loss that constrains multi-lag finite increments and targets temporal-correlation structure. We evaluate MeLISA with compact UNet and scalable DiT backbones on two high-resolution benchmarks, extended 2D Kolmogorov flow at $256 \times 256$ and turbulent channel-flow slice at $192 \times 192$. 
MeLISA outperforms neural-operator baselines on short-term forecasting accuracy and long-horizon statistical metrics, including energy spectra, turbulent kinetic energy, and mixing-rate-related dynamics, while achieving inference speeds comparable to, and in some cases faster than, neural operators. Compact 3.7--5.7M-parameter variants already deliver strong parameter efficiency, and DiT variants provide a scalable path up to 150M parameters. Overall, MeLISA benefits both rollout efficiency and long-horizon statistical accuracy. To our knowledge, this is the \textbf{first} method for high-resolution one-step generation for physical dynamical systems with performance comparable to state-of-the-art deterministic surrogates.
\end{abstract}

\section{Introduction}
Accurate yet efficient simulation of complex dynamical systems remains a central challenge in computational physics. These systems are typically governed by nonlinear partial differential equations (PDEs)~\cite{evans2022partial}, for which obtaining exact solutions is often intractable. As a result, a wide range of numerical techniques have been developed, including Direct Numerical Simulation (DNS)~\cite{lee2015direct, strikwerda2004finite, eymard2000finite, reddy1993introduction}. While such methods can achieve high fidelity by resolving fine-scale structures, they often come with substantial computational cost, making them impractical for many real-world scenarios. To alleviate this computational burden, a variety of reduced-fidelity approaches have been proposed that trade physical resolution for efficiency, including Reynolds-Averaged Navier--Stokes (RANS)~\cite{alfonsi2009reynolds} and Large Eddy Simulation (LES)~\cite{piomelli1999large}. Nevertheless, these methods can remain prohibitively costly in regimes that demand fine spatial resolution, for instance, near solid boundaries where steep gradients must be resolved. Motivated by these limitations and enabled by recent advances in deep learning, data-driven surrogate models have emerged as a promising alternative. Such surrogates aim to approximate the underlying dynamics directly from data, and have been successfully applied across a broad range of scientific domains, including weather forecasting~\cite{gao2023prediff, bi2022pangu}, quantum chemistry~\cite{von2022self, wong2018ferminets}, materials science~\cite{luo2025crystalflow}, and fluid dynamics~\cite{xue2026uni}.

Deterministic neural operators have become a widely used class of data-driven surrogates for PDE-governed and spatiotemporal forecasting problems~\cite{li2020fourier, lu2021learning, rahman2022u, liu2024neural}. When used autoregressively, they map a finite context window to future states and can roll out efficiently. However, long-horizon autoregressive prediction remains challenging: small one-step errors are repeatedly fed back as inputs, causing error accumulation and distribution shift over the rollout~\cite{mccabe2023towards}. This issue is especially pronounced in turbulent or chaotic regimes, where small biases in high-frequency content or temporal correlations can lead to drift in trajectory-level statistics such as energy spectra, turbulent kinetic energy, and invariant-measure-related quantities~\cite{jiang2024training,oommen2024integrating,khodakarami2025mitigating}. 

A complementary direction uses generative models, including diffusion models~\cite{ho2020denoising, song2020score}, flow matching~\cite{lipman2022flow}, and one-step variants such as consistency models~\cite{song2023consistency, lu2024simplifying}, which view sample generation as a transport process from a base distribution to the data distribution. These methods have recently been adopted as autoregressive surrogates for physical dynamics~\cite{kohl2026benchmarking, ruhling2023dyffusion, cachay2025elucidated, xue2026uni, price2025probabilistic, gao2023prediff} and explicitly model conditional distributions over future states. However, in their current form they typically (i) require multi-step denoising or SDE/ODE integration at every rollout step, leading to long inference latency, (ii) rely on frame-wise progressive noise schedules~\cite{cachay2025elucidated, ruhe2024rolling} to roll out stably, and (iii) operate in a latent space induced by an auxiliary VAE or encoder~\cite{ruhe2024rolling, kohl2026benchmarking}, introducing additional training and inference complexity.

In this work, we build on the recently proposed pixel MeanFlow (p-MF) framework~\cite{lu2026one}, a one-step generative model that operates directly in pixel space and avoids both multi-step solvers and latent encoders. To turn p-MF into an autoregressive surrogate that retains fast rollout speed while preserving long-horizon statistical structure, we address two specific questions: how should single-frame MeanFlow be extended to a \emph{window-conditioned spatiotemporal} generator so that one-step generation becomes non-trivial under multi-frame temporal context, and how can long-horizon temporal correlations and mixing behavior be explicitly enforced when only short windows are observed during training? Our answers, \emph{Window-Consistency MeanFlow} and \emph{Time Increment Consistency}, together yield a stochastic autoregressive surrogate that requires a single function evaluation per forecast block (1 NFE per block, where NFE denotes number of function evaluations), combining direct autoregressive rollout with the conditional-distribution modeling of generative methods. See Appendix~\ref{appendix:overview} for an extensive discussion. Our key contributions are summarized as follows:
\begin{itemize}
\item We introduce \textbf{Me}anFlow \textbf{L}ong-term \textbf{I}nvariant \textbf{S}patiotemporal Consistency \textbf{A}utoregressive Models (\textbf{MeLISA}), a stochastic autoregressive surrogate that requires 1-NFE per block, combining direct autoregressive rollout with the conditional-distribution modeling of generative methods, without progressive noise schedules or multi-step SDE solvers.
\item We propose \emph{Window-Consistency MeanFlow}, the first extension of pixel MeanFlow from single-frame generation to a window-conditioned spatiotemporal transition kernel, where masked temporal context is what makes one-step generative forecasting non-trivial.
\item We propose \emph{Time Increment Consistency}, a finite-lag regularizer that directly constrains temporal covariance and mixing structure of the rollout, supplying long-horizon constraints that pointwise state-reconstruction losses provably cannot.
\item MeLISA performs generative forecasting directly in pixel space at up to $256 \times 256$, removing the VAE / latent-encoder / fidelity-enhancement dependencies that current diffusion-based scientific surrogates typically rely on.
\item On extended 2D Kolmogorov flow ($256 \times 256$) and a projected turbulent channel-flow observable ($192 \times 192$), MeLISA matches or exceeds neural-operator baselines on short-term accuracy at comparable rollout speed, while substantially improving recovery of energy spectra, turbulent kinetic energy, and mixing-rate-related dynamics. Compact 3.7--5.7M-parameter variants are already parameter-efficient, and a DiT instantiation scales to 150M parameters.
\end{itemize}

\section{Related Works}
\paragraph{Generative models for dynamical systems.}
Diffusion models and their latent-space variants---including Latent Diffusion Models (LDM)~\cite{ruhling2023dyffusion, cachay2025elucidated, du2024confild} and Latent Flow Matching (LFM)~\cite{lim2024elucidating, mokady2022nulltextinversioneditingreal}, have been widely adopted for dynamical systems forecasting and have achieved state-of-the-art performance. Nevertheless, in the absence of distillation or related acceleration techniques, these approaches typically require many sampling steps at inference time, leading to substantial computational overhead. In contrast, our method builds on a one-step generative formulation, eliminating the need for specialized multi-step sampling procedures. Moreover, we perform generative modeling directly in pixel space, avoiding dimensionality reduction altogether. This enables high-resolution generation up to $256 \times 256$ without training auxiliary latent-space components such as variational autoencoders (VAEs) or fidelity enhancement modules~\cite{shu2023physics, oommen2024integrating, lin2026decoupleddiffusionsamplinginverse, yang2026menomeanflowenhancedneuraloperators}. This design also makes our approach more general and robust, as it reduces the number of system-specific architectural and training choices required in practice. In particular, the one-step nature of our model removes the need to specify a progressive noise schedule, which is a component that most rolling or autoregressive diffusion methods depend on for multi-step sampling~\cite{cachay2025elucidated, ruhe2024rolling, karras2022elucidating, molinaro2024generative}.

\paragraph{One-step generative models.}
One-step (or few-step) generative models have recently gained momentum, motivated by the view of generative modeling as distribution transport~\cite{song2020score}. Representative examples include Consistency Models (CM/sCM)~\cite{song2023consistency, lu2024simplifying}, inductive Moment Matching (iMM)~\cite{zhou2025inductive}, and Shortcut Diffusion~\cite{frans2024one}. In the image generation domain, the MeanFlow (MF) family, including the original MF~\cite{geng2025mean}, improved MeanFlow (i-MF)~\cite{geng2025improved}, and pixel MeanFlow (p-MF)~\cite{lu2026one}, has shown particularly strong performance on both conditional and unconditional generation tasks. Modern generative models for image and video synthesis often achieve few-step or one-step sampling by distilling large pre-trained diffusion models. For example, Consistency Distillation (CD/sCD)~\cite{song2023consistency, lu2024simplifying} has been successfully applied to pre-trained image generators~\cite{luo2023latent} as well as video diffusion models~\cite{mao2025osv, wang2023videolcm}. In contrast, training one-step generative models \emph{from scratch} on video-like data has received comparatively little attention; to the best of our knowledge, this setting has not been systematically explored, nor has it been studied in the context of physical dynamical systems.

\section{Background}
\paragraph{Problem setup.}
Let the sequential dataset be denoted by $\mathcal{D} \in \mathbb{R}^{B \times T \times F}$, where $B$ is the number of trajectories, $T$ is the trajectory length, and $F$ is the feature dimension. We represent a sampled window from discrete physical time $\tau \in \mathbb{Z}^+$ to $\tau+W$ as
\[
\bar{x}^{\tau:\tau+W}
:=
(x^\tau,\ldots,x^{\tau+W-1})
\in \mathbb{R}^{W\times F},
\]
and use the shorthand $\bar{x}_W^\tau$ when the meaning is clear. In probabilistic forecasting, given an input window $\bar{x}_{W_{\mathrm{in}}}^{\tau}$, the goal is to predict a future window $\hat{\bar{x}}_{W_{\mathrm{out}}}^{\tau + W_{\mathrm{in}}}$, where $W_{\mathrm{in}}$ and $W_{\mathrm{out}}$ denote the input and output window lengths, respectively, and may be chosen arbitrarily. This forecasting problem is formalized by the predictive distribution
\begin{equation}
    \hat{\bar{X}}_{W_{\mathrm{out}}}^{\tau + W_{\mathrm{in}}}
    \sim
    p\!\left(
        \hat{\bar{X}}_{W_{\mathrm{out}}}^{\tau + W_{\mathrm{in}}}
        \mid
        \bar{x}_{W_{\mathrm{in}}}^{\tau}
    \right).
\end{equation}

\paragraph{Diffusion models and flow matching.}
Diffusion models can be viewed as distribution transport methods that map a simple base distribution $p_{\mathrm{base}}$ (typically Gaussian) to the target data distribution $p_{\mathrm{data}}$. In denoising diffusion probabilistic models (DDPM)~\cite{ho2020denoising}, this transport is realized through a Markovian diffusion process indexed by diffusion time $t$. A neural denoiser $D_\theta$ is trained to predict the injected noise at each time step:
\begin{equation}
    \mathcal{L}_\mathrm{DDPM}=
        \mathbb{E}_{t,x,\epsilon} \left[\left\|
            D_\theta(z_t, t) - \epsilon
        \right\|_2^2\right],
\end{equation}
where $z_t$ denotes a noisy sample along the diffusion path at time $t$~\cite{ho2020denoising}. This forward--reverse diffusion process can equivalently be described through a stochastic differential equation (SDE) formulation~\cite{song2020score}.

Flow matching approaches the same transport problem from a different perspective by modeling the dynamics with an ordinary differential equation (ODE). In particular, one defines an interpolating noisy state
\begin{equation}
    z_t = a_t x + b_t \epsilon,
\end{equation}
where $\epsilon \sim \mathcal{N}(0, I)$ is Gaussian noise, and $a_t$ and $b_t$ are time-dependent coefficients. In this work, we focus on the linear parameterization $a_t = 1-t$ and $b_t = t$~\cite{liu2022rectified}. The model $D_\theta$ is then trained to predict the transport velocity field 
\begin{equation}
    v(z_t | x, \epsilon) = \frac{\mathrm{d} z_t}{\mathrm{d}t} = a'_t x + b'_t \epsilon,
\end{equation}
via the conditional FM loss~\cite{lipman2022flow}
\begin{equation}
    \mathcal{L}_\mathrm{FM} = 
       \mathbb{E}_{t,x,\epsilon}\left[\left\|
            D_\theta(z_t, t)
            -
                v(z_t | x, \epsilon)
        \right\|^2\right],
\end{equation}
where $z_t$ is the noisy interpolated sample induced by the FM parameterization.

\paragraph{MeanFlow models.}
Standard flow-matching sampling requires multiple evaluations of the velocity network. MeanFlow improves efficiency by predicting the \emph{average} velocity over an interval $[r,t]$~\cite{geng2025mean}:
\begin{equation}
u(z_t,r,t)=\frac{1}{t-r}\int_r^t v(z_\tau \mid x,\epsilon)\,d\tau,
\end{equation}
which enables one-step transport via $z_r = z_t - (t-r)u(z_t,r,t)$. MeanFlow trains a network $D_\theta$ to predict $u$ using a self-consistency identity derived from this definition. To improve stability, \citet{geng2025improved} define
\begin{equation}\label{eq:i-mf}
V_\theta(z_t,r,t)
=
D_\theta
+
(t-r)\,\operatorname{sg}\!\left(
\partial_t D_\theta + v(z_t \mid x,\epsilon)\,\partial_{z_t}D_\theta
\right),
\end{equation}
and optimize the direct regression objective, namely the improved MeanFlow (i-MF) objective
\begin{equation}\label{eq:i-mf_loss}
\mathcal{L}_{\mathrm{i\text{-}MF}}
=
\mathbb{E}_{t,x,\epsilon}
\left[
\left\|
V_\theta(z_t,r,t)-v(z_t \mid x,\epsilon)
\right\|^2
\right].
\end{equation}
This reformulation yields more stable training and better sample quality. Pixel MeanFlow (p-MF) further reparameterizes the model to predict a pixel-space target~\cite{li2026basicsletdenoisinggenerative}:
\begin{equation}\label{eq:pmf}
D_\theta(z_t,r,t)= z_t - t\,u(z_t,r,t),
\end{equation}
so that the same i-MF objective can be applied while biasing prediction toward the data manifold.

\section{Methodology: MeanFlow Long-term Invariant Spatiotemporal Consistency Autoregressive Models}

\begin{figure*}[t]
    \centering    \includegraphics[width=1.\linewidth]{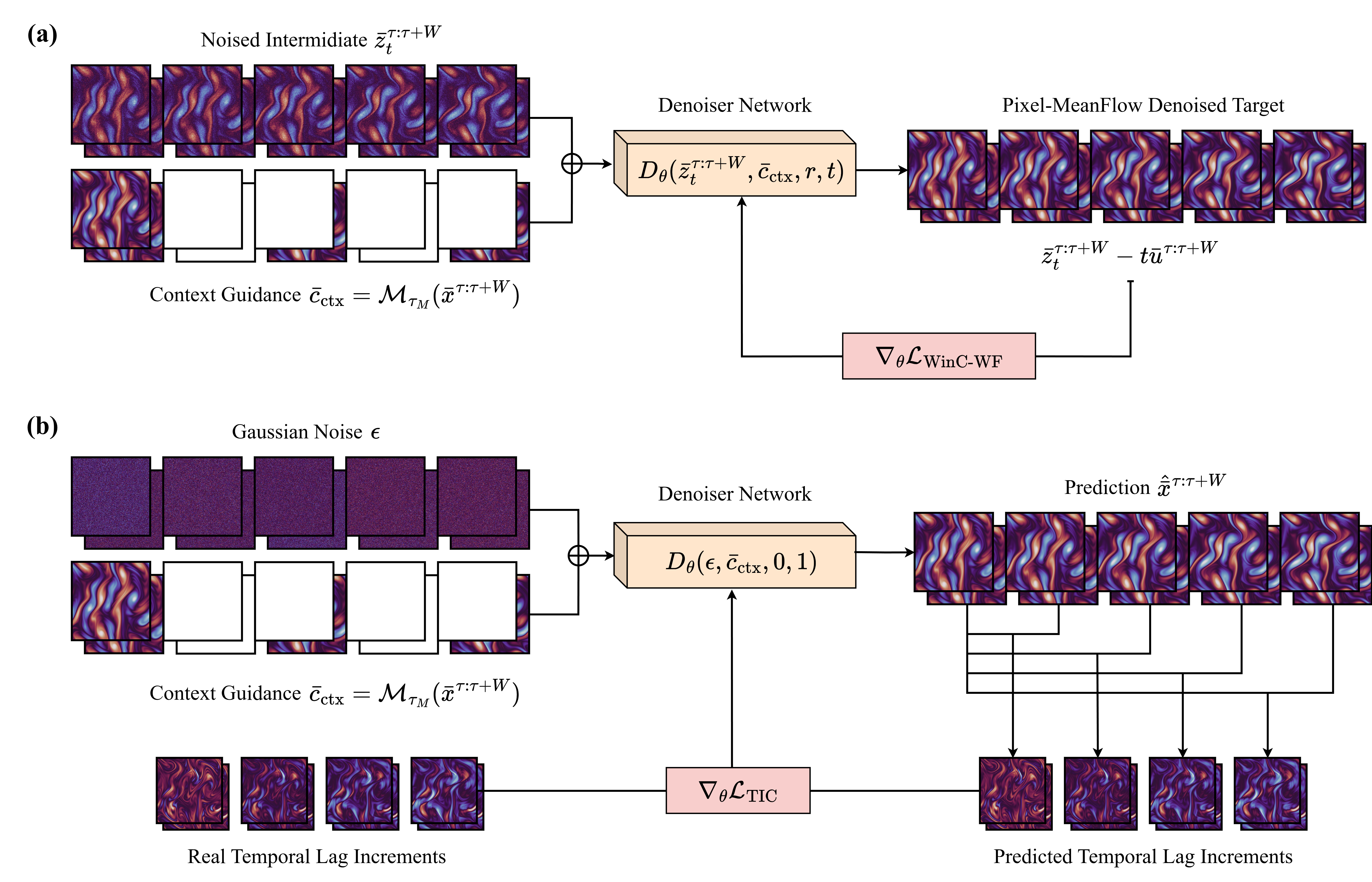}
    \caption{This diagram illustrates the two key mechanisms of MeanFlow Long-term Invariant Spatiotemporal Consistency Autoregressive (MeLISA) models. Panel~(a) shows the \textbf{Window-Consistency MeanFlow} (WinC-MF) objective, in which a MeanFlow denoiser is trained on a temporal window with partially observed, unmasked guidance frames. Panel~(b) shows the \textbf{Time Increment Consistency} (TIC) loss: given a context window, MeLISA first produces a reconstruction, and TIC then penalizes the discrepancy between the predicted and ground-truth temporal increments across different lags.}
    \label{fig:MeLISA_framework}
    \vspace{-1em}
\end{figure*}

Our method, \textbf{MeLISA}, incorporates physics-aware prediction into the MeanFlow framework through two key mechanisms: the \textbf{Window-Consistency MeanFlow} objective and the \textbf{Time Increment Consistency} regularizer.

\subsection{Window Consistency MeanFlow}
Prior work on autoregressive forecasting has shown that long-horizon stability improves with larger input or output windows~\cite{mccabe2025walrus,li2020fourier,rahman2022u,molinaro2024generative}. This is especially important for consistency-sensitive physical statistics, such as autocorrelation functions. Motivated by this observation, we propose a MeanFlow formulation that explicitly enforces temporal consistency over a window.

Following Eq.~\ref{eq:pmf}, the average velocity for a single sample is
\begin{equation}
    u_\theta(z_t,r,t)=\frac{1}{t}\bigl[z_t-D_\theta(z_t,r,t)\bigr],
\end{equation}
which induces the instantaneous velocity
\begin{equation}\label{eq:imf_main}
    V_\theta(z_t,r,t)
    =
    u_\theta
    +
    (t-r)\,\operatorname{sg}\!\left(
        \partial_t u_\theta
        +
        v(z_t \mid x,\epsilon)\,\partial_{z_t}u_\theta
    \right).
\end{equation}

We extend this construction from a single frame to a temporal window. For a clean window $\bar{x}^{\tau:\tau+W}$, we define the noisy window
\begin{equation}
    \bar{z}_t^{\tau:\tau+W}
    =
    (1-t)\,\bar{x}^{\tau:\tau+W}
    +
    t\,\epsilon,
\end{equation}
where the diffusion times $r$ and $t$ are shared across all frames in the window. The corresponding average velocity is
\begin{equation}
    \bar{u}_\theta^{\tau:\tau+W}
    (\bar{z}_t^{\tau:\tau+W},r,t)
    =
    \frac{1}{t}
    \left[
        \bar{z}_t^{\tau:\tau+W}
        -
        D_\theta(\bar{z}_t^{\tau:\tau+W},r,t)
    \right].
\end{equation}

To enforce consistency under partial observation, let $\mathcal{M}_{\tau_M}$ denote a masking operator that removes the frames indexed by $\tau_M$ within the window. We then define
\begin{equation}
    \bar{u}_\theta^{\tau:\tau+W}
    (\bar{z}_t^{\tau:\tau+W},r,t,\tau_M)
    =
    \frac{1}{t}
    \left[
        \bar{z}_t^{\tau:\tau+W}
        -
        D_\theta\!\left(\bar{z}_t^{\tau:\tau+W}, 
            \mathcal{M}_{\tau_M}(\bar{x}^{\tau:\tau+W}),\,r,\,t
        \right)
    \right].
\end{equation}
The on-window instantaneous velocity is
\begin{equation}
    \bar{V}_\theta^{\tau:\tau+W}
    (\bar{z}_t^{\tau:\tau+W},r,t,\tau_M)
    =
    \bar{u}_\theta^{\tau:\tau+W}
    +
    (t-r)\,\operatorname{sg}\!\left(
        \partial_t \bar{u}_\theta^{\tau:\tau+W}
        +
        \bar{v}^{\tau:\tau+W}\,
        \partial_{\bar{z}_t^{\tau:\tau+W}}
        \bar{u}_\theta^{\tau:\tau+W}
    \right),
\end{equation}
where
\begin{equation}
    \bar{v}^{\tau:\tau+W}
    \bigl(
        \bar{z}_t^{\tau:\tau+W}
        \mid
        \bar{x}^{\tau:\tau+W},\epsilon
    \bigr)
    =
    \epsilon-\bar{x}^{\tau:\tau+W}.
\end{equation}
This yields the \textbf{Window-Consistency MeanFlow} (WinC-MF) objective,
\begin{equation}\label{eq:winc_mf}
    \mathcal{L}_{\mathrm{WinC\text{-}MF}}
    =
    \mathbb{E}_{t,r,x,\epsilon,\tau,\tau_M}
        \left\|
            \bar{V}_\theta^{\tau:\tau+W}
            (\bar{z}_t^{\tau:\tau+W},r,t,\tau_M)
            -
            \bar{v}^{\tau:\tau+W}
            \bigl(
                \bar{z}_t^{\tau:\tau+W}
                \mid
                \bar{x}^{\tau:\tau+W},\epsilon
            \bigr)
        \right\|^2_2.
\end{equation}
This objective encourages temporally consistent predictions from partially observed windows; its connection to self-supervised learning (SSL) is discussed in Appendix~\ref{appendix:winc_mf}. This is illustrated in the panel (a) of Fig. \ref{fig:MeLISA_framework}. In practice, we sample the mask over later frames from a Bernoulli distribution, while always retaining the first frame as a temporal reference.

\subsection{Time Increment Consistency}

Let $x^\tau$ denote the ground-truth field at physical time $\tau$, and $\hat{x}^\tau$ its reconstruction. For a window of length $W$, we define the lag-$w$ temporal increment as
\begin{equation}
    \Delta x^{\tau,\tau+w} = x^{\tau+w} - x^\tau,
    \qquad
    w=1,\dots,W-1.
\end{equation}
Given a window $\bar{x}^{\tau:\tau+W}$, the MeLISA reconstruction is
\begin{equation}
    \hat{\bar{x}}^{\tau:\tau+W}
=
D_\theta\!\left(
\mathrm{Concat}\!\left[
\epsilon,\,
\mathcal{M}_{\tau_M}(\bar{x}^{\tau:\tau+W}),\,
\tau_M
\right],
0,1
\right).
\end{equation}
We then define the \textbf{Time Increment Consistency} (TIC) loss as
\begin{equation}
    \mathcal{L}_{\mathrm{TIC}}
    =
    \sum_{w=1}^{W-1}
    \kappa_w\,
    \mathbb{E}_{x,\tau,\tau_M, \epsilon}
    \left[
        \left\|
            \Delta x^{\tau,\tau+w}
            -
            \Delta \hat{x}^{\tau,\tau+w}
        \right\|_2^2
    \right],
\end{equation}
where $\kappa_w$ is a lag-dependent weight. This objective enforces agreement between the temporal increments of the reconstructed and ground-truth trajectories across multiple lags. The full training objective is
\begin{equation}
    \mathcal{L}_{\mathrm{MeLISA}}
    =
    \mathcal{L}_{\mathrm{WinC\text{-}MF}}
    +
    \mathcal{L}_{\mathrm{TIC}}.
\end{equation}
TIC encourages recovery of long-range spatiotemporal correlations by constraining second-order temporal structure, which cannot be captured by per-frame state reconstruction losses alone. This mechanism is illustrated in the panel (b) of Fig. \ref{fig:MeLISA_framework}. Further discussion is provided in Appendix~\ref{appendix:tic}. The training procedure of MeLISA is summarized in Algorithm \ref{alg:training}. Inference for each forecast block requires only \textbf{one model evaluation} (1-NFE), see Algorithm \ref{alg:inference}. We report that MeLISA models inference speed is as fast as neural operators, see Appendix \ref{appendix:cost_analysis} for details.

\section{Experiments}
\paragraph{Dataset.}
In this section, we evaluate MeLISA on two high-resolution benchmark datasets: the $192 \times 192$ turbulent channel flow dataset (\textbf{TCF192}) and the $256 \times 256$ 2D Kolmogorov flow dataset (\textbf{KF256}). These datasets represent two distinct types of two-dimensional dynamics. KF256 is governed by a closed-form 2D Partial Different Equation (PDE) with periodic external forcing, whereas TCF192 is obtained by slicing a 3D turbulent flow and therefore corresponds to projected dynamics rather than a fully observed closed system. As a result, TCF192 contains stronger non-Markovian effects, making its long-range statistics more challenging to model. See Appendix \ref{appendix:dataset} for more details. 

\begin{figure*}[t]
    \centering    \includegraphics[width=1.\linewidth]{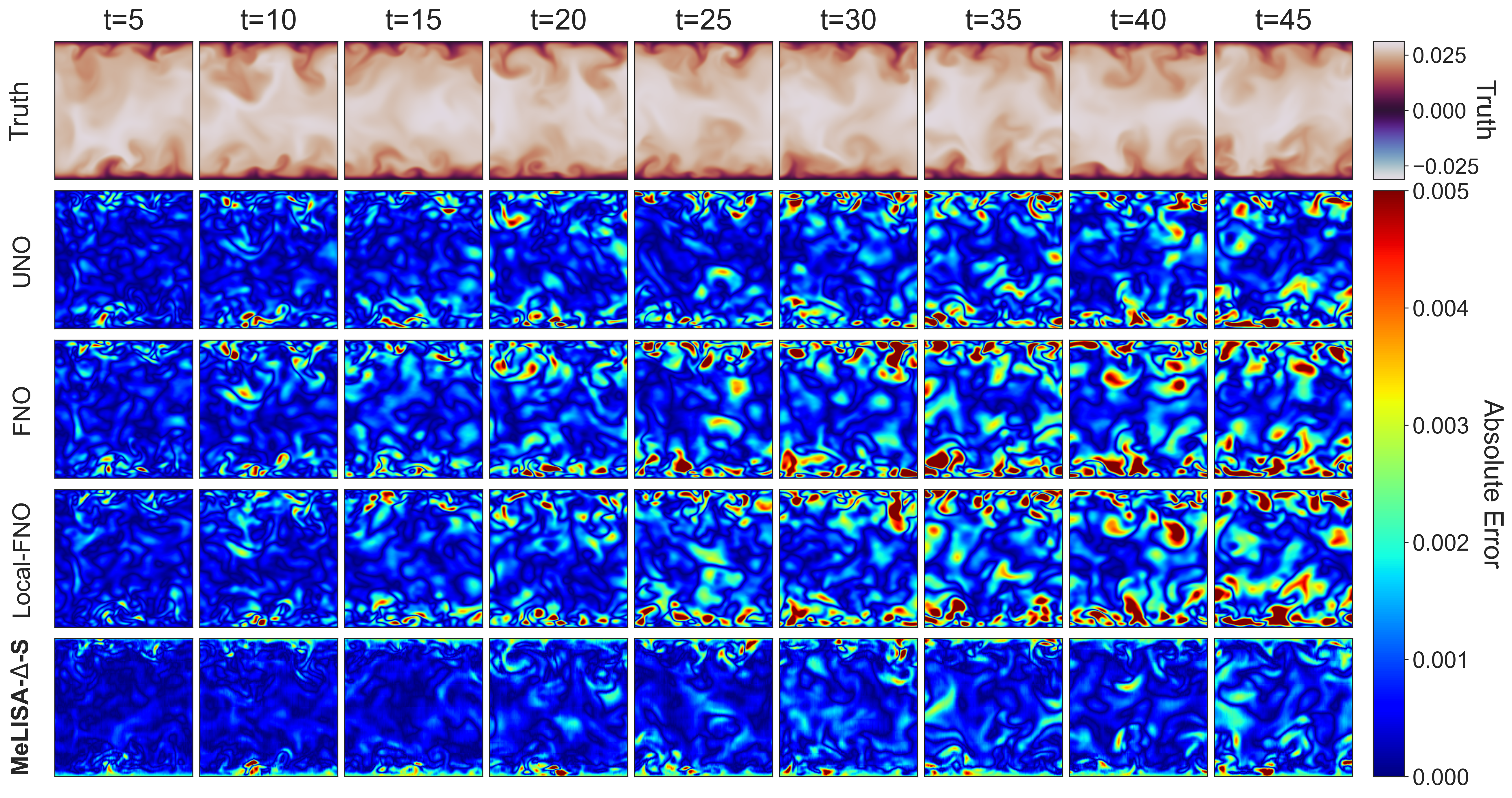}
    \caption{Rollout results on an uncurated test trajectory from \textbf{TCF192}, shown as absolute error with respect to the ground truth over the prediction horizon. We compare three autoregressive baselines with one MeLISA variant, MeLISA-$\Delta$-S. Here, $t$ denotes the frame index within the trajectory. MeLISA exhibits substantially improved stability and markedly slower error accumulation.}
    \label{fig:tcf_rollout}
    \vspace{-1em}
\end{figure*}

\paragraph{MeLISA model instantiations.}
We evaluate MeLISA with two backbone families. \textbf{MeLISA-$\Upsilon$} uses an attention-augmented UNet architecture following the DDPM style~\cite{ho2020denoising,nichol2021improved}, extended with 3D convolutions to process spatiotemporal inputs. We consider two compact variants: MeLISA-$\Upsilon$-XS (3M parameters) and MeLISA-$\Upsilon$-S (5M parameters), highlighting the parameter efficiency of the framework. To study scaling behavior, we also instantiate MeLISA with diffusion transformers (DiTs)~\cite{peebles2023scalable} under the pixel-space formulation of \citet{li2026basicsletdenoisinggenerative}. We denote this family by \textbf{MeLISA-$\Delta$}, and scale it from 10M parameters (MeLISA-$\Delta$-S) to 150M parameters (MeLISA-$\Delta$-B), limited by the size of the training data. Following prior work~\cite{li2026basicsletdenoisinggenerative}, we train MeLISA-$\Delta$ with the Muon optimizer~\cite{liu2025muon}, which has been shown to provide faster convergence and greater stability for the p-MF formulation~\cite{lu2026one}. Importantly, to remain consistent with the design of WinC-MF, we \textbf{do not introduce explicit temporal attention modules} in either backbone family. Full architectural details and training hyperparameters are provided in Table~\ref{tab:melisa_table}.

\paragraph{Baseline models.}
We benchmark MeLISA against deterministic autoregressive baselines. \textbf{Neural operators}~\cite{li2020fourier,lu2021learning} are widely used for dynamical-system forecasting and have shown strong performance across many benchmarks. We consider three representative neural-operator models: the original Fourier Neural Operator (FNO)~\cite{li2020fourier}, the U-shaped Neural Operator (UNO)~\cite{rahman2022u}, which adopts a UNet-style architecture for improved parameter efficiency, and the Local Fourier Neural Operator (Local FNO)~\cite{liu2024neural}, which incorporates local convolutional kernels to improve predictive accuracy. For a fair comparison, all neural-operator baselines are trained with two input frames and four output frames, matching the MeLISA setting. We defer discussion of \textbf{rolling generative models}, the family most closely related to MeLISA and increasingly used for spatiotemporal prediction~\cite{ruhling2023dyffusion,cachay2025elucidated,lim2024elucidating,molinaro2024generative}, to Appendix~\ref{appendix:overview}.

\paragraph{Metrics.}
We evaluate all models in an autoregressive rollout setting, initialized with the first two frames from each test trajectory. Performance is measured using five complementary metrics: relative $L_2$ error (R$L_2$), structural similarity index measure (SSIM), power spectral density discrepancy (PSDD), turbulent kinetic energy difference (TKED), and mixing rate difference (MRD). SSIM captures perceptual and structural agreement by emphasizing local patterns such as edges, textures, and contrast~\cite{wang2004image,hore2010image}, while R$L_2$ measures pointwise reconstruction accuracy. Since both metrics primarily reflect instantaneous prediction quality, we report them over the \textbf{short-term prediction horizon}, corresponding to the first \textbf{40 frames} for both sets. In contrast, PSDD measures how well a model preserves the distribution of energy across spatial frequencies, which is crucial for multi-scale dynamics and turbulent flows~\cite{hess2023generalized,volkmann2024scalable}. TKED quantifies the discrepancy in trajectory-level turbulent kinetic energy (TKE) relative to the ground truth. The mixing rate is defined as the decay exponent of the full-trajectory autocorrelation curve, estimated from the first 20 frames, and MRD measures the corresponding error in this quantity. Because PSDD, TKED, and MRD characterize trajectory-level statistical behavior, they are computed over the \textbf{full rollout trajectory}. Further details are provided in Appendix~\ref{appendix:metrics}. All reported results are averaged over evaluations.

\begin{figure*}[t]
    \centering    \includegraphics[width=1.\linewidth]{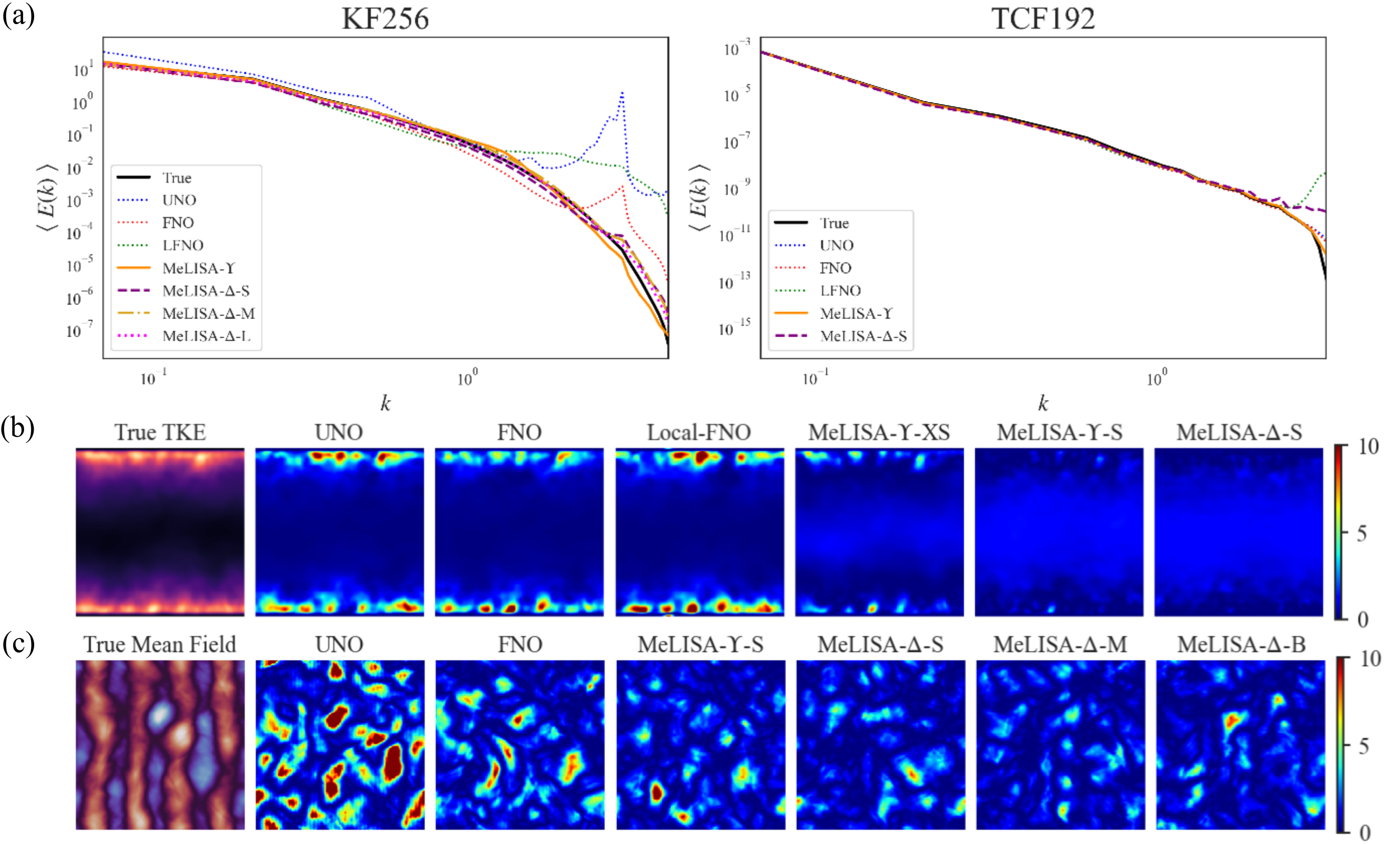}
    \caption{(a) Radially averaged energy spectra for KF256 and TCF192. The neural-operator baselines tend to overemphasize low-frequency modes and introduce spurious peaks at higher frequencies. In contrast, MeLISA accurately reproduces the high-frequency tail without any explicit spectral regularization. 
    (b) Spatial distribution of turbulent kinetic energy (TKE), shown together with scaled \textbf{absolute error} relative to the ground truth for all models. Only the MeLISA variants correctly recover the kinetic energy near the boundaries. 
    (c) Time-averaged mean field on KF256 test trajectories, shown with scaled absolute error. MeLISA again reproduces the correct large-scale pattern more faithfully. Local-FNO is omitted because it cannot roll out stably over the full trajectory length. Error scaling is applied only for visualization.}
    \label{fig:stats}
    \vspace{-1em}
\end{figure*}

\subsection{Turbulent Channel Flow at $192 \times 192$}
We report results on the TCF192 dataset in this section. Rollout visualizations are shown in Fig.~\ref{fig:tcf_rollout}, and the corresponding metrics are summarized in Table~\ref{tab:tcf_table}. Despite having only \mbox{$\sim$3M} parameters, MeLISA-$\Upsilon$-XS performs competitively against autoregressive neural operators and achieves the best TKED score. The MeLISA-$\Delta$ variants obtain the strongest short-term metrics (R$L_2$ and SSIM). The radial energy spectrum for TCF192 is shown in the right panel of Fig.~\ref{fig:stats}(a). All models capture the low- to mid-frequency range reasonably well, but most exhibit a spurious peak in the high-frequency tail, deviating from the expected spectral decay. Among all methods, the $\Upsilon$ variants match the tail behavior most faithfully. The spatial distribution of turbulent kinetic energy is shown in Fig.~\ref{fig:stats}(b). Neural operators fail to reproduce the correct near-boundary behavior, whereas all MeLISA variants recover the boundary energy distribution much more accurately. 

\begin{table}[!h]
\centering
\caption{TCF192 results for all models on short-term and long-term evaluation metrics with uncertainties. The $\Delta$ variants achieve the strongest short-term predictive performance, whereas the $\Upsilon$ variants preserve long-range statistical structure more effectively. Short-term metrics are computed over the first 40 frames, and long-term metrics are evaluated over the full 625-frame trajectory.}
\setlength{\tabcolsep}{0.4em}
\begin{tabular}{lc|cc|ccc}
\toprule
\multicolumn{2}{c|}{\textbf{Model Specs}}  
& \multicolumn{2}{c|}{\textbf{Short-term metrics}} 
& \multicolumn{3}{c}{\textbf{Long-term metrics}} \\
\midrule
\textbf{Model Name} & \textbf{Model Size}  
& \textbf{R$L_2$}$\downarrow$ 
& \textbf{SSIM}$\uparrow$ 
& \textbf{PSDD}$\downarrow$ 
& \textbf{TKED}$\downarrow$ 
& \textbf{MRD}$\downarrow$ \\
\midrule

\rowcolor{gray!15} 
\multicolumn{7}{l}{\textit{MeLISA Models (1-NFE)}}\\

MeLISA-$\Upsilon$-XS 
& 3.68M 
& 0.0831\unc{12} 
& 0.882\unc{20} 
& 0.0897\unc{21} 
& \textbf{1.20}\unc{12}\,$\times 10^{-6}$ 
& 0.475\unc{12} \\

MeLISA-$\Upsilon$-S 
& 5.65M 
& 0.0532\unc{88} 
& 0.925\unc{16} 
& \textbf{0.0635}\unc{90} 
& 1.42\unc{13}\,$\times 10^{-6}$ 
& \textbf{0.137}\unc{13} \\

MeLISA-$\Delta$-S   
& 10.5M 
& \textbf{0.0396}\unc{33} 
& \textbf{0.956}\unc{8} 
& 0.0660\unc{5} 
& 1.47\unc{1}\,$\times 10^{-6}$ 
& 0.196\unc{3} \\

\rowcolor{gray!15} 
\multicolumn{7}{l}{\textit{Autoregressive Neural Operators}}\\

FNO                 
& 6.68M 
& 0.0617\unc{37} 
& 0.908\unc{7} 
& 0.0649\unc{13} 
& 1.59\unc{12}\,$\times 10^{-6}$ 
& 0.476\unc{36} \\

UNO                 
& 5.98M 
& 0.0617\unc{80} 
& 0.906\unc{16} 
& 0.0639\unc{15} 
& 1.89\unc{23}\,$\times 10^{-6}$ 
& 0.538\unc{19} \\

Local-FNO           
& 6.76M 
& 0.0579\unc{47} 
& 0.913\unc{9} 
& 0.0772\unc{54} 
& 1.99\unc{54}\,$\times 10^{-6}$ 
& 0.507\unc{6} \\

\bottomrule
\end{tabular}
\label{tab:tcf_table}
\end{table}

\subsection{Kolmogorov Flow at $256 \times 256$}
We summarize the results on KF256 in this section, with particular emphasis on the scaling behavior of MeLISA. The metrics are summarized in Table \ref{tab:km_table}. We train three MeLISA-$\Delta$ variants on KF256, ranging from 10.5M to 150M parameters. Interestingly, although the short-term metrics remain competitive, they slightly degrade with increasing model size, whereas the long-term metrics, especially PSDD and TKED, generally improve with scale, albeit not monotonically, which is due to our cost-matching training setup (See Appendix \ref{appendix:training_setting}) and the limited size of dataset. Comparing the two backbone families, the MeLISA-$\Upsilon$ variants achieve stronger short-term prediction metrics than the $\Delta$ variants. Overall, all MeLISA models outperform the neural-operator baselines on short-term accuracy and on most long-term statistical metrics. The radial energy spectrum is shown in the left panel of Fig.~\ref{fig:stats}(a). Neural operators exhibit a clear bias at high frequencies, with UNO in particular producing a spurious peak in the spectral tail. In contrast, all MeLISA variants recover the correct high-frequency decay much more faithfully. The time-averaged mean field is shown in Fig.~\ref{fig:stats}(c), where all MeLISA models reproduce the overall spatial pattern accurately.

\begin{table}[ht]
\centering
\caption{KF256 results for all models on short-term and long-term evaluation metrics with uncertainties. Short-term metrics are computed over the first 40 frames, and long-term metrics are evaluated over the full 320-frame trajectory. Long-term metrics for Local-FNO are omitted because Local-FNOs cannot roll out stably over the full trajectory length.}
\setlength{\tabcolsep}{0.5em}
\begin{tabular}{lc|cc|ccc}
\toprule
\multicolumn{2}{c|}{\textbf{Model Specs}}  
& \multicolumn{2}{c|}{\textbf{Short-term metrics}} 
& \multicolumn{3}{c}{\textbf{Long-term metrics}} \\
\midrule
\textbf{Model Name} & \textbf{Model Size} 
& \textbf{R$L_2$}$\downarrow$ 
& \textbf{SSIM}$\uparrow$ 
& \textbf{PSDD}$\downarrow$ 
& \textbf{TKED}$\downarrow$ 
& \textbf{MRD}$\downarrow$ \\
\midrule

\rowcolor{gray!15} 
\multicolumn{7}{l}{\textit{MeLISA Models (1-NFE)}}\\

MeLISA-$\Upsilon$-XS 
& 3.68M 
& \textbf{0.130}\unc{15} 
& \textbf{0.925}\unc{10} 
& 0.353\unc{16} 
& 6.89\unc{72} 
& \textbf{0.065}\unc{54} \\

MeLISA-$\Upsilon$-S 
& 5.65M 
& 0.137\unc{15} 
& 0.919\unc{11} 
& 0.347\unc{18} 
& 6.73\unc{71} 
& 0.068\unc{39} \\

MeLISA-$\Delta$-S   
& 10.5M  
& 0.194\unc{33} 
& 0.867\unc{29} 
& 0.450\unc{10} 
& 5.17\unc{50} 
& 0.096\unc{41} \\

MeLISA-$\Delta$-M   
& 58.3M  
& 0.156\unc{31} 
& 0.903\unc{44} 
& 0.445\unc{78} 
& \textbf{4.24}\unc{64} 
& 0.086\unc{20} \\

MeLISA-$\Delta$-B   
& 150M  
& 0.174\unc{44} 
& 0.887\unc{36} 
& \textbf{0.342}\unc{37} 
& 4.73\unc{26} 
& 0.130\unc{4} \\

\rowcolor{gray!15} 
\multicolumn{7}{l}{\textit{Autoregressive Neural Operators}}\\

FNO       
& 6.68M  
& 0.192\unc{12} 
& 0.845\unc{10} 
& 1.15\unc{5} 
& 5.72\unc{92} 
& 0.168\unc{88} \\

UNO       
& 5.98M  
& 0.187\unc{19} 
& 0.857\unc{9} 
& 1.42\unc{19} 
& 6.00\unc{57} 
& 0.338\unc{22} \\

Local-FNO 
& 6.76M  
& 0.549\unc{85} 
& 0.744\unc{31} 
& - 
& - 
& - \\

\bottomrule
\end{tabular}
\label{tab:km_table}
\end{table}
\vspace{-0.6em}

\section{Conclusion}

We introduced \textbf{MeLISA}, a latent-free autoregressive generative surrogate for dynamical systems that combines the efficiency of pixel-space MeanFlow with training objectives designed to improve long-horizon stability and physical consistency. By formulating forecasting directly in pixel space, MeLISA avoids the additional modeling and inference overhead associated with latent compression, while still enabling one-step generation over multiple future frames. To address the difficulty of preserving realistic dynamics under repeated rollout, we proposed two key components: the \emph{Window-Consistency MeanFlow} objective, which encourages temporally coherent predictions from partially observed windows, and the \emph{Time Increment Consistency} regularizer, which constrains temporal increments across multiple lags to better preserve long-range spatiotemporal structure.

Across two challenging high-resolution benchmarks, turbulent channel flow at $192 \times 192$ and Kolmogorov flow at $256 \times 256$, MeLISA consistently delivered strong short-term predictive accuracy while more faithfully recovering trajectory-level physical statistics than autoregressive neural-operator baselines. In particular, the proposed framework showed improved preservation of energy spectra, turbulent kinetic energy, and mixing-related behavior over long rollouts, indicating that accurate frame-wise prediction alone is insufficient for physically realistic surrogate modeling, and that explicit regularization of temporal structure is beneficial.

Overall, these results position MeLISA as a promising next-generation generative surrogate for scientific machine learning: it is efficient at inference, scalable to high-resolution fields, and better aligned with the statistical requirements of physical dynamical systems. Future work may extend this framework to fully three-dimensional systems, longer-context and multi-scale forecasting settings, and broader classes of partially observed or non-Markovian dynamics.

\section{Acknowledgments}

 Authors acknowledge the use of resources provided by the Isambard-AI National AI Research Resource (AIRR). Isambard-AI is operated by the University of Bristol and funded by the UK Government's Department for Science, Innovation and Technology (DSIT) through UK Research and Innovation (UKRI). This work was supported under project proposals ``GenTurb3D".

\bibliographystyle{unsrtnat}
\bibliography{ref}

@article{li2020fourier,
  title={Fourier neural operator for parametric partial differential equations},
  author={Li, Zongyi and Kovachki, Nikola and Azizzadenesheli, Kamyar and Liu, Burigede and Bhattacharya, Kaushik and Stuart, Andrew and Anandkumar, Anima},
  journal={arXiv preprint arXiv:2010.08895},
  year={2020}
}

@article{li2022learning,
  title={Learning chaotic dynamics in dissipative systems},
  author={Li, Zongyi and Liu-Schiaffini, Miguel and Kovachki, Nikola and Azizzadenesheli, Kamyar and Liu, Burigede and Bhattacharya, Kaushik and Stuart, Andrew and Anandkumar, Anima},
  journal={Advances in Neural Information Processing Systems},
  volume={35},
  pages={16768--16781},
  year={2022}
}

@article{lu2021learning,
  title={Learning nonlinear operators via DeepONet based on the universal approximation theorem of operators},
  author={Lu, Lu and Jin, Pengzhan and Pang, Guofei and Zhang, Zhongqiang and Karniadakis, George Em},
  journal={Nature machine intelligence},
  volume={3},
  number={3},
  pages={218--229},
  year={2021},
  publisher={Nature Publishing Group UK London}
}

@article{jiang2024training,
  title={Training neural operators to preserve invariant measures of chaotic attractors},
  author={Jiang, Ruoxi and Lu, Peter Y and Orlova, Elena and Willett, Rebecca},
  journal={Advances in Neural Information Processing Systems},
  volume={36},
  year={2024}
}

@article{yan2024emerging,
  title={Emerging opportunities and challenges for the future of reservoir computing},
  author={Yan, Min and Huang, Can and Bienstman, Peter and Tino, Peter and Lin, Wei and Sun, Jie},
  journal={Nature Communications},
  volume={15},
  number={1},
  pages={2056},
  year={2024},
  publisher={Nature Publishing Group UK London}
}

@article{rahman2022u,
  title={U-no: U-shaped neural operators},
  author={Rahman, Md Ashiqur and Ross, Zachary E and Azizzadenesheli, Kamyar},
  journal={arXiv preprint arXiv:2204.11127},
  year={2022}
}

@article{xue2022synthetic,
  title={Synthetic turbulence generator for lattice {B}oltzmann method at the interface between {RANS} and {LES}},
  author={Xue, Xiao and Yao, Hua-Dong and Davidson, Lars},
  journal={Physics of Fluids},
  volume={34},
  number={5},
  pages={055118},
  year={2022},
  publisher={AIP Publishing LLC}
}

@book{succi2001lattice,
  title={The {L}attice {B}oltzmann {E}quation for {F}luid {D}ynamics and {B}eyond},
  author={Succi, Sauro},
  year={2001},
  publisher={Oxford University Press}
}

@article{latt2008straight,
  title={Straight velocity boundaries in the lattice Boltzmann method},
  author={Latt, Jonas and Chopard, Bastien and Malaspinas, Orestis and Deville, Michel and Michler, Andreas},
  journal={Physical Review E—Statistical, Nonlinear, and Soft Matter Physics},
  volume={77},
  number={5},
  pages={056703},
  year={2008},
  publisher={APS}
}

@article{Kochkov2021-ML-CFD,
  author = {Kochkov, Dmitrii and Smith, Jamie A. and Alieva, Ayya and Wang, Qing and Brenner, Michael P. and Hoyer, Stephan},
  title = {Machine learning{\textendash}accelerated computational fluid dynamics},
  volume = {118},
  number = {21},
  elocation-id = {e2101784118},
  year = {2021},
  doi = {10.1073/pnas.2101784118},
  publisher = {National Academy of Sciences},
  issn = {0027-8424},
  URL = {https://www.pnas.org/content/118/21/e2101784118},
  journal = {Proceedings of the National Academy of Sciences}
}

@article{lee2015direct,
  title={Direct numerical simulation of turbulent channel flow up to},
  author={Lee, Myoungkyu and Moser, Robert D},
  journal={Journal of fluid mechanics},
  volume={774},
  pages={395--415},
  year={2015},
  publisher={Cambridge University Press}
}

@article{alfonsi2009reynolds,
  title={Reynolds-averaged Navier--Stokes equations for turbulence modeling},
  author={Alfonsi, Giancarlo},
  year={2009}
}

@article{piomelli1999large,
  title={Large-eddy simulation: achievements and challenges},
  author={Piomelli, Ugo},
  journal={Progress in aerospace sciences},
  volume={35},
  number={4},
  pages={335--362},
  year={1999},
  publisher={Elsevier}
}

@book{evans2022partial,
  title={Partial differential equations},
  author={Evans, Lawrence C},
  volume={19},
  year={2022},
  publisher={American mathematical society}
}

@article{khodakarami2025mitigating,
  title={Mitigating spectral bias in neural operators via high-frequency scaling for physical systems},
  author={Khodakarami, Siavash and Oommen, Vivek and Bora, Aniruddha and Karniadakis, George Em},
  journal={arXiv preprint arXiv:2503.13695},
  year={2025}
}

@article{ho2020denoising,
  title={Denoising diffusion probabilistic models},
  author={Ho, Jonathan and Jain, Ajay and Abbeel, Pieter},
  journal={Advances in neural information processing systems},
  volume={33},
  pages={6840--6851},
  year={2020}
}

@article{lipman2022flow,
  title={Flow matching for generative modeling},
  author={Lipman, Yaron and Chen, Ricky TQ and Ben-Hamu, Heli and Nickel, Maximilian and Le, Matt},
  journal={arXiv preprint arXiv:2210.02747},
  year={2022}
}

@article{geng2025mean,
  title={Mean flows for one-step generative modeling},
  author={Geng, Zhengyang and Deng, Mingyang and Bai, Xingjian and Kolter, J Zico and He, Kaiming},
  journal={arXiv preprint arXiv:2505.13447},
  year={2025}
}

@article{geng2025improved,
  title={Improved Mean Flows: On the Challenges of Fastforward Generative Models},
  author={Geng, Zhengyang and Lu, Yiyang and Wu, Zongze and Shechtman, Eli and Kolter, J Zico and He, Kaiming},
  journal={arXiv preprint arXiv:2512.02012},
  year={2025}
}

@article{wang2004image,
  title={Image quality assessment: from error visibility to structural similarity},
  author={Wang, Zhou and Bovik, Alan C and Sheikh, Hamid R and Simoncelli, Eero P},
  journal={IEEE transactions on image processing},
  volume={13},
  number={4},
  pages={600--612},
  year={2004},
  publisher={IEEE}
}

@inproceedings{hore2010image,
  title={Image quality metrics: PSNR vs. SSIM},
  author={Hore, Alain and Ziou, Djemel},
  booktitle={2010 20th international conference on pattern recognition},
  pages={2366--2369},
  year={2010},
  organization={IEEE}
}

@article{shu2023physics,
  title={A physics-informed diffusion model for high-fidelity flow field reconstruction},
  author={Shu, Dule and Li, Zijie and Farimani, Amir Barati},
  journal={Journal of Computational Physics},
  volume={478},
  pages={111972},
  year={2023},
  publisher={Elsevier}
}

@article{hess2023generalized,
  title={Generalized teacher forcing for learning chaotic dynamics},
  author={Hess, Florian and Monfared, Zahra and Brenner, Manuel and Durstewitz, Daniel},
  journal={arXiv preprint arXiv:2306.04406},
  year={2023}
}

@article{volkmann2024scalable,
  title={A scalable generative model for dynamical system reconstruction from neuroimaging data},
  author={Volkmann, Eric and Br{\"a}ndle, Alena and Durstewitz, Daniel and Koppe, Georgia},
  journal={Advances in Neural Information Processing Systems},
  volume={37},
  pages={80328--80362},
  year={2024}
}

@inproceedings{mokady2022nulltextinversioneditingreal,
  title={Null-text inversion for editing real images using guided diffusion models},
  author={Mokady, Ron and Hertz, Amir and Aberman, Kfir and Pritch, Yael and Cohen-Or, Daniel},
  booktitle={Proceedings of the IEEE/CVF conference on computer vision and pattern recognition},
  pages={6038--6047},
  year={2023}
}

@article{song2023consistency,
  title={Consistency models},
  author={Song, Yang and Dhariwal, Prafulla and Chen, Mark and Sutskever, Ilya},
  year={2023}
}

@article{frans2024one,
  title={One step diffusion via shortcut models},
  author={Frans, Kevin and Hafner, Danijar and Levine, Sergey and Abbeel, Pieter},
  journal={arXiv preprint arXiv:2410.12557},
  year={2024}
}

@article{zhou2025inductive,
  title={Inductive moment matching},
  author={Zhou, Linqi and Ermon, Stefano and Song, Jiaming},
  journal={arXiv preprint arXiv:2503.07565},
  year={2025}
}

@article{lu2024simplifying,
  title={Simplifying, stabilizing and scaling continuous-time consistency models},
  author={Lu, Cheng and Song, Yang},
  journal={arXiv preprint arXiv:2410.11081},
  year={2024}
}

@article{oommen2024integrating,
  title={Integrating neural operators with diffusion models improves spectral representation in turbulence modeling},
  author={Oommen, Vivek and Bora, Aniruddha and Zhang, Zhen and Karniadakis, George Em},
  journal={arXiv preprint arXiv:2409.08477},
  year={2024}
}

@book{strikwerda2004finite,
  title={Finite difference schemes and partial differential equations},
  author={Strikwerda, John C},
  year={2004},
  publisher={SIAM}
}

@article{eymard2000finite,
  title={Finite volume methods},
  author={Eymard, Robert and Gallou{\"e}t, Thierry and Herbin, Rapha{\`e}le},
  journal={Handbook of numerical analysis},
  volume={7},
  pages={713--1018},
  year={2000},
  publisher={Elsevier}
}

@article{reddy1993introduction,
  title={An introduction to the finite element method},
  author={Reddy, Junuthula Narasimha},
  journal={New York},
  volume={27},
  number={14},
  year={1993}
}

@article{cachay2025elucidated,
  title={Elucidated Rolling Diffusion Models for Probabilistic Forecasting of Complex Dynamics},
  author={Cachay, Salva R{\"u}hling and Aittala, Miika and Kreis, Karsten and Brenowitz, Noah and Vahdat, Arash and Mardani, Morteza and Yu, Rose},
  journal={arXiv preprint arXiv:2506.20024},
  year={2025}
}

@article{ruhling2023dyffusion,
  title={Dyffusion: A dynamics-informed diffusion model for spatiotemporal forecasting},
  author={R{\"u}hling Cachay, Salva and Zhao, Bo and Joren, Hailey and Yu, Rose},
  journal={Advances in neural information processing systems},
  volume={36},
  pages={45259--45287},
  year={2023}
}

@article{karras2022elucidating,
  title={Elucidating the design space of diffusion-based generative models},
  author={Karras, Tero and Aittala, Miika and Aila, Timo and Laine, Samuli},
  journal={Advances in neural information processing systems},
  volume={35},
  pages={26565--26577},
  year={2022}
}

@article{song2020score,
  title={Score-based generative modeling through stochastic differential equations},
  author={Song, Yang and Sohl-Dickstein, Jascha and Kingma, Diederik P and Kumar, Abhishek and Ermon, Stefano and Poole, Ben},
  journal={arXiv preprint arXiv:2011.13456},
  year={2020}
}

@article{blattmann2023stable,
  title={Stable video diffusion: Scaling latent video diffusion models to large datasets},
  author={Blattmann, Andreas and Dockhorn, Tim and Kulal, Sumith and Mendelevitch, Daniel and Kilian, Maciej and Lorenz, Dominik and Levi, Yam and English, Zion and Voleti, Vikram and Letts, Adam and others},
  journal={arXiv preprint arXiv:2311.15127},
  year={2023}
}

@article{kohl2026benchmarking,
  title={Benchmarking autoregressive conditional diffusion models for turbulent flow simulation},
  author={Kohl, Georg and Chen, Li-Wei and Thuerey, Nils},
  journal={Neural Networks},
  pages={108641},
  year={2026},
  publisher={Elsevier}
}

@article{price2025probabilistic,
  title={Probabilistic weather forecasting with machine learning},
  author={Price, Ilan and Sanchez-Gonzalez, Alvaro and Alet, Ferran and Andersson, Tom R and El-Kadi, Andrew and Masters, Dominic and Ewalds, Timo and Stott, Jacklynn and Mohamed, Shakir and Battaglia, Peter and others},
  journal={Nature},
  volume={637},
  number={8044},
  pages={84--90},
  year={2025},
  publisher={Nature Publishing Group UK London}
}

@article{ruhe2024rolling,
  title={Rolling diffusion models},
  author={Ruhe, David and Heek, Jonathan and Salimans, Tim and Hoogeboom, Emiel},
  journal={arXiv preprint arXiv:2402.09470},
  year={2024}
}

@article{lu2026one,
  title={One-step Latent-free Image Generation with Pixel Mean Flows},
  author={Lu, Yiyang and Lu, Susie and Sun, Qiao and Zhao, Hanhong and Jiang, Zhicheng and Wang, Xianbang and Li, Tianhong and Geng, Zhengyang and He, Kaiming},
  journal={arXiv preprint arXiv:2601.22158},
  year={2026}
}

@article{du2024confild,
  title={Conditional neural field latent diffusion model for generating spatiotemporal turbulence},
  author={Du, Pan and Parikh, Meet Hemant and Fan, Xiantao and Liu, Xin-Yang and Wang, Jian-Xun},
  journal={Nature Communications},
  volume={15},
  number={1},
  pages={10416},
  year={2024},
  publisher={Nature Publishing Group UK London}
}

@article{lim2024elucidating,
  title={Elucidating the design choice of probability paths in flow matching for forecasting},
  author={Lim, Soon Hoe and Wang, Yijin and Yu, Annan and Hart, Emma and Mahoney, Michael W and Li, Xiaoye S and Erichson, N Benjamin},
  journal={arXiv preprint arXiv:2410.03229},
  year={2024}
}

@article{luo2023latent,
  title={Latent consistency models: Synthesizing high-resolution images with few-step inference},
  author={Luo, Simian and Tan, Yiqin and Huang, Longbo and Li, Jian and Zhao, Hang},
  journal={arXiv preprint arXiv:2310.04378},
  year={2023}
}

@inproceedings{mao2025osv,
  title={Osv: One step is enough for high-quality image to video generation},
  author={Mao, Xiaofeng and Jiang, Zhengkai and Wang, Fu-Yun and Zhang, Jiangning and Chen, Hao and Chi, Mingmin and Wang, Yabiao and Luo, Wenhan},
  booktitle={Proceedings of the Computer Vision and Pattern Recognition Conference},
  pages={12585--12594},
  year={2025}
}

@article{wang2023videolcm,
  title={Videolcm: Video latent consistency model},
  author={Wang, Xiang and Zhang, Shiwei and Zhang, Han and Liu, Yu and Zhang, Yingya and Gao, Changxin and Sang, Nong},
  journal={arXiv preprint arXiv:2312.09109},
  year={2023}
}

@article{gao2023prediff,
  title={Prediff: Precipitation nowcasting with latent diffusion models},
  author={Gao, Zhihan and Shi, Xingjian and Han, Boran and Wang, Hao and Jin, Xiaoyong and Maddix, Danielle and Zhu, Yi and Li, Mu and Wang, Yuyang Bernie},
  journal={Advances in Neural Information Processing Systems},
  volume={36},
  pages={78621--78656},
  year={2023}
}

@article{bi2022pangu,
  title={Pangu-weather: A 3d high-resolution model for fast and accurate global weather forecast},
  author={Bi, Kaifeng and Xie, Lingxi and Zhang, Hengheng and Chen, Xin and Gu, Xiaotao and Tian, Qi},
  journal={arXiv preprint arXiv:2211.02556},
  year={2022}
}

@article{von2022self,
  title={A self-attention ansatz for ab-initio quantum chemistry},
  author={von Glehn, Ingrid and Spencer, James S and Pfau, David},
  journal={arXiv preprint arXiv:2211.13672},
  year={2022}
}

@article{wong2018ferminets,
  title={Ferminets: Learning generative machines to generate efficient neural networks via generative synthesis},
  author={Wong, Alexander and Shafiee, Mohammad Javad and Chwyl, Brendan and Li, Francis},
  journal={arXiv preprint arXiv:1809.05989},
  year={2018}
}

@article{luo2025crystalflow,
  title={CrystalFlow: a flow-based generative model for crystalline materials},
  author={Luo, Xiaoshan and Wang, Zhenyu and Wang, Qingchang and Shao, Xuechen and Lv, Jian and Wang, Lei and Wang, Yanchao and Ma, Yanming},
  journal={Nature Communications},
  volume={16},
  number={1},
  pages={9267},
  year={2025},
  publisher={Nature Publishing Group UK London}
}

@article{xue2026uni,
  title={Uni-Flow: a unified autoregressive-diffusion model for complex multiscale flows},
  author={Xue, Xiao and Yang, Tianyue and Gao, Mingyang and Pan, Leyu and Wang, Maida and Zhu, Kewei and Wang, Shuo and Li, Jiuling and ten Eikelder, Marco FP and Coveney, Peter V},
  journal={arXiv preprint arXiv:2602.15592},
  year={2026}
}

@article{molinaro2024generative,
  title={Generative ai for fast and accurate statistical computation of fluids},
  author={Molinaro, Roberto and Lanthaler, Samuel and Raoni{\'c}, Bogdan and Rohner, Tobias and Armegioiu, Victor and Simonis, Stephan and Grund, Dana and Ramic, Yannick and Wan, Zhong Yi and Sha, Fei and others},
  journal={arXiv preprint arXiv:2409.18359},
  year={2024}
}

@article{liu2022rectified,
  title={Rectified flow: A marginal preserving approach to optimal transport},
  author={Liu, Qiang},
  journal={arXiv preprint arXiv:2209.14577},
  year={2022}
}

@misc{li2026basicsletdenoisinggenerative,
      title={Back to Basics: Let Denoising Generative Models Denoise}, 
      author={Tianhong Li and Kaiming He},
      year={2026},
      eprint={2511.13720},
      archivePrefix={arXiv},
      primaryClass={cs.CV},
      url={https://arxiv.org/abs/2511.13720}, 
}

@article{mccabe2025walrus,
  title={Walrus: A cross-domain foundation model for continuum dynamics},
  author={McCabe, Michael and Mukhopadhyay, Payel and Marwah, Tanya and Blancard, Bruno Regaldo-Saint and Rozet, Francois and Diaconu, Cristiana and Meyer, Lucas and Wong, Kaze WK and Sotoudeh, Hadi and Bietti, Alberto and others},
  journal={arXiv preprint arXiv:2511.15684},
  year={2025}
}

@article{liu2024neural,
  title={Neural operators with localized integral and differential kernels},
  author={Liu-Schiaffini, Miguel and Berner, Julius and Bonev, Boris and Kurth, Thorsten and Azizzadenesheli, Kamyar and Anandkumar, Anima},
  journal={arXiv preprint arXiv:2402.16845},
  year={2024}
}

@misc{wu2023maskedtrajectorymodelsprediction,
      title={Masked Trajectory Models for Prediction, Representation, and Control}, 
      author={Philipp Wu and Arjun Majumdar and Kevin Stone and Yixin Lin and Igor Mordatch and Pieter Abbeel and Aravind Rajeswaran},
      year={2023},
      eprint={2305.02968},
      archivePrefix={arXiv},
      primaryClass={cs.LG},
      url={https://arxiv.org/abs/2305.02968}, 
}

@misc{he2021maskedautoencodersscalablevision,
      title={Masked Autoencoders Are Scalable Vision Learners}, 
      author={Kaiming He and Xinlei Chen and Saining Xie and Yanghao Li and Piotr Dollár and Ross Girshick},
      year={2021},
      eprint={2111.06377},
      archivePrefix={arXiv},
      primaryClass={cs.CV},
      url={https://arxiv.org/abs/2111.06377}, 
}

@misc{lin2026decoupleddiffusionsamplinginverse,
      title={Decoupled Diffusion Sampling for Inverse Problems on Function Spaces}, 
      author={Thomas Y. L. Lin and Jiachen Yao and Lufang Chiang and Julius Berner and Anima Anandkumar},
      year={2026},
      eprint={2601.23280},
      archivePrefix={arXiv},
      primaryClass={cs.LG},
      url={https://arxiv.org/abs/2601.23280}, 
}

@misc{yang2026menomeanflowenhancedneuraloperators,
      title={MENO: MeanFlow-Enhanced Neural Operators for Dynamical Systems}, 
      author={Tianyue Yang and Xiao Xue},
      year={2026},
      eprint={2604.06881},
      archivePrefix={arXiv},
      primaryClass={cs.LG},
      url={https://arxiv.org/abs/2604.06881}, 
}

@inproceedings{nichol2021improved,
  title={Improved denoising diffusion probabilistic models},
  author={Nichol, Alexander Quinn and Dhariwal, Prafulla},
  booktitle={International conference on machine learning},
  pages={8162--8171},
  year={2021},
  organization={PMLR}
}

@inproceedings{peebles2023scalable,
  title={Scalable diffusion models with transformers},
  author={Peebles, William and Xie, Saining},
  booktitle={Proceedings of the IEEE/CVF international conference on computer vision},
  pages={4195--4205},
  year={2023}
}

@article{liu2025muon,
  title={Muon is scalable for llm training},
  author={Liu, Jingyuan and Su, Jianlin and Yao, Xingcheng and Jiang, Zhejun and Lai, Guokun and Du, Yulun and Qin, Yidao and Xu, Weixin and Lu, Enzhe and Yan, Junjie and others},
  journal={arXiv preprint arXiv:2502.16982},
  year={2025}
}

@software{flax2020github,
  author = {Jonathan Heek and Anselm Levskaya and Avital Oliver and Marvin Ritter and Bertrand Rondepierre and Andreas Steiner and Marc van {Z}ee},
  title = {{F}lax: A neural network library and ecosystem for {JAX}},
  url = {http://github.com/google/flax},
  version = {0.12.6},
  year = {2024},
}

@article{kingma2014adam,
  title={Adam: A method for stochastic optimization},
  author={Kingma, Diederik P and Ba, Jimmy},
  journal={arXiv preprint arXiv:1412.6980},
  year={2014}
}

@inproceedings{gupta2018shampoo,
  title={Shampoo: Preconditioned stochastic tensor optimization},
  author={Gupta, Vineet and Koren, Tomer and Singer, Yoram},
  booktitle={International Conference on Machine Learning},
  pages={1842--1850},
  year={2018},
  organization={PMLR}
}

@article{wu2023ar,
  title={Ar-diffusion: Auto-regressive diffusion model for text generation},
  author={Wu, Tong and Fan, Zhihao and Liu, Xiao and Zheng, Hai-Tao and Gong, Yeyun and Jiao, Jian and Li, Juntao and Guo, Jian and Duan, Nan and Chen, Weizhu and others},
  journal={Advances in Neural Information Processing Systems},
  volume={36},
  pages={39957--39974},
  year={2023}
}

@article{yang2023diffusion,
  title={Diffusion probabilistic modeling for video generation},
  author={Yang, Ruihan and Srivastava, Prakhar and Mandt, Stephan},
  journal={Entropy},
  volume={25},
  number={10},
  pages={1469},
  year={2023},
  publisher={MDPI}
}

@article{kim2024fifo,
  title={Fifo-diffusion: Generating infinite videos from text without training},
  author={Kim, Jihwan and Kang, Junoh and Choi, Jinyoung and Han, Bohyung},
  journal={Advances in Neural Information Processing Systems},
  volume={37},
  pages={89834--89868},
  year={2024}
}

@inproceedings{chen2023seine,
  title={Seine: Short-to-long video diffusion model for generative transition and prediction},
  author={Chen, Xinyuan and Wang, Yaohui and Zhang, Lingjun and Zhuang, Shaobin and Ma, Xin and Yu, Jiashuo and Wang, Yali and Lin, Dahua and Qiao, Yu and Liu, Ziwei},
  booktitle={The Twelfth International Conference on Learning Representations},
  year={2023}
}

@article{ergasti2025r,
  title={$^{R}$FLAV: Rolling Flow matching for infinite Audio Video generation},
  author={Ergasti, Alex and Tarollo, Giuseppe Gabriele and Botti, Filippo and Fontanini, Tomaso and Ferrari, Claudio and Bertozzi, Massimo and Prati, Andrea},
  journal={arXiv preprint arXiv:2503.08307},
  year={2025}
}

@article{elgazzar2025probabilistic,
  title={Probabilistic forecasting via autoregressive flow matching},
  author={ElGazzar, Ahmed and van Gerven, Marcel},
  journal={arXiv preprint arXiv:2503.10375},
  year={2025}
}

@article{armegioiu2025rectified,
  title={Rectified flows for fast multiscale fluid flow modeling},
  author={Armegioiu, Victor and Ramic, Yannick and Mishra, Siddhartha},
  journal={arXiv preprint arXiv:2506.03111},
  year={2025}
}

@inproceedings{wang2021deep,
  title={Deep generative learning via schr{\"o}dinger bridge},
  author={Wang, Gefei and Jiao, Yuling and Xu, Qian and Wang, Yang and Yang, Can},
  booktitle={International conference on machine learning},
  pages={10794--10804},
  year={2021},
  organization={PMLR}
}

@article{lin2025design,
  title={On the Design of One-step Diffusion via Shortcutting Flow Paths},
  author={Lin, Haitao and Hu, Peiyan and Ren, Minsi and Gao, Zhifeng and Ma, Zhi-Ming and Wu, Tailin and Li, Stan Z and others},
  journal={arXiv preprint arXiv:2512.11831},
  year={2025}
}

@article{zhang2025alphaflow,
  title={Alphaflow: Understanding and improving meanflow models},
  author={Zhang, Huijie and Siarohin, Aliaksandr and Menapace, Willi and Vasilkovsky, Michael and Tulyakov, Sergey and Qu, Qing and Skorokhodov, Ivan},
  journal={arXiv preprint arXiv:2510.20771},
  year={2025}
}

@article{zwanzig1961memory,
  title={Memory effects in irreversible thermodynamics},
  author={Zwanzig, Robert},
  journal={Physical Review},
  volume={124},
  number={4},
  pages={983},
  year={1961},
  publisher={APS}
}

@article{chen1998lattice,
  title={Lattice Boltzmann method for fluid flows},
  author={Chen, Shiyi and Doolen, Gary D},
  journal={Annual review of fluid mechanics},
  volume={30},
  number={1},
  pages={329--364},
  year={1998},
  publisher={Annual Reviews 4139 El Camino Way, PO Box 10139, Palo Alto, CA 94303-0139, USA}
}

@inproceedings{assran2023self,
  title={Self-supervised learning from images with a joint-embedding predictive architecture},
  author={Assran, Mahmoud and Duval, Quentin and Misra, Ishan and Bojanowski, Piotr and Vincent, Pascal and Rabbat, Michael and LeCun, Yann and Ballas, Nicolas},
  booktitle={Proceedings of the IEEE/CVF conference on computer vision and pattern recognition},
  pages={15619--15629},
  year={2023}
}

@article{mccabe2023towards,
  title={Towards stability of autoregressive neural operators},
  author={McCabe, Michael and Harrington, Peter and Subramanian, Shashank and Brown, Jed},
  journal={arXiv preprint arXiv:2306.10619},
  year={2023}
}

@inproceedings{hoogeboom2023simple,
  title={simple diffusion: End-to-end diffusion for high resolution images},
  author={Hoogeboom, Emiel and Heek, Jonathan and Salimans, Tim},
  booktitle={International Conference on Machine Learning},
  pages={13213--13232},
  year={2023},
  organization={PMLR}
}
\clearpage

%%%%%%%%%%%%%%%%%%%%%%%%%%%%%%%%%%%%%%%%%%%%%%%%%%%%%%%%%%%%

\appendix

\startcontents[appendix]
\section*{Appendix}
\printcontents[appendix]{}{1}{\setcounter{tocdepth}{2}}
\clearpage

\section{Hyperparameters}
\subsection{MeLISA Settings}\label{appendix:training_setting}
We list the backbone architecture hyperparameters for MeLISA in Table~\ref{tab:melisa_table}. Here, \texttt{Depth} refers to the number of spatial downsampling stages in the UNet backbone, and to the number of transformer layers in the DiT backbone. Our UNet implementation uses the \texttt{flax.nnx} API (\texttt{flax}~\cite{flax2020github} is a deep learning library for JAX) and is adapted from the \texttt{flax.linen} DDPM-style UNet of \citet{song2020score}. Our DiT implementation also uses \texttt{flax.nnx} and follows \citet{lu2026one}, which builds on the design of \citet{li2026basicsletdenoisinggenerative}. In both backbones, we do not introduce separate temporal attention; instead, all input frames are processed jointly as a single large image.

\begin{table}[!h]
\centering
\caption{Architecture hyperparameter setting for MeLISA backbone models.}
\setlength{\tabcolsep}{0.4em}
\begin{tabular}{lccccc}
\toprule
\textbf{Model} & Attention Resolution  & Depth & Residual Layers & Width & Patch Size \\
\midrule
\rowcolor{gray!15} \multicolumn{6}{l}{\textit{UNet-based Variants}}\\
MeLISA-$\Upsilon$-XS & - & 4 & 1 & 32 & - \\
MeLISA-$\Upsilon$-S & 32 & 4 & 2 & 32 & - \\
\rowcolor{gray!15} \multicolumn{6}{l}{\textit{DiT-based Variants}}\\
MeLISA-$\Delta$-S & - & 16  & - & 240 & 16 \\
MeLISA-$\Delta$-M & - & 24 & - & 480 & 16 \\
MeLISA-$\Delta$-B  & - & 28 & - & 720 & 16 \\
\bottomrule
\end{tabular}
\label{tab:melisa_table}
\end{table}

The model hyperparameters for MeLISA are listed in Table~\ref{tab:melisa_setting}. The masking rate is chosen such that, on average, two frames are revealed during training, including the first frame, which is always retained as a temporal reference. The TIC lag weights increase with lag, thereby penalizing error accumulation over temporal evolution more strongly at longer horizons. Our sampling strategy for the cases $t \neq r$, corresponding to the flow matching limit of MeanFlow models~\cite{geng2025mean,geng2025improved,lu2026one}, as well as the joint sampler for $t$ and $r$, is directly adapted from \citet{lu2026one}.

\begin{table}[!h]
\centering
\caption{MeLISA model setting.}
\setlength{\tabcolsep}{0.3em}
\begin{tabular}{lcccc}
\toprule
\textbf{MeLISA Settings} & Masking Rate & TIC lag weighting & Ratio of $t\neq r$ & $t, \ r$ Sampler \\
\midrule
  \textbf{Values} & 0.8 & (0.4, 0.5, 0.8, 1.1, 1.2) & 0.5 & logit-normal(0.8, 0.8)  \\
\bottomrule
\end{tabular}
\label{tab:melisa_setting}
\end{table}

The optimization hyperparameters are listed in Table~\ref{tab:melisa_optim}. For the UNet-based variants, we use a linearly decaying learning-rate schedule, which we found helpful for learning fine-scale details. For the DiT-based variants, we follow \citet{lu2026one} and use the Muon optimizer~\cite{liu2025muon}, a second-order method based on the Newton--Schulz iteration and developed for large-scale transformer training as an improvement over Shampoo~\cite{gupta2018shampoo}. Its learning rate is therefore \textbf{not} directly comparable to that of Adam. For larger models, we adopt cosine annealing schedules to reduce overfitting. We also decrease the number of training epochs as model size increases in order to approximately match training cost across models. The training random seed is fixed at 42 for all models. 

\begin{table}[!h]
\centering
\caption{Optimization hyperparameter setting for MeLISA backbone models.}
\setlength{\tabcolsep}{0.4em}
\begin{tabular}{lccccc}
\toprule
\textbf{Model} & Optimizer & Learning Rate & Schedule & Epochs & Batch Size \\
\midrule
\rowcolor{gray!15} \multicolumn{6}{l}{\textit{UNet-based Variants}}\\
MeLISA-$\Upsilon$-XS & Adam~\cite{kingma2014adam} & 1e-4 & Linear Decay & 500 & 256 \\
MeLISA-$\Upsilon$-S & Adam~\cite{kingma2014adam} & 1e-4 & Linear Decay & 500 & 256 \\
\rowcolor{gray!15} \multicolumn{6}{l}{\textit{DiT-based Variants}}\\
MeLISA-$\Delta$-S & Muon~\cite{liu2025muon} & 5e-3 & Constant & 300 & 256 \\
MeLISA-$\Delta$-M & Muon~\cite{liu2025muon} & 5e-3 & Cosine Annealing & 200 & 256 \\
MeLISA-$\Delta$-B & Muon~\cite{liu2025muon} & 5e-3 & Cosine Annealing & 150 & 256 \\
\bottomrule
\end{tabular}
\label{tab:melisa_optim}
\end{table}

\subsection{Baseline Settings}

For the neural operator baselines, we follow \citet{li2020fourier} and retain as many Fourier modes as feasible to improve spectral fidelity. We keep the model depth fixed across architectures and match overall model capacity as closely as possible by adjusting the channel expansion factor. All baselines are trained with two input frames and four output frames, consistent with the MeLISA setting. The architecture hyperparameters are summarized in Table~\ref{tab:no_table}.

\begin{table}[!h]
\centering
\caption{Architecture hyperparameter setting for neural operator models.}
\setlength{\tabcolsep}{0.3em}
\begin{tabular}{lccccc}
\toprule
\textbf{Model} & Fourier Modes & Channel MLP Expansion & Depth & Width & Factorization \\
\midrule
FNO       & 42 & 0.5 & 7 & 32 & Tucker \\
UNO       & 32,32,32,16,32,32,32 & 2 & 7 & 32 & Tucker \\
Local-FNO & 42 & 0.5 & 7 & 32 & Tucker \\
\bottomrule
\end{tabular}
\label{tab:no_table}
\end{table}

The optimization settings for the neural operator baselines are listed in Table~\ref{tab:no_optim}. These settings are kept identical across all models. We adopt the learning rate configuration of \citet{li2020fourier}, replacing the piecewise-constant decay schedule with a smooth linear decay.

\begin{table}[!h]
\centering
\caption{Optimization hyperparameter setting for neural operator models.}
\setlength{\tabcolsep}{0.3em}
\begin{tabular}{lccccc}
\toprule
\textbf{Model} & Optimizer & Learning Rate & Schedule & Epochs & Batch Size \\
\midrule
FNO       & Adam & 1e-3 & Linear Decay & 1000 & 128 \\
UNO       & Adam & 1e-3 & Linear Decay & 1000 & 128 \\
Local-FNO & Adam & 1e-3 & Linear Decay & 1000 & 128 \\
\bottomrule
\end{tabular}
\label{tab:no_optim}
\end{table}
\clearpage

\subsection{MeLISA Training}
\begin{algorithm}[h]
\caption{MeLISA Model Training}
\label{alg:training}
\begin{algorithmic}[1]
\REQUIRE Ordered dataset $\mathcal{D}$ (empirical distribution $\hat p_{\mathcal D}$), denoiser $D_\theta$, window length $W$, iterations $K$, batch size $B$, learning rate $\eta$, masking rate $\upsilon$, lag weights $\{\kappa_w\}_{w=1}^{W-1}$
\STATE Initialize parameters $\theta$ of network $D_\theta$
\FOR{$k=1,\dots,K$}
    \STATE Sample a batch of windows $\{\bar{x}_i^{\tau_i:\tau_i+W}\}_{i=1}^B \sim \hat p_{\mathcal D}$
    \STATE Sample $\{\epsilon_i\}_{i=1}^B \sim \mathcal N(0,I)$, diffusion times $\{t_i\}_{i=1}^B$, reference times $\{r_i\}_{i=1}^B$
    \STATE Sample masks $\{\tau_{M,i}\}_{i=1}^B$ over later frames from $\mathrm{Bernoulli}(\upsilon)$, always retaining the first frame
    \FOR{$i=1,\dots,B$}
        \STATE $\bar{z}_{t_i}^{\tau_i:\tau_i+W}\gets (1-t_i)\bar{x}_i^{\tau_i:\tau_i+W}+t_i\epsilon_i$
        \STATE $\bar{v}_i^{\tau_i:\tau_i+W}\gets \epsilon_i-\bar{x}_i^{\tau_i:\tau_i+W}$
        \STATE $\bar{u}_{\theta,i}\gets \dfrac{1}{t_i}\left[\bar{z}_{t_i}^{\tau_i:\tau_i+W}-D_\theta\!\left( \mathrm{Concat} \left[\bar{z}_{t_i}^{\tau_i:\tau_i+W}, \ \mathcal{M}_{\tau_{M,i}}\!\left(\bar{x}_i^{\tau_i:\tau_i+W}\right), \tau_{M, i}\right],r_i,t_i\right)\right]$
        \STATE $\bar{V}_{\theta,i}\gets \bar{u}_{\theta,i}+(t_i-r_i)\cdot \operatorname{sg}\!\left(\partial_t \bar{u}_{\theta,i}+{\bar{v}_i^{\tau_i:\tau_i+W}} \ ^{\top}\nabla_{\bar z}\bar{u}_{\theta,i}\right)$
        \STATE $\hat{\bar{x}}_i^{\tau_i:\tau_i+W}\gets D_\theta\!\left( \mathrm{Concat}\left[\epsilon_i, \mathcal{M}_{\tau_{M,i}}\!\left(\bar{x}_i^{\tau_i:\tau_i+W}\right), \tau_{M, i} \right],0,1\right)$
    \ENDFOR
    \STATE $\mathcal{L}_{\mathrm{WinC\text{-}MF}}\gets \dfrac{1}{B}\sum_{i=1}^B \left\|\bar{V}_{\theta,i}-\bar{v}_i^{\tau_i:\tau_i+W}\right\|_2^2$
    \STATE $\mathcal{L}_{\mathrm{TIC}}\gets \dfrac{1}{B}\sum_{i=1}^B \sum_{w=1}^{W-1}\kappa_w
    \left\|
    \bigl(x_i^{\tau_i+w}-x_i^{\tau_i}\bigr)
    -
    \bigl(\hat{x}_i^{\tau_i+w}-\hat{x}_i^{\tau_i}\bigr)
    \right\|_2^2$
    \STATE $\mathcal{L}_{\mathrm{MeLISA}}\gets \mathcal{L}_{\mathrm{WinC\text{-}MF}}+\mathcal{L}_{\mathrm{TIC}}$
    \STATE $\theta \gets \theta-\eta \nabla_\theta \mathcal{L}_{\mathrm{MeLISA}}$
\ENDFOR
\end{algorithmic}
\end{algorithm}

\section{Overview of Autoregressive Diffusion Models}\label{appendix:overview}
We provide a conceptual and theoretical overview of diffusion-based autoregressive modeling for physical dynamical systems in this section, and discuss the position of MeLISA models.

\subsection{Diffusion Autoregressive Models}
\paragraph{Rolling diffusion.} One major line of work extends \textbf{video diffusion models}~\cite{blattmann2023stable,yang2023diffusion,kim2024fifo,chen2023seine} to temporal forecasting, typically by applying a window-based noise schedule. A representative early framework is the \textbf{rolling sequence diffusion model}~\cite{wu2023ar,ruhe2024rolling}, which, like MeLISA, operates on sampled windows $\bar{x}_i^{\tau_i:\tau_i+W}$. These methods define a progressive noise schedule $\bar{\sigma}(t_i;\sigma_{\min},\sigma_{\max},\rho)$, parameterized by the minimum noise level $\sigma_{\min}$, maximum noise level $\sigma_{\max}$, and curvature parameter $\rho$. Traditional DDPM-based variants, such as rolling diffusion models~\cite{ruhe2024rolling}, typically use either constant-noise or variance-preserving schedules; see \citet{song2020score} for details. More recent methods, such as elucidated rolling diffusion models (ERDM)~\cite{cachay2025elucidated}, adopt EDM-inspired schedules~\cite{karras2022elucidating} with explicit curvature control. These approaches are generally framed as \textbf{probabilistic forecasting models}, whose main advantage over deterministic autoregressive methods, such as neural operators, is their ability to represent predictive uncertainty. The general training procedure is summarized in Algorithm~\ref{alg:rolling_diffusion}. Because inference in these models typically requires tens to hundreds of diffusion steps, their \textbf{sampling speed is substantially slower than that of deterministic models}.

Theoretically, following the score-based SDE formulation of diffusion models~\cite{song2020score}, rolling diffusion models can be interpreted as a physical-time-dependent extension of probabilistic diffusion. For a window-dependent variance schedule, the time derivative of the covariance matrix is
\begin{equation}
\dot{\mathbf{G}}_{\mathrm{window}}(t)=\operatorname{diag}\!\bigl(
2\bar{\sigma}_1(t)\dot{\bar{\sigma}}_1(t)\mathbf{I}_D,
\ldots,
2\bar{\sigma}_W(t)\dot{\bar{\sigma}}_W(t)\mathbf{I}_D
\bigr).
\end{equation}
Note that $\dot{\mathbf{G}}_{\mathrm{window}}(t)$ is block-diagonal and depends on the position of each frame within the rollout window. The corresponding forward diffusion SDE can be written as
\begin{equation}
\mathrm{d}\mathbf{x}_t =
\mathbf{K}_{\mathrm{window}}(t)\,\mathrm{d}\omega(t),
\end{equation}
where $\mathbf{K}_{\mathrm{window}}(t)$ is any matrix square root satisfying
\begin{equation}
\mathbf{K}_{\mathrm{window}}(t)\mathbf{K}_{\mathrm{window}}(t)^\top
=
\dot{\mathbf{G}}_{\mathrm{window}}(t).
\end{equation}
Under standard regularity assumptions, the associated reverse-time SDE is
\begin{equation}
\mathrm{d}\mathbf{x}_t
=
-\dot{\mathbf{G}}_{\mathrm{window}}(t)\,
\nabla_{\mathbf{x}}
\log p_t(\mathbf{x}_t)\,\mathrm{d}t
+
\mathbf{K}_{\mathrm{window}}(t)\,\mathrm{d}\bar{\omega}(t),
\end{equation}
and the corresponding probability-flow ODE is
\begin{equation}
\frac{\mathrm{d}\mathbf{x}_t}{\mathrm{d}t}
=
-\frac{1}{2}\,
\dot{\mathbf{G}}_{\mathrm{window}}(t)\,
\nabla_{\mathbf{x}}
\log p_t(\mathbf{x}_t).
\end{equation}
In practice, the score is approximated by a denoiser network through the noisy conditional distribution induced by the rolling schedule. Hence, inference in rolling diffusion models amounts to numerically integrating either the reverse-time SDE or its probability-flow ODE counterpart while repeatedly advancing the physical rollout window.

\begin{algorithm}[t]
\caption{General Rolling Diffusion Model Training}
\label{alg:rolling_diffusion}
\begin{algorithmic}[1]
\REQUIRE Ordered dataset $\mathcal{D}$ (empirical distribution $\hat p_{\mathcal{D}}$), denoiser $D_\theta$, window length $W$, iterations $K$, batch size $B$, learning rate $\eta$, noise schedule parameters $\sigma_{\min}, \sigma_{\max}, \rho$, preconditioning parameters $P_{\mathrm{mean}}, P_{\mathrm{std}}$
\STATE Initialize parameters $\theta$ of network $D_\theta$
\FOR{$k = 1,\ldots,K$}
    \STATE Sample a batch of windows $\left\{\bar{x}_i^{\tau_i:\tau_i+W}\right\}_{i=1}^B \sim \hat p_{\mathcal{D}}$
    \STATE Sample $\left\{\epsilon_i\right\}_{i=1}^B \sim \mathcal{N}(0, I_{W \times D})$ and noise times $\left\{t_i\right\}_{i=1}^B \sim \mathcal{U}([0,1))$
    \FOR{$i = 1,\ldots,B$}
        \STATE $\boldsymbol{\sigma}_i = (\sigma_{i,1},\ldots,\sigma_{i,W}) \leftarrow \bar{\sigma}(t_i;\sigma_{\min},\sigma_{\max},\rho)$
        \STATE $\bar{z}_i^{\tau_i:\tau_i+W} \leftarrow \bar{x}_i^{\tau_i:\tau_i+W} + \boldsymbol{\sigma}_i \odot \epsilon_i$
        \STATE $\hat{x}_i^{\tau_i:\tau_i+W} \leftarrow c_{\mathrm{skip}}(\boldsymbol{\sigma}_i)\odot \bar{z}_i^{\tau_i:\tau_i+W} + c_{\mathrm{out}}(\boldsymbol{\sigma}_i)\odot D_\theta\!\left(c_{\mathrm{in}}(\boldsymbol{\sigma}_i)\odot \bar{z}_i^{\tau_i:\tau_i+W},\, c_{\mathrm{noise}}(\boldsymbol{\sigma}_i)\right)$
    \ENDFOR
    \STATE $\mathcal{L}_{\mathrm{RD}} \leftarrow \frac{1}{B}\sum_{i=1}^B \frac{1}{W}\sum_{w=1}^W \lambda(\sigma_{i,w})\,f(\sigma_{i,w};P_{\mathrm{mean}},P_{\mathrm{std}})\,\left\|\left(\bar{x}_i^{\tau_i:\tau_i+W}\right)_w - \left(\hat{x}_i^{\tau_i:\tau_i+W}\right)_w\right\|_2^2$
    \STATE $\theta \leftarrow \theta - \eta \nabla_\theta \mathcal{L}_{\mathrm{RD}}$
\ENDFOR
\end{algorithmic}
\end{algorithm}

This viewpoint clarifies both the strength and limitation of rolling diffusion models. Their main strength is that they provide a principled probabilistic formulation of uncertainty propagation over multiple future frames. However, because generation is tied to reverse-time integration, they are inherently \emph{solver-based}: each forecast block requires multiple denoising evaluations across diffusion time, followed by a rolling update of the window. As a result, their inference procedure is not naturally compatible with a one-step autoregressive schema.

\paragraph{Autoregressive transition kernel.} A complementary theoretical perspective is to view autoregressive generative forecasting directly through a \emph{transition kernel}.  Let $\bar{c}$ denote the observed context and let $\hat{\bar{X}}_{\mathrm{blk}}$ denote a future block to be generated. Then an autoregressive generative surrogate may be written abstractly as
\begin{equation}
\hat{\bar{X}}_{\mathrm{blk}}
\sim
p_\theta\!\left(
\hat{\bar{X}}_{\mathrm{blk}}
\mid
\bar{c}
\right).
\end{equation}

This approach has been validated in both pixel-space~\cite{molinaro2024generative, ruhling2023dyffusion}, and VAE-based latent space~\cite{gao2023prediff}. In these existing autoregressive diffusion formulations, this kernel is instantiated in a relatively restrictive way: the model is conditioned on a fixed, short temporal context and predicts either a single next snapshot or a fixed future block under a pre-specified input--output geometry. Such formulations are probabilistically sound, but they do not always fully exploit the structure of a partially observed spatiotemporal window, and they offer limited flexibility with respect to varying context length or forecast horizon.

\subsection{Flow Matching Autoregressive Models}

\paragraph{Rolling flow matching models.}
Analogous to rolling diffusion models, window-based rolling flow matching models have recently been explored in video generation~\cite{ergasti2025r}, although the literature remains much sparser than that for diffusion-based methods. Simple rolling autoregressive schemes have also been studied for forecasting physical dynamical systems~\cite{elgazzar2025probabilistic}. However, to the best of our knowledge, there is still no dedicated frame-dependent noise schedule for flow-matching models comparable to the schedules used in diffusion-based approaches such as ERDM~\cite{cachay2025elucidated}.

\paragraph{Flow matching transition kernel models.}
Transition kernel formulations have received greater attention in the flow-matching literature. Recent work has employed rectified flows~\cite{armegioiu2025rectified} for spatiotemporal prediction, while \citet{lim2024elucidating} use flow matching-based Schr\"odinger bridge models~\cite{wang2021deep} to learn a transition kernel in latent space. 

\subsection{MeanFlow Model Family}

Before discussing the role and advantages of MeLISA, we briefly review the development of MeanFlow models. The original MeanFlow (o-MF) model~\cite{geng2025mean} was introduced as a one-step alternative to standard optimal-transport flow matching (OT-FM). OT-FM defines the interpolation path
\begin{equation}
    z_t = (1-t)x + t\epsilon,
\end{equation}
where $\epsilon \sim \mathcal{N}(0,\mathbf{I})$. MeanFlow models the time-averaged velocity between diffusion time $t$ and a reference time $r$:
\begin{equation}\label{eq:mf_int}
    u(z_t,r,t)
    =
    \frac{1}{t-r}
    \int_r^t
    v(z_\tau \mid x,\epsilon)\,d\tau,
\end{equation}
which enables single-step transport from $z_t$ to $z_r$ via
\(
z_r = z_t - (t-r)u(z_t,r,t)
\).
Differentiating Eq.~\ref{eq:mf_int} yields the \textbf{MeanFlow identity}
\begin{equation}\label{eq:meanflow_identity}
    u(z_t,r,t)
    =
    v(z_t \mid x,\epsilon)
    -
    (t-r)\frac{d}{dt}u(z_t,r,t),
\end{equation}
which leads to the o-MF objective
\begin{equation}\label{eq:meanflow-loss}
    \mathcal{L}_{\mathrm{o\text{-}MF}}
    =
    \left\|
        u_\theta(z_t,r,t)
        -
        \operatorname{sg}(u_{\mathrm{tgt}})
    \right\|^2.
\end{equation}
This can be interpreted as a \textbf{self-supervised loss}~\cite{lin2025design} defined over a velocity interval. However, o-MF has been reported to exhibit unstable training dynamics, including the phenomenon that the training loss may increase even as generation quality improves~\cite{geng2025improved,zhang2025alphaflow}. This behavior is largely attributed to samples near the flow-matching limit $t=r$, where MeanFlow degenerates to standard OT-FM. $\alpha$-Flow~\cite{zhang2025alphaflow} addresses this issue by introducing an annealing schedule that smoothly recovers the flow-matching objective. A more direct remedy was proposed by \citet{geng2025improved}, who introduced improved MeanFlow (i-MF). Instead of supervising the interval-averaged velocity directly, i-MF uses Eq.~\ref{eq:meanflow_identity} to construct an estimate of the instantaneous velocity:
\begin{equation}
    V_\theta(z_t,r,t)
    \equiv
    u_\theta
    +
    (t-r)\,\operatorname{sg}\!\left(
        \partial_t u_\theta
        +
        v(z_t \mid x,\epsilon)\,\partial_{z_t}u_\theta
    \right),
\end{equation}
and optimizes the objective
\begin{equation}
    \mathcal{L}_{\mathrm{i\text{-}MF}}
    =
    \mathbb{E}_{t,x,\epsilon}
    \left[
        \left\|
            V_\theta(z_t,r,t)
            -
            v(z_t \mid x,\epsilon)
        \right\|^2
    \right].
\end{equation}

Both o-MF and i-MF were originally developed in VAE-induced latent space for $256 \times 256$ image generation. To remove the VAE bottleneck, \citet{li2026basicsletdenoisinggenerative} proposed the Just-image Transformer (JiT), enabling pixel-space image generation up to $1024 \times 1024$. Building on this idea, \citet{lu2026one} introduced pixel MeanFlow (p-MF), which reparameterizes the prediction target in pixel space as
\begin{equation}
    X_\theta(z_t,r,t)
    =
    z_t - t\,u(z_t,r,t),
    \qquad
    u(z_t,r,t)
    =
    \frac{1}{t}\bigl[z_t - X_\theta(z_t,r,t)\bigr].
\end{equation}
The induced $u$ is then used within the i-MF objective. In physical dynamical-system forecasting, many diffusion-based methods operate either in VAE latent space~\cite{gao2023prediff,lim2024elucidating} or at relatively low pixel resolutions (typically up to $128 \times 128$). To the best of our knowledge, p-MF is the only one-step diffusion-based formulation that has been validated in high-dimensional pixel space. MeLISA is therefore built on top of p-MF.

\subsection{Birth of MeLISA: The One-Step Dilemma}

We now position MeLISA relative to existing diffusion-based forecasting paradigms. From the perspective of rolling diffusion, one-step generation appears at odds with the core idea of iterative denoising: to fully exploit one-step efficiency, all future frames would ideally be generated in a single network evaluation. From the transition-kernel perspective, however, we still seek to preserve the richer temporal conditioning provided by window-based generative models.

MeLISA occupies an intermediate point between these two views. On the one hand, it remains a probabilistic autoregressive model, since stochasticity enters through Gaussian noise injected into the masked future portion of the input window. On the other hand, unlike rolling SDE-based models, MeLISA does not require multi-step reverse-time integration. Its sampling procedure, summarized in Algorithm~\ref{alg:inference}, takes a context buffer $\bar{c}$ and rollout block size $S = W - W_{\mathrm{ctx}}$, constructs
\begin{equation}
    \tilde{\bar{x}}
    \gets
    \mathrm{Concat}\!\left[
        \epsilon,\;
        \mathcal{M}_{\mathrm{fut}}
        \!\left(
            [\bar{c};\,\mathbf{0}_{S \times F}]
        \right),\;
        \tau_M
    \right],
\end{equation}
and applies a single denoising map
\begin{equation}
    \hat{\bar{x}}
    =
    D_\theta(\tilde{\bar{x}},0,1),
    \qquad
    \hat{\bar{X}}_{\mathrm{blk}}
    =
    \operatorname{Tail}_S(\hat{\bar{x}}).
\end{equation}
This defines the stochastic transition kernel
\begin{equation}
    p_\theta\!\left(
        \hat{\bar{X}}_{\mathrm{blk}}
        \mid
        \bar{c}
    \right)
    =
    \int
    \delta\!\Bigl(
        \hat{\bar{X}}_{\mathrm{blk}}
        -
        \operatorname{Tail}_S\!\bigl(
            D_\theta(\tilde{\bar{x}}(\epsilon),0,1)
        \bigr)
    \Bigr)\,
    p(\epsilon)\,d\epsilon,
\end{equation}
where $\delta(\cdot)$ denotes the Dirac delta distribution. Hence, MeLISA preserves the probabilistic transition-kernel interpretation of autoregressive generation, but defines the kernel over a \emph{partially observed spatiotemporal window} rather than a single-step state transition.

This perspective yields two key advantages. First, compared with standard autoregressive diffusion kernels, MeLISA conditions on a richer object (a masked temporal window), which allows it to exploit multi-frame structure more effectively. Second, compared with SDE-based rolling diffusion models such as ERDM, MeLISA supports one-step inference: each future block is generated with a single model evaluation instead of iterative reverse-time integration. In this sense, MeLISA combines the probabilistic expressivity of stochastic autoregressive modeling with the efficiency and flexibility of direct window-conditioned generation.

This interpretation also explains MeLISA's practical flexibility at inference time. Since the model conditions only on the most recent $W_{\mathrm{ctx}}$ frames and predicts blocks of size $S$, it naturally supports arbitrary observed input lengths $W_{\mathrm{in}} \geq W_{\mathrm{ctx}}$ and arbitrary forecast horizons $W_{\mathrm{out}}$ via repeated blockwise rollout. Unlike classical rolling diffusion models, whose formulations are tightly coupled to diffusion-time integration, MeLISA decouples \emph{probabilistic forecasting} from \emph{multi-step SDE solving}. This yields a one-step stochastic autoregressive model that is particularly well suited for long-horizon dynamical prediction. Table~\ref{tab:model_comparison} compares MeLISA with the models discussed in this section. We note that, due to the novelty of our formulation, MeLISA has \textbf{no strict counterpart} that can serve as a direct benchmark. 

\begin{table}[t]
  \centering
  \small
  \setlength{\tabcolsep}{2.2pt}
  \renewcommand{\arraystretch}{1.0}
  \caption{
  Model-level comparison between MeLISA and existing diffusion-based generative models.
  Here, DA denotes distribution-aware modeling, whose methods primarily focus on matching generated and target distributions and are typically not designed for autoregressive rollouts; hence, they are not directly comparable to MeLISA.
  TK denotes transition-kernel modeling.
  CoNFiLD performs long-horizon prediction in a non-autoregressive manner by predicting all latent variables in a single diffusion batch.
  In contrast, MeLISA is 1-NFE, latent-free, and does not require a specialized backbone.
  Under this comparison, MeLISA has no strict counterpart among existing methods.
  }
  \label{tab:model_comparison}
  \begin{tabular}{l|cccccc}
    \toprule
    \textbf{Model Name} 
    & \textbf{Model Type} 
    & \textbf{Max. Resolution} 
    & \textbf{Formalism} 
    & \textbf{1-NFE?} 
    & \textbf{Special Backbone?} 
    & \textbf{Latent-Free?} \\
    \midrule

    \rowcolor{gray!15}
    \multicolumn{7}{l}{\textit{Existing Models}} \\
    
    DYffusion~\cite{ruhling2023dyffusion} 
    & AR 
    & $221 \times 42$ 
    & Rolling 
    & \cellcolor{red!15}\ding{53} 
    & \cellcolor{green!15} No -- UNet~\cite{ruhling2023dyffusion} 
    & \cellcolor{green!15} Yes \\

    ERDM~\cite{cachay2025elucidated} 
    & AR 
    & $240 \times 121$ 
    & Rolling 
    & \cellcolor{red!15}\ding{53}  
    & \cellcolor{red!15} Yes 
    & \cellcolor{green!15} Yes \\

    RDM~\cite{ruhe2024rolling} 
    & AR 
    & $64 \times 64$ 
    & Rolling 
    & \cellcolor{red!15}\ding{53} 
    & \cellcolor{green!15} No -- UViT~\cite{hoogeboom2023simple} 
    & \cellcolor{green!15} Yes \\

    GenCFD~\cite{molinaro2024generative} 
    & \cellcolor{gray!15}\textcolor{gray}{DA} 
    & \cellcolor{gray!15}\textcolor{gray}{$128^3$} 
    & \cellcolor{gray!15}\textcolor{gray}{TK} 
    & \cellcolor{gray!15}\textcolor{gray}{\ding{53}} 
    & \cellcolor{gray!15}\textcolor{gray}{No -- UViT~\cite{molinaro2024generative}} 
    & \cellcolor{gray!15}\textcolor{gray}{Yes} \\

    Rectified Flow~\cite{armegioiu2025rectified} 
    & \cellcolor{gray!15}\textcolor{gray}{DA}
    & \cellcolor{gray!15}\textcolor{gray}{$128 \times 128$} 
    & \cellcolor{gray!15}\textcolor{gray}{TK} 
    & \cellcolor{gray!15}\textcolor{gray}{\checkmark} 
    & \cellcolor{gray!15}\textcolor{gray}{No -- UNet} 
    & \cellcolor{gray!15}\textcolor{gray}{Yes} \\

    CoNFiLD~\cite{du2024confild}  
    & Non-AR 
    & $100 \times 400$ 
    & TK 
    & \cellcolor{red!15}\ding{53} 
    & \cellcolor{red!15} Yes 
    & \cellcolor{red!15} No (VAE) \\

    \addlinespace[2pt]

    \rowcolor{gray!15}
    \multicolumn{7}{l}{\textit{MeLISA Models}} \\

    MeLISA-$\Delta$ 
    & AR 
    & $256 \times 256$ 
    & Mix 
    & \cellcolor{green!15}\checkmark 
    & \cellcolor{green!15} No -- DiT~\cite{peebles2023scalable} 
    & \cellcolor{green!15} Yes \\

    MeLISA-$\Upsilon$ 
    & AR 
    & $256 \times 256$ 
    & Mix 
    & \cellcolor{green!15}\checkmark 
    & \cellcolor{green!15} No -- UNet~\cite{ho2020denoising} 
    & \cellcolor{green!15} Yes \\

    \bottomrule
  \end{tabular}
\end{table}

\begin{algorithm}[t]
\caption{MeLISA Autoregressive Inference}
\label{alg:inference}
\begin{algorithmic}[1]
\REQUIRE Trained denoiser $D_\theta$, observed context $\bar{x}_{W_{\mathrm{in}}}^{\tau}$, forecast horizon $W_{\mathrm{out}}$, model window length $W$, conditioning length $W_{\mathrm{ctx}}<W$
\STATE Set rollout block size $S \gets W - W_{\mathrm{ctx}}$
\STATE Let $\operatorname{Head}_q(\cdot)$ and $\operatorname{Tail}_q(\cdot)$ denote the first and last $q$ frames of a sequence, respectively
\STATE Initialize context buffer $\bar{c} \gets \operatorname{Tail}_{W_{\mathrm{ctx}}}\!\left(\bar{x}_{W_{\mathrm{in}}}^{\tau}\right)$
\STATE Initialize forecast sequence $\hat{\mathcal{Y}} \gets [\ ]$
\WHILE{$|\hat{\mathcal{Y}}| < W_{\mathrm{out}}$}
    \STATE Form a partially observed rollout window
    \[
    \tilde{\bar{x}}
    \gets
    \mathrm{Concat}\left[\epsilon, \ \mathcal{M}_{\mathrm{fut}}
    \!\left(
        \left[\bar{c};\,\mathbf{0}
        _{S\times F}\right]
    \right), \ \tau_M\right],
    \]
    where $\mathcal{M}_{\mathrm{fut}}$ masks the final $S$ frames and preserves the first $W_{\mathrm{ctx}}$ frames, and $\epsilon$ is Gaussian noise and $\tau_M$ is the masking time indices
    \STATE \(
    \hat{\bar{x}}
    \gets
    D_\theta\!\left(\tilde{\bar{x}},\,0,\,1\right)
    \)
    \STATE \(
    \hat{\bar{x}}_{\mathrm{blk}}
    \gets
    \operatorname{Tail}_{S}\!\left(\hat{\bar{x}}\right)
    \)
    \STATE $q \gets \min\!\left(S,\; W_{\mathrm{out}}-|\hat{\mathcal{Y}}|\right)$
    \STATE \(
    \hat{\mathcal{Y}}
    \gets
    \left[
        \hat{\mathcal{Y}};\,
        \operatorname{Head}_{q}\!\left(\hat{\bar{x}}_{\mathrm{blk}}\right)
    \right]
    \)
    \STATE Update the context buffer with the most recent $W_{\mathrm{ctx}}$ frames:
    \[
    \bar{c}
    \gets
    \operatorname{Tail}_{W_{\mathrm{ctx}}}
    \!\left(
        \left[
            \bar{c};\,
            \operatorname{Head}_{q}\!\left(\hat{\bar{x}}_{\mathrm{blk}}\right)
        \right]
    \right)
    \]
\ENDWHILE
\RETURN $\hat{\bar{X}}_{W_{\mathrm{out}}}^{\tau+W_{\mathrm{in}}} \gets \hat{\mathcal{Y}}$
\end{algorithmic}
\end{algorithm}
\clearpage

\section{Theoretical Origin of MeLISA}
\subsection{Window Consistency: A Self-supervised Learning Perspective}\label{appendix:winc_mf}
We now discuss the motivation behind the Window-Consistency MeanFlow (WinC-MF) objective in Eq.~\ref{eq:winc_mf}. The main challenge is that MeanFlow is fundamentally a one-step generative method: the model should receive enough temporal context to infer meaningful dynamics, but not so much information that forecasting reduces to a deterministic copy task. Our solution is to condition on \emph{masked trajectories}. In this way, the model can exploit partial temporal structure while uncertainty remains over the missing portions of the window. This design is closely related to masked modeling approaches such as Masked Autoencoders (MAEs)~\cite{he2021maskedautoencodersscalablevision} and Masked Trajectory Models (MTMs)~\cite{wu2023maskedtrajectorymodelsprediction}, which likewise learn predictive representations from partially observed inputs.

Using the notation from the main text, and suppressing the explicit $\mathrm{Concat}$ operation for simplicity, let
\begin{equation}
    \mathcal{F}_{\mathcal{M}}
=
\sigma\!\left(
\bar{z}_t^{\tau:\tau+W},\,
\mathcal{M}_{\tau_M}(\bar{x}^{\tau:\tau+W}),\,
\tau_M,\,
t,\,
r
\right)
\end{equation}
denote the information available to the model: the noisy window, the mask pattern, and the diffusion-time variables. Define the target field as
\begin{equation}
    \hat{X}_t^{\tau:\tau+W}
    =
    \bar{z}_t^{\tau:\tau+W}
    -
    t\,u\!\left(
        \bar{z}_t^{\tau:\tau+W},\,
        r,\,
        t
    \right).
\end{equation}
From a self-supervised learning viewpoint, WinC-MF asks the model to predict this target from partial information. Under squared loss, the optimal predictor is the conditional expectation
\begin{equation}
    \mathbb{E}\!\left[
        \hat{X}_t^{\tau:\tau+W}
        \mid
        \mathcal{F}_{\mathcal{M}}
    \right].
\end{equation}
Indeed, for any $\mathcal{F}_{\mathcal{M}}$-measurable predictor $D$, the prediction error decomposes into an irreducible term and an approximation term:
\begin{equation}
\begin{aligned}
    \mathbb{E}\!\left[
        \left\|
            \hat{X}_t^{\tau:\tau+W} - D
        \right\|^2
    \right]
    &=
    \mathbb{E}\!\left[
        \left\|
            \hat{X}_t^{\tau:\tau+W}
            -
            \mathbb{E}\!\left[
                \hat{X}_t^{\tau:\tau+W}
                \mid
                \mathcal{F}_{\mathcal{M}}
            \right]
        \right\|^2
    \right] \\
    &\quad+
    \mathbb{E}\!\left[
        \left\|
            \mathbb{E}\!\left[
                \hat{X}_t^{\tau:\tau+W}
                \mid
                \mathcal{F}_{\mathcal{M}}
            \right]
            -
            D
        \right\|^2
    \right].
\end{aligned}
\end{equation}
The first term does not depend on $D$, so the loss is minimized uniquely by the conditional expectation. Although MeLISA is not trained with a direct regression loss in $X$-space, the same conclusion applies to WinC-MF because its objective differs only by linear transformations of the underlying target.

This perspective makes the role of masking especially clear. If the full clean trajectory were revealed, then the target would become measurable with respect to the conditioning variables, so the conditional variance would collapse to zero. In that regime, there would be no irreducible uncertainty left in the task: the model could simply recover the target directly from the observed trajectory, and forecasting would lose its probabilistic character. By contrast, under nontrivial masking, some future components remain hidden and are not determined by $\mathcal{F}_{\mathcal{M}}$. Whenever the target depends genuinely on those hidden components, it is no longer fully measurable given the conditioning information, and the corresponding conditional variance remains strictly positive on a set of nonzero probability. In other words, masking prevents the trivial \emph{copy-everything} solution by preserving uncertainty over the unobserved part of the window.

Therefore, WinC-MF can be viewed as a form of self-supervised conditional inference: the model learns to reconstruct a target defined over a temporal window from a partially observed and noisy version of that same window. The observed entries provide enough structure to encode local temporal regularities, while the masked entries force the model to infer plausible hidden dynamics rather than memorize the full trajectory. This is precisely what allows WinC-MF to enrich MeanFlow with temporal context without collapsing the forecasting problem into deterministic reconstruction.

\subsection{Time Increment Consistency Loss}\label{appendix:tic}

We now discuss the role of the Time Increment Consistency (TIC) loss in MeLISA. The central question is why a one-step pixel-space autoregressive generative model should remain physically consistent over long rollouts, especially in turbulent regimes. Our answer is that MeLISA does not rely on state reconstruction alone. In addition to the MeanFlow objective, it imposes an explicit finite-lag temporal-increment constraint, which regularizes how the resolved observable evolves across time. This provides a concrete mechanism for the benchmark classes considered in this work. In closed systems such as KF256, TIC approximates consistency with the integrated PDE tendency of the vorticity field. In projected systems such as TCF192, which is obtained from an LBM-LES turbulent channel-flow simulation, TIC constrains the finite-lag evolution of the reduced observable in exactly the sense suggested by projected-dynamics formalisms such as Mori-Zwanzig~\cite{zwanzig1961memory}. More broadly, TIC acts on lagged fluctuation structure rather than only on pointwise reconstruction, which is the level at which many physically relevant turbulent statistics are defined.

\paragraph{Setup and notation.}
Let $H \cong \mathbb{R}^d$ denote the finite-dimensional resolved state space after spatial discretization of the observable of interest. At discrete physical time $\tau \in \mathbb{Z}^+$, let
\begin{equation}
    x^\tau \in H
\end{equation}
denote the ground-truth resolved state, and let $\hat{x}^\tau$ denote the corresponding model reconstruction or prediction. For a rollout window of length $W$, define the lag-$w$ temporal increment by
\begin{equation}
    \Delta x^{\tau,\tau+w}
    :=
    x^{\tau+w} - x^\tau,
    \qquad
    w = 1,\dots,W-1,
\end{equation}
and similarly
\begin{equation}
    \Delta \hat{x}^{\tau,\tau+w}
    :=
    \hat{x}^{\tau+w} - \hat{x}^\tau.
\end{equation}
The TIC loss used in MeLISA is
\begin{equation}
    \mathcal{L}_{\mathrm{TIC}}
    =
    \sum_{w=1}^{W-1}
    \kappa_w\,
    \mathbb{E}_{x,\tau,\tau_M,\epsilon}
    \left[
        \left\|
            \Delta x^{\tau,\tau+w}
            -
            \Delta \hat{x}^{\tau,\tau+w}
        \right\|_2^2
    \right],
\end{equation}
with lag weights $\kappa_w \ge 0$. This is the anchored increment form implemented in MeLISA: each future frame is compared to the same anchor time $\tau$, rather than only to its immediate predecessor. The key point is that TIC is not a generic smoothness prior. In turbulence, rapid fluctuation is often physical, so a smoothness penalty by itself would be poorly justified. TIC instead penalizes mismatch in finite-lag evolution, which is much closer to the way temporal structure is characterized in turbulence theory and in statistical descriptions of dynamical systems.

A first useful observation is that finite-lag increments are directly tied to temporal covariance decay. If $\{X^\tau\}_{\tau\in\mathbb{Z}}$ is a second-order stationary $H$-valued process with mean $\mu=\mathbb{E}[X^\tau]$ and covariance operator
\begin{equation}
    \Gamma(w)
    :=
    \mathbb{E}\!\left[
        (X^{\tau+w}-\mu)\otimes(X^\tau-\mu)
    \right],
\end{equation}
then expanding the squared increment gives
\begin{equation}
    \mathbb{E}\!\left[
        \|X^{\tau+w}-X^\tau\|_2^2
    \right]
    =
    2\,\operatorname{tr}\!\left(
        \Gamma(0)-\Gamma(w)
    \right).
\end{equation}
So at second order, finite-lag increments are not auxiliary quantities: they are exactly equivalent to temporal covariance decay. In particular, matching increment statistics necessarily constrains how quickly temporal correlations are lost along the rollout.

The same point admits a spectral interpretation. For any fixed $a\in H$, define the scalar projected process
\begin{equation}
    z^\tau := \langle a, X^\tau \rangle
\end{equation}
with autocovariance
\begin{equation}
    R_z(w)
    :=
    \mathbb{E}\!\left[
        (z^{\tau+w}-\mathbb{E}z^\tau)
        (z^\tau-\mathbb{E}z^\tau)
    \right].
\end{equation}
Then
\begin{equation}
    \mathbb{E}\!\left[
        |z^{\tau+w}-z^\tau|^2
    \right]
    =
    2\bigl(R_z(0)-R_z(w)\bigr).
\end{equation}
Under the Wiener--Khinchin representation
\begin{equation}
    R_z(w)
    =
    \frac{1}{2\pi}
    \int_{-\pi}^{\pi}
    e^{iw\omega}S_z(\omega)\,\mathrm{d}\omega,
\end{equation}
this becomes
\begin{equation}
    \mathbb{E}\!\left[
        |z^{\tau+w}-z^\tau|^2
    \right]
    =
    \frac{1}{2\pi}
    \int_{-\pi}^{\pi}
    2\bigl(1-\cos(w\omega)\bigr)\,
    S_z(\omega)\,\mathrm{d}\omega.
\end{equation}
Hence TIC is also a temporal-spectrum-aware regularizer: it penalizes mismatch in the energy carried by temporal frequencies, with lag-dependent weighting through the filter $2(1-\cos(w\omega))$.

A second observation is that TIC reduces to tendency matching at short lags. If $x:[0,T]\to H$ is a smooth trajectory satisfying
\begin{equation}
    \dot{x}(t)=F(x(t)),
\end{equation}
then the fundamental theorem of calculus gives
\begin{equation}
    x(t+\delta)-x(t)
    =
    \int_t^{t+\delta}
    F(x(s))\,\mathrm{d}s,
\end{equation}
and a first-order Taylor expansion yields
\begin{equation}
    \frac{x(t+\delta)-x(t)}{\delta}
    =
    F(x(t))
    +
    O(\delta),
    \qquad
    \delta\to 0.
\end{equation}
Thus, at sufficiently short lags, TIC acts as a discrete tendency-matching regularizer: it encourages the model not only to predict plausible states, but also to reproduce the local time evolution that connects them.

This also clarifies why a state-only objective is insufficient. Consider a deterministic one-step predictor $f(c^\tau)$ trained from context $c^\tau$ by the squared loss
\begin{equation}
    \mathcal{L}_{\mathrm{MSE}}(f)
    =
    \mathbb{E}
    \left[
        \|f(c^\tau)-x^{\tau+1}\|_2^2
    \right].
\end{equation}
Conditioning on $c^\tau$ and completing the square shows that the minimizer is
\begin{equation}
    f^\star(c^\tau)
    =
    \mathbb{E}\!\left[
        x^{\tau+1}
        \mid
        c^\tau
    \right].
\end{equation}
So a pointwise squared-loss predictor is driven toward a conditional mean. In chaotic, turbulent, or partially observed systems, however, the physically relevant object is often a full conditional law, or at least a correct multi-time organization of the resolved observable. MeLISA addresses the first issue through its Window Consistency MeanFlow formulation, but a state-only loss still does not directly constrain finite-lag structure. TIC supplies that missing constraint by forcing the rollout to evolve with the correct lagged increments.

These general observations become especially transparent on our two benchmarks.

\paragraph{KF256: finite-lag consistency with the vorticity dynamics.}
For KF256, the learned state is the discretized vorticity field. Let $\omega$ denote the continuous vorticity, let $S_t$ be the exact flow map of the underlying Kolmogorov-flow dynamics, and let $\mathcal{O}$ be the observation operator mapping $\omega$ to the discretized field used by the network. Then
\begin{equation}
    x^\tau = \mathcal{O}\omega(\tau\Delta t),
    \qquad
    x^{\tau+w} = \mathcal{O}S_{w\Delta t}\omega(\tau\Delta t),
\end{equation}
so the finite-lag increment is
\begin{equation}
    \Delta x^{\tau,\tau+w}
    =
    \mathcal{O}\!\left(
        S_{w\Delta t}\omega(\tau\Delta t)-\omega(\tau\Delta t)
    \right).
\end{equation}
Equivalently,
\begin{equation}
    \Delta x^{\tau,\tau+w}
    =
    \int_{\tau\Delta t}^{(\tau+w)\Delta t}
    \mathcal{O}\,\partial_t\omega(s)\,\mathrm{d}s.
\end{equation}
If the vorticity equation is written abstractly as
\begin{equation}
    \partial_t\omega = \mathcal{N}(\omega) + g_{\mathrm{K}},
\end{equation}
where $\mathcal{N}$ collects the advective and dissipative terms and $g_{\mathrm{K}}$ denotes the Kolmogorov forcing contribution, then
\begin{equation}
    \Delta x^{\tau,\tau+w}
    =
    \int_{\tau\Delta t}^{(\tau+w)\Delta t}
    \mathcal{O}\bigl(
        \mathcal{N}(\omega(s)) + g_{\mathrm{K}}(s)
    \bigr)\,\mathrm{d}s.
\end{equation}
This gives TIC a direct interpretation on KF256: it penalizes mismatch in the time-integrated vorticity tendency over finite lags. In the short-lag regime,
\begin{equation}
    \Delta x^{\tau,\tau+w}
    =
    w\Delta t\,
    \mathcal{O}\bigl(\partial_t\omega(\tau\Delta t)\bigr)
    +
    O\!\left((w\Delta t)^2\right),
\end{equation}
so TIC becomes a discrete consistency condition on the true PDE evolution. This is the cleanest setting for TIC, because the resolved field is generated by a closed equation and finite-lag increments are simply finite-time displacements under that dynamics.

\paragraph{TCF192: finite-lag consistency for a projected LBM-LES observable.}
The interpretation for TCF192 is different. Here the target is a 2D observable extracted from a turbulent channel-flow simulation generated by LBM-LES~\cite{chen1998lattice, succi2001lattice}, rather than the full state of a closed 2D system. Accordingly, the learned field should be viewed as a projected observable of a higher-dimensional flow, not as a state that necessarily satisfies a closed autonomous equation on its own. This is exactly the setting in which projected-dynamics formalisms such as Mori-Zwanzig~\cite{zwanzig1961memory} become relevant.

Let $\mathcal{X}(t)\in\mathcal{X}$ denote the full channel-flow state, evolving according to
\begin{equation}
    \dot{\mathcal{X}}(t)=\mathcal{R}(\mathcal{X}(t)).
\end{equation}
Let
\begin{equation}
    \Pi:\mathcal{X}\to H
\end{equation}
be the observation operator extracting the 2D field used for learning, and define the reduced observable
\begin{equation}
    x(t):=\Pi \mathcal{X}(t).
\end{equation}
Because $\Pi$ removes unresolved degrees of freedom, $x(t)$ is generally non-Markovian when viewed in isolation. In the Mori-Zwanzig framework, this reduced evolution can be written as the sum of a resolved drift term, a memory term, and an orthogonal or noise term. If $\mathcal{L}$ denotes the Liouville operator and $\mathcal{P}$ is a projection onto resolved observables with $\mathcal{Q}=I-\mathcal{P}$, then a standard identity has the form
\begin{equation}
\begin{aligned}
    \frac{\mathrm{d}}{\mathrm{d}t}\xi(t)
    =
    \mathcal{P}\mathcal{L}\xi(t)
    +
    \mathcal{P}\mathcal{L}
    \int_0^t
    e^{(t-s)\mathcal{Q}\mathcal{L}}
    \mathcal{Q}\mathcal{L}\xi(s)\,\mathrm{d}s
    +
    \mathcal{P}\mathcal{L}
    e^{t\mathcal{Q}\mathcal{L}}
    \mathcal{Q}\phi_0,
\end{aligned}
\end{equation}
where $\xi(t)$ is the resolved observable and $\phi_0$ is the initial observable. Integrating this identity over a finite lag $\delta$ shows that the increment of the reduced observable naturally decomposes into
\begin{equation}
    x(t+\delta)-x(t)
    =
    \text{resolved drift contribution}
    +
    \text{memory contribution}
    +
    \text{unresolved contribution}.
\end{equation}
This is the key point for TCF192: finite-lag increments of the learned slice accumulate not only instantaneous resolved dynamics, but also memory and unresolved forcing inherited from the full turbulent system. A finite-lag increment loss is therefore exactly the right kind of regularizer for this setting.

In other words, TIC is not imposing an artificial closed 2D PDE on the TCF192 field. Instead, it regularizes the part of the projected dynamics that is actually observable and learnable, namely its finite-lag evolution. This viewpoint is physically meaningful beyond formal projection theory. Space-time correlations are classical objects in wall-bounded turbulence, and recent channel-flow studies report well-defined lagged relationships among turbulent kinetic energy, dissipation, and wall-normal structure. From this perspective, preserving finite-lag increments is a principled way to preserve the temporal organization of the reduced observable, even when no closed equation is available for that observable alone.

In summary, TIC regularizes the temporal organization of MeLISA rollouts at a level that state reconstruction alone cannot enforce. On KF256, where the resolved variable is the vorticity field of a closed system, TIC approximates consistency with the integrated governing dynamics. On TCF192, where the target is a projected observable produced by LBM-LES, TIC constrains the finite-lag behavior of the reduced field, including the effects of memory and unresolved forcing. This is precisely why TIC improves long-horizon physical realism: it encourages the model not only to predict plausible states, but also to evolve them with the correct lagged structure.
\clearpage

\section{Ablation Studies}
\subsection{Effect of WinC-MF Mechanism}
\begin{figure*}[h]
    \centering    \includegraphics[width=0.9\linewidth]{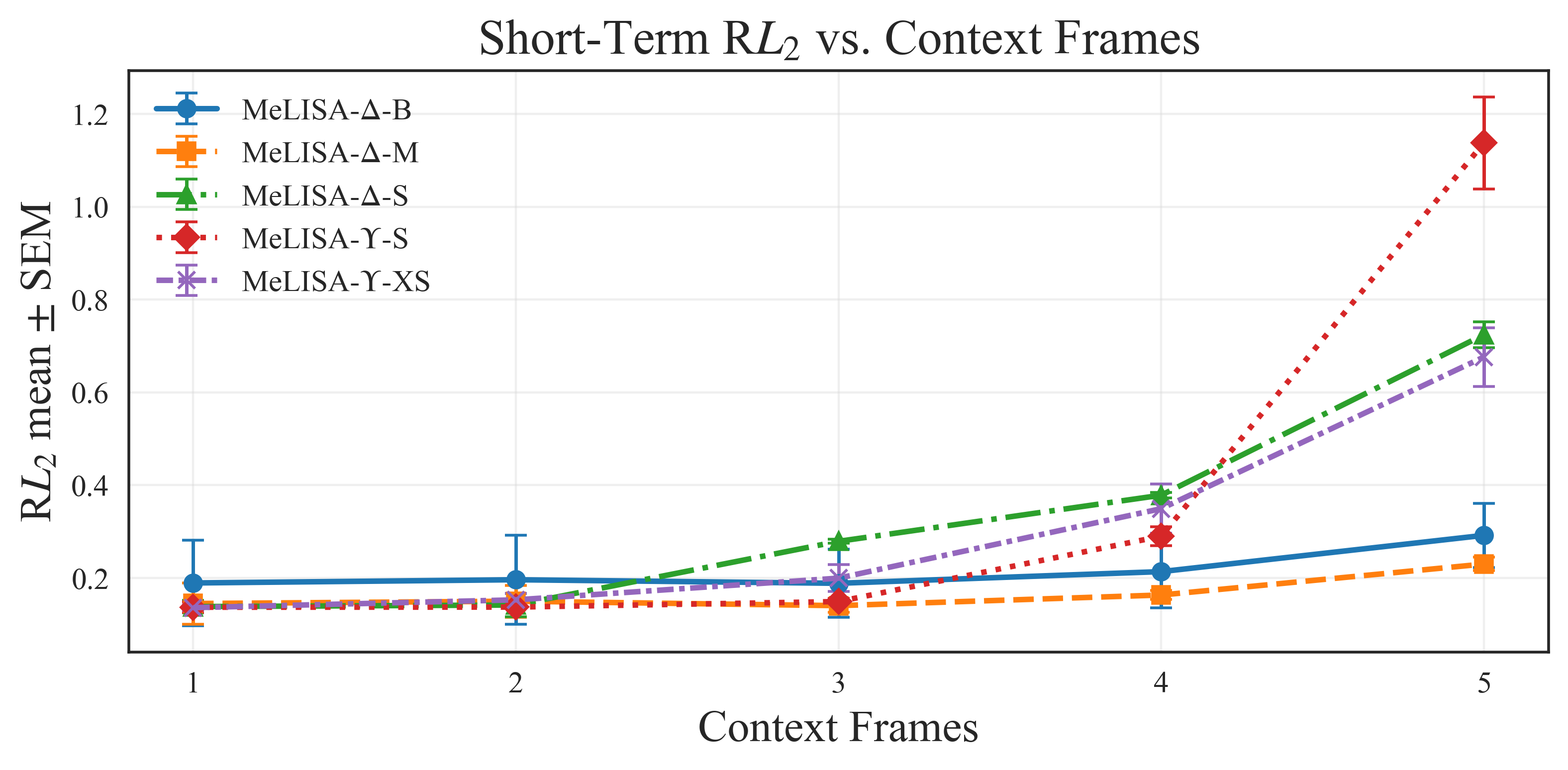}
    \caption{Prediction R$L_2$ loss against number of MeLISA guidance frames. }
    \label{fig:rl2_guidance}
\end{figure*}

In this section, we study how MeLISA depends on the number of guidance frames used during autoregressive rollout. We note that \textbf{WinC-MF cannot be removed or ablated}, as it is the principal mechanism that enables MeLISA to perform spatiotemporal prediction. The distinction between WinC-MF and other possible formulations is discussed in detail in Appendix~\ref{appendix:overview}. Specifically, on the KF256 dataset, we examine the trend of both short-term and long-term metrics as the number of guidance frames varies from 1 to 5. For short-term prediction, we focus on the R$L_2$ metric, shown in Fig.~\ref{fig:rl2_guidance}, computed over the first 40 frames as in the main text. Most models remain stable with 1--3 guidance frames but deteriorate sharply when 4 or 5 frames are provided. In contrast, MeLISA-$\Delta$-M and MeLISA-$\Delta$-B remain stable across the entire range. This behavior is consistent with prior observations that pixel-space self-supervised objectives typically require sufficiently large models to be effective~\cite{he2021maskedautoencodersscalablevision,assran2023self,yan2024emerging}. Under the training masking rate of $0.8$, the expected number of revealed frames is $1 + 5 \times (1 - 0.8) = 2$, including the first frame that is always observed. Interestingly, the largest model, MeLISA-$\Delta$-B, achieves its best performance with 3 guidance frames. This suggests that, with further scaling and larger training datasets, MeLISA may exhibit stronger emergent context utilization, pointing toward the possibility of a generative foundation model for dynamical systems.

\begin{figure*}[t]
    \centering    \includegraphics[width=0.9\linewidth]{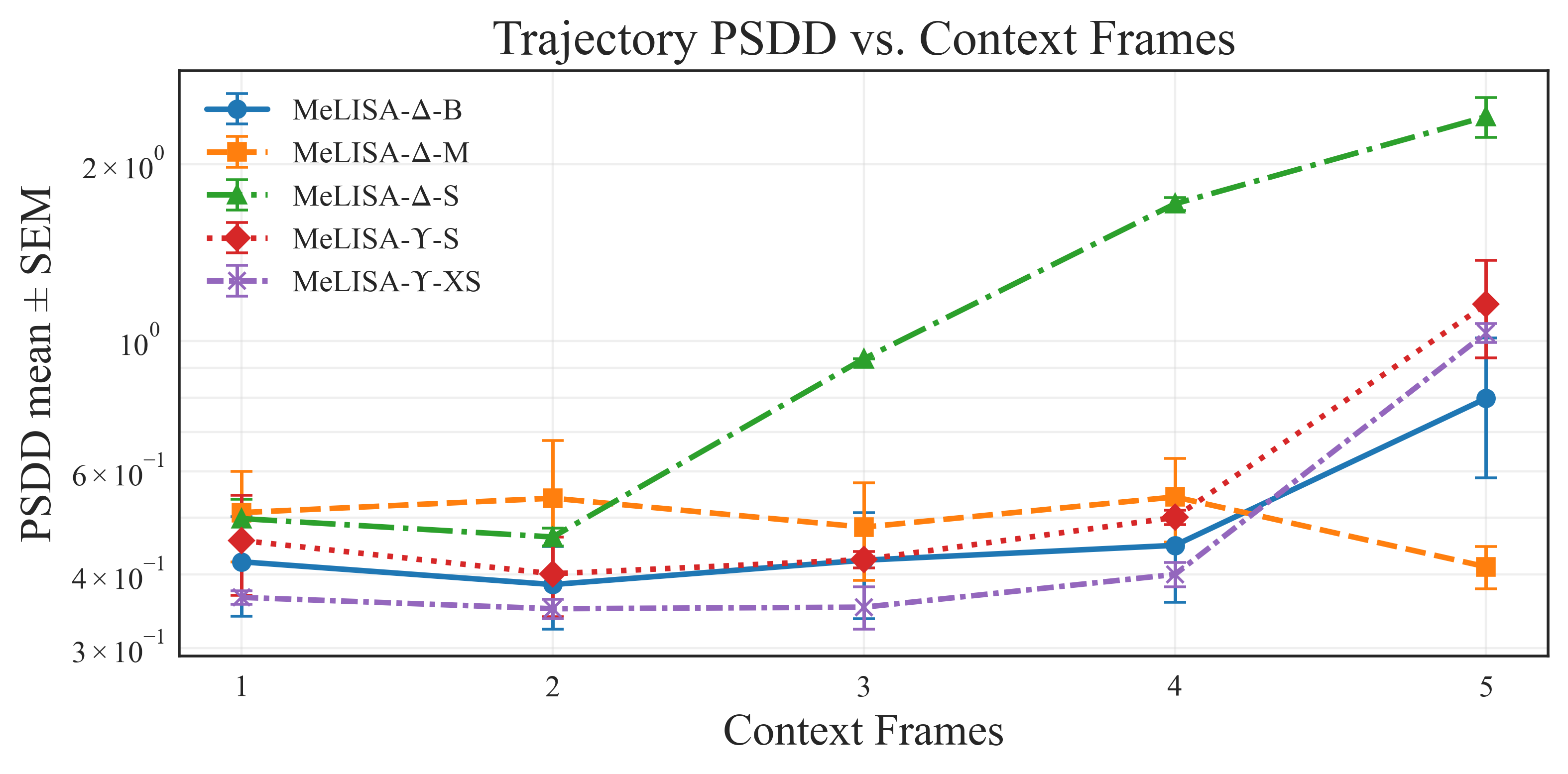}
    \caption{Long-term PSDD loss against number of MeLISA guidance frames. The y-axis is in log scale.}
    \label{fig:psdd_guidance}
\end{figure*}

\begin{figure*}[t]
    \centering    \includegraphics[width=0.9\linewidth]{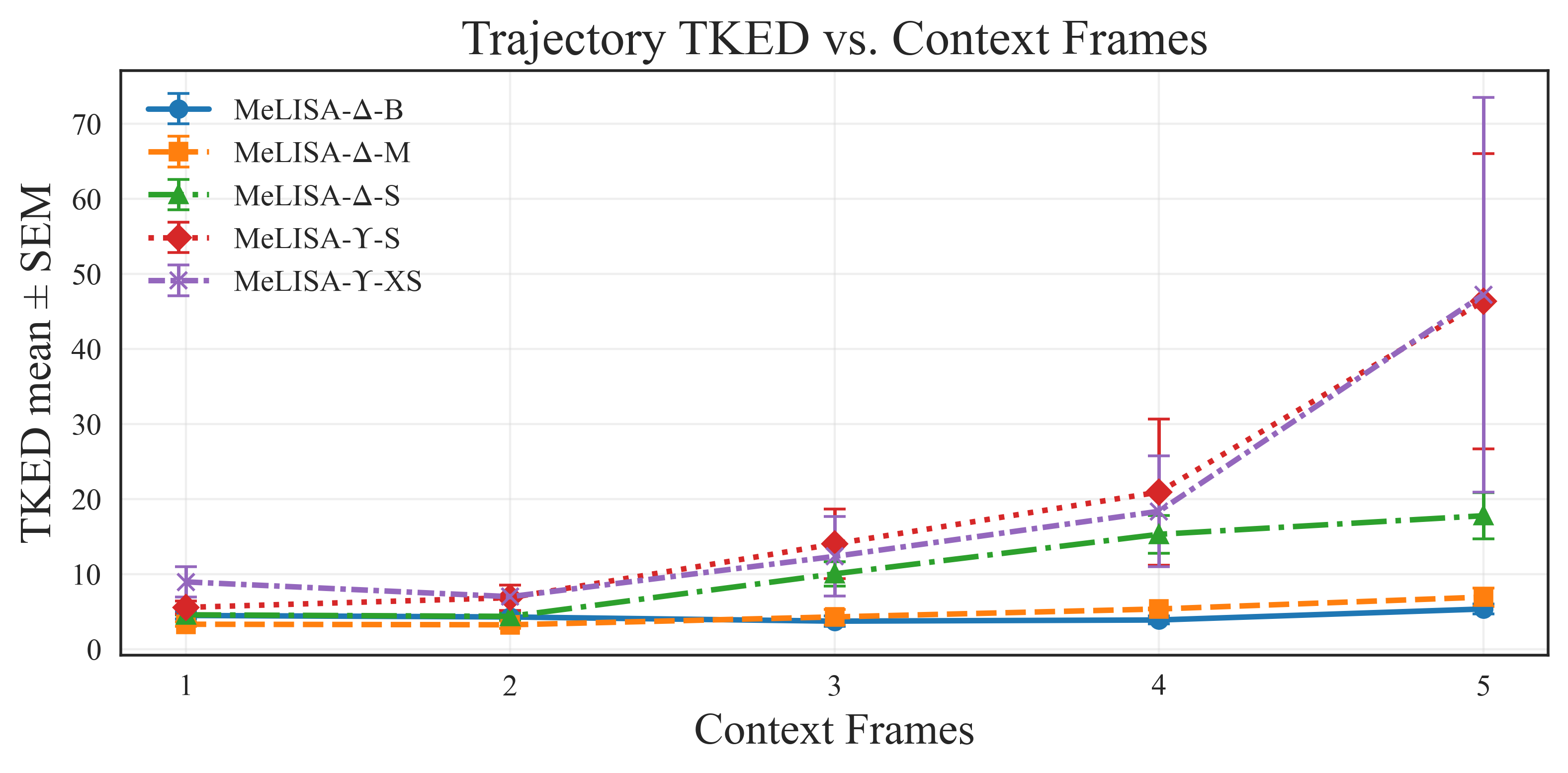}
    \caption{TKED loss against number of MeLISA guidance frames. The y-axis is in log scale.}
    \label{fig:tked_guidance}
\end{figure*}

Another important metric is PSDD. The dependence of full-trajectory PSDD on the number of guidance frames is shown in Fig.~\ref{fig:psdd_guidance}. Interestingly, MeLISA-$\Delta$-M exhibits a consistent improvement in PSDD as more context frames are provided. MeLISA-$\Upsilon$-XS, despite being the smallest model, achieves the best overall spectral accuracy, highlighting the parameter efficiency. To further assess physical consistency, we also examine TKED as a function of the number of guidance frames. The trend in Fig.~\ref{fig:tked_guidance} shows that MeLISA-$\Delta$-M and MeLISA-$\Delta$-B achieve the strongest long-term physical consistency, with TKED values that are smaller when 3 or 5 frames are revealed.

Overall, these results suggest that MeLISA-$\Delta$-M and MeLISA-$\Delta$-B make the most effective use of additional context. This provides encouraging evidence that larger MeLISA models can develop stronger context utilization and more faithfully capture the underlying physical dynamics, pointing toward the potential for MeLISA-based foundation models for dynamical systems.


\subsection{Effect of TIC Loss}

\begin{figure*}[t]
    \centering
    \includegraphics[width=1.\linewidth]{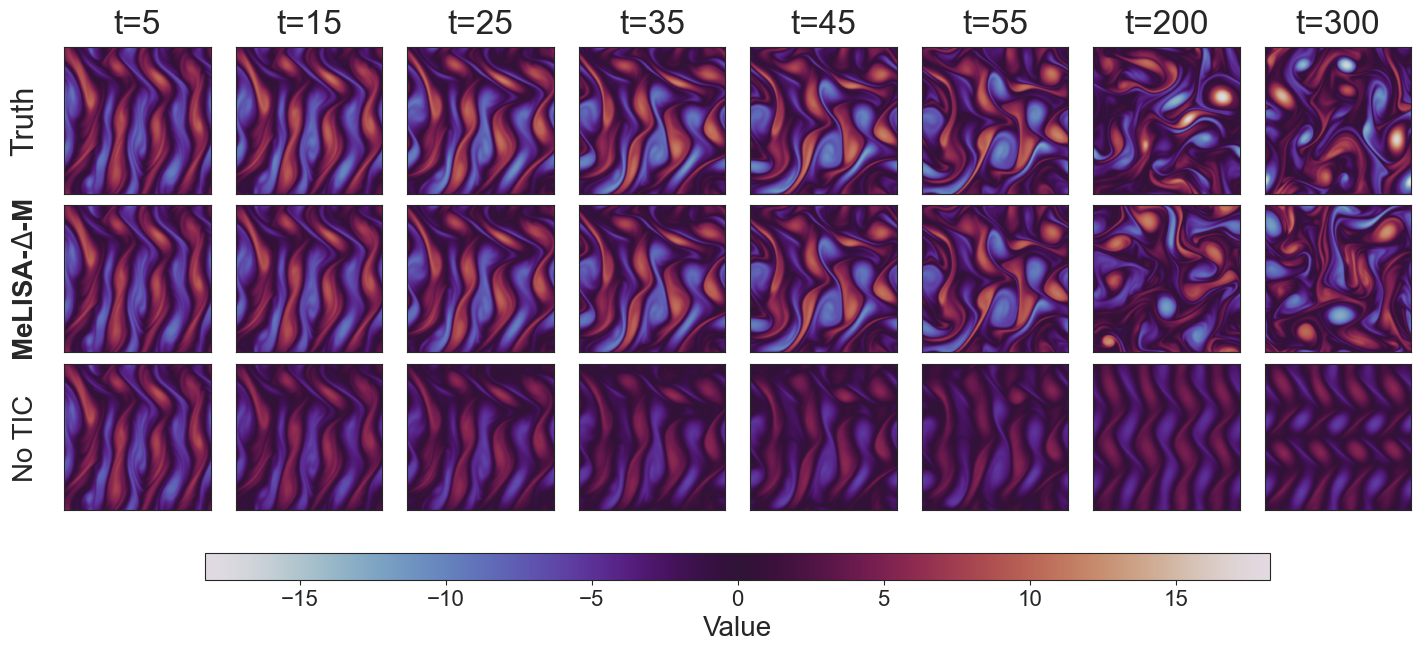}
    \caption{Autoregressive rollout results for MeLISA-$\Delta$-M with two context frames, comparing the full model against a variant trained without TIC under the same optimization setting. $t$ denotes the frame index within the trajectory.}
    \label{fig:tic}
\end{figure*}

To evaluate the role of the TIC loss, we additionally trained MeLISA-$\Delta$-M on KF256 without the TIC term. Autoregressive rollout results for an uncurated initial condition are shown in Fig.~\ref{fig:tic}, and the corresponding statistics are summarized in Table~\ref{tab:tic_table}. The ablation reveals a characteristic failure mode: without TIC, the prediction gradually collapses toward a channel-flow-like mean field. The long-term spectral error (PSDD) also degrades substantially. This behavior is consistent with the theoretical analysis in Appendix~\ref{appendix:tic}: WinC-MF alone is insufficient to recover the correct process, because the model can partially satisfy the objective by regressing toward a \textbf{mean field} solution rather than preserving the correct long-range dynamics.

\begin{table}[!h]
\centering
\caption{KF256 results for MeLISA-$\Delta$-M trained with and without TIC loss. Short-term metrics are computed over the first 40 frames, while long-term metrics are evaluated over the full 320-frame trajectory.}
\setlength{\tabcolsep}{0.5em}
\begin{tabular}{lc|cc|ccc}
\toprule
\multicolumn{2}{c|}{\textbf{Model Specs}}  & \multicolumn{2}{c|}{\textbf{Short-term metrics}} 
  & \multicolumn{3}{c}{\textbf{Long-term metrics}} 
  \\
\midrule
\textbf{Model Name} & \textbf{Model Size} & \textbf{R$L_2$}$\downarrow$ & \textbf{SSIM}$\uparrow$ & \textbf{PSDD}$\downarrow$ & \textbf{TKED}$\downarrow$ & \textbf{MRD}$\downarrow$  \\
\midrule
MeLISA-$\Delta$-M           & 58.3M  & 0.150 & 0.908 & 0.540 & 1.73 & 0.211  \\
MeLISA-$\Delta$-M (No TIC)  & 58.3M  & 0.434 & 0.666 & 1.46  & 13.9 & 0.831  \\
\bottomrule
\end{tabular}
\label{tab:tic_table}
\end{table}

\subsection{Rollout Stability}

\begin{figure*}[t]
    \centering
    \includegraphics[width=0.8\linewidth]{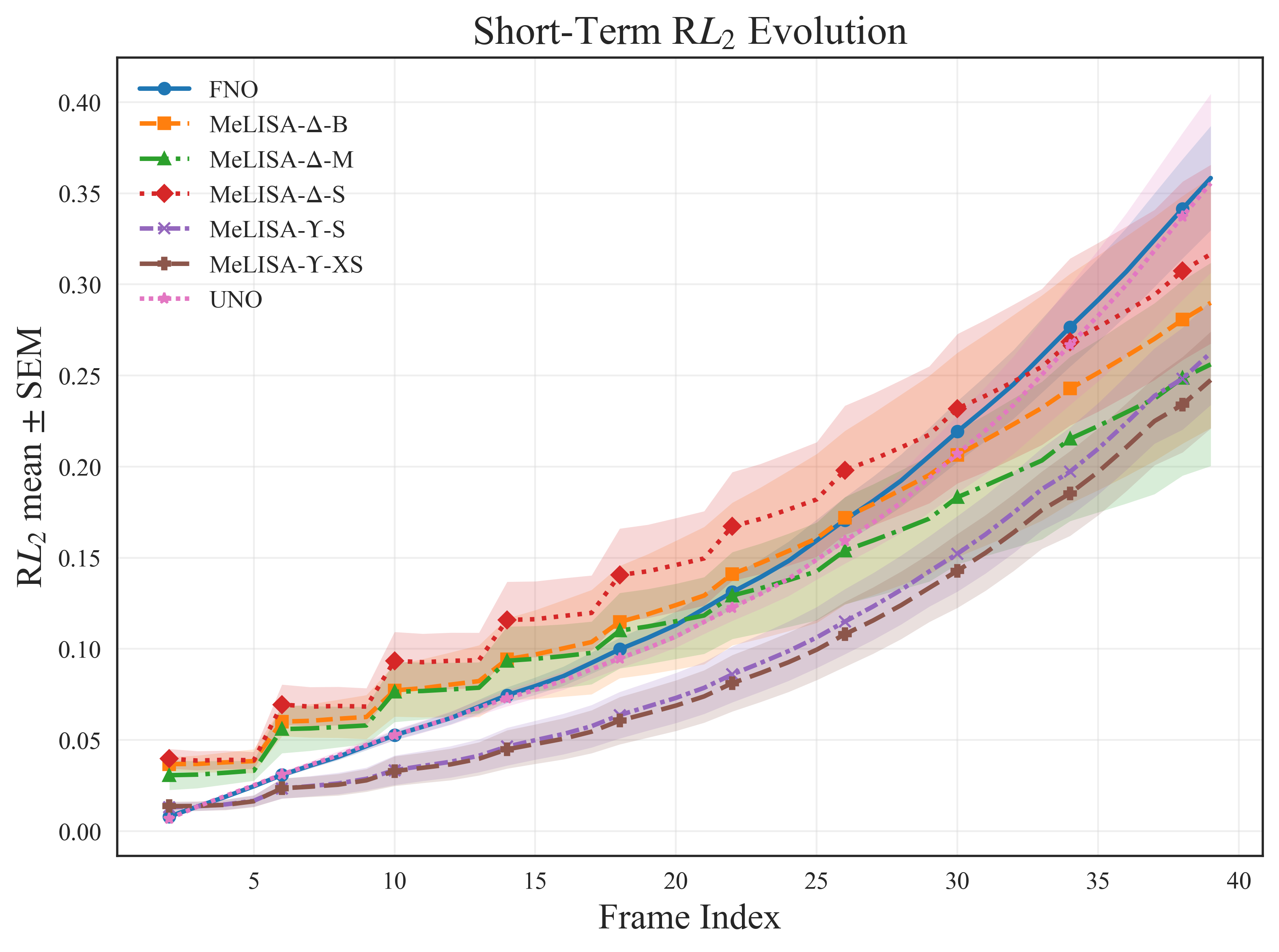}
    \caption{Accumulation of R$L_2$ error over the first 40 rollout frames on KF256.}
    \label{fig:rl2_rollout}
\end{figure*}

We study the temporal rollout stability of MeLISA on the KF256 dataset. Short-range error accumulation is shown in Fig.~\ref{fig:rl2_rollout}. Overall, MeLISA exhibits better long-horizon error stability, whereas the neural-operator baselines achieve lower error only during the first few rollout steps.

\begin{figure*}[t]
    \centering
    \includegraphics[width=1.\linewidth]{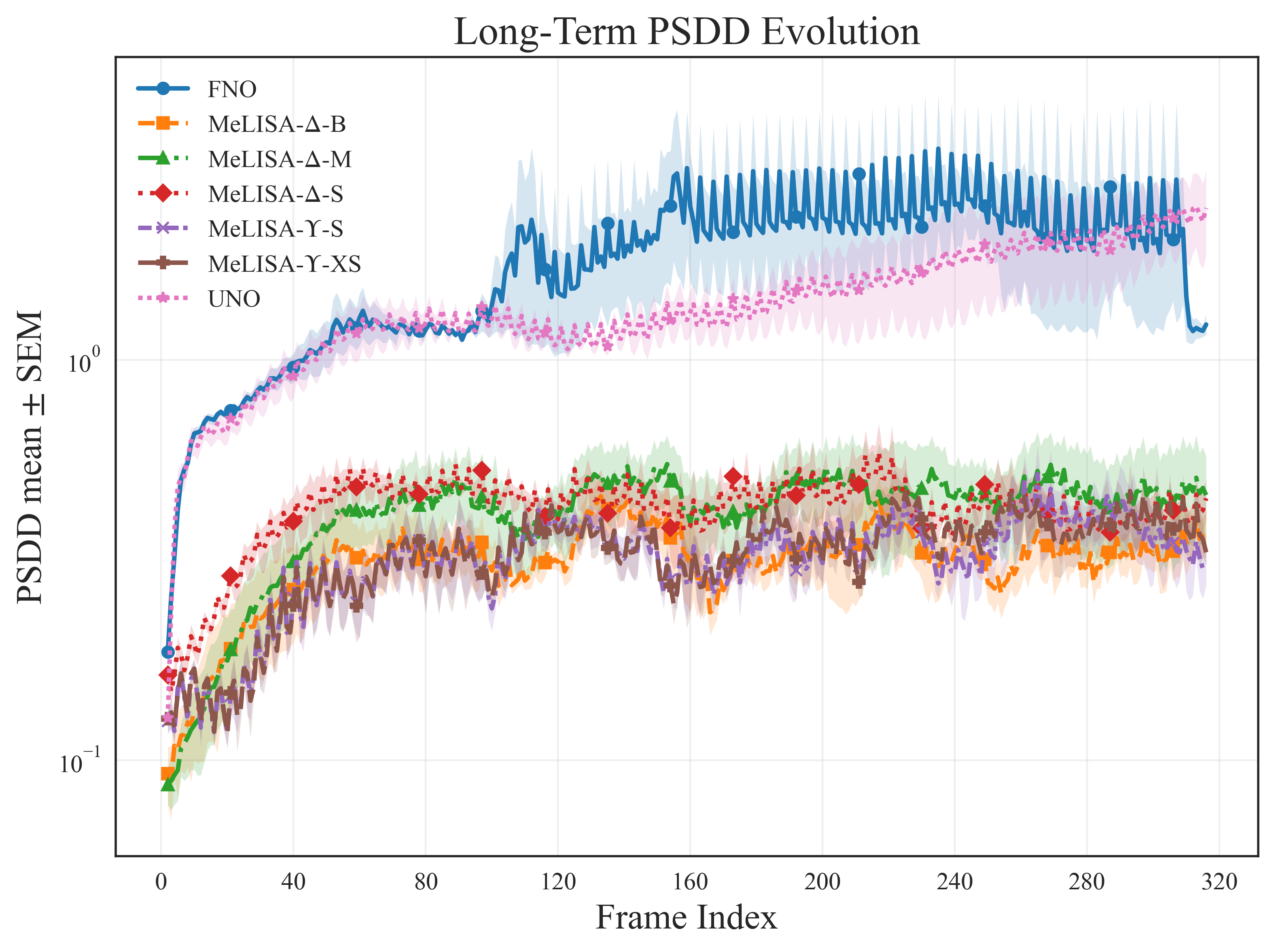}
    \caption{Accumulation of PSDD over autoregressive rollouts on KF256, evaluated over the full 320-frame trajectory.}
    \label{fig:psdd_rollout}
\end{figure*}

\begin{figure*}[t]
    \centering
    \includegraphics[width=1.\linewidth]{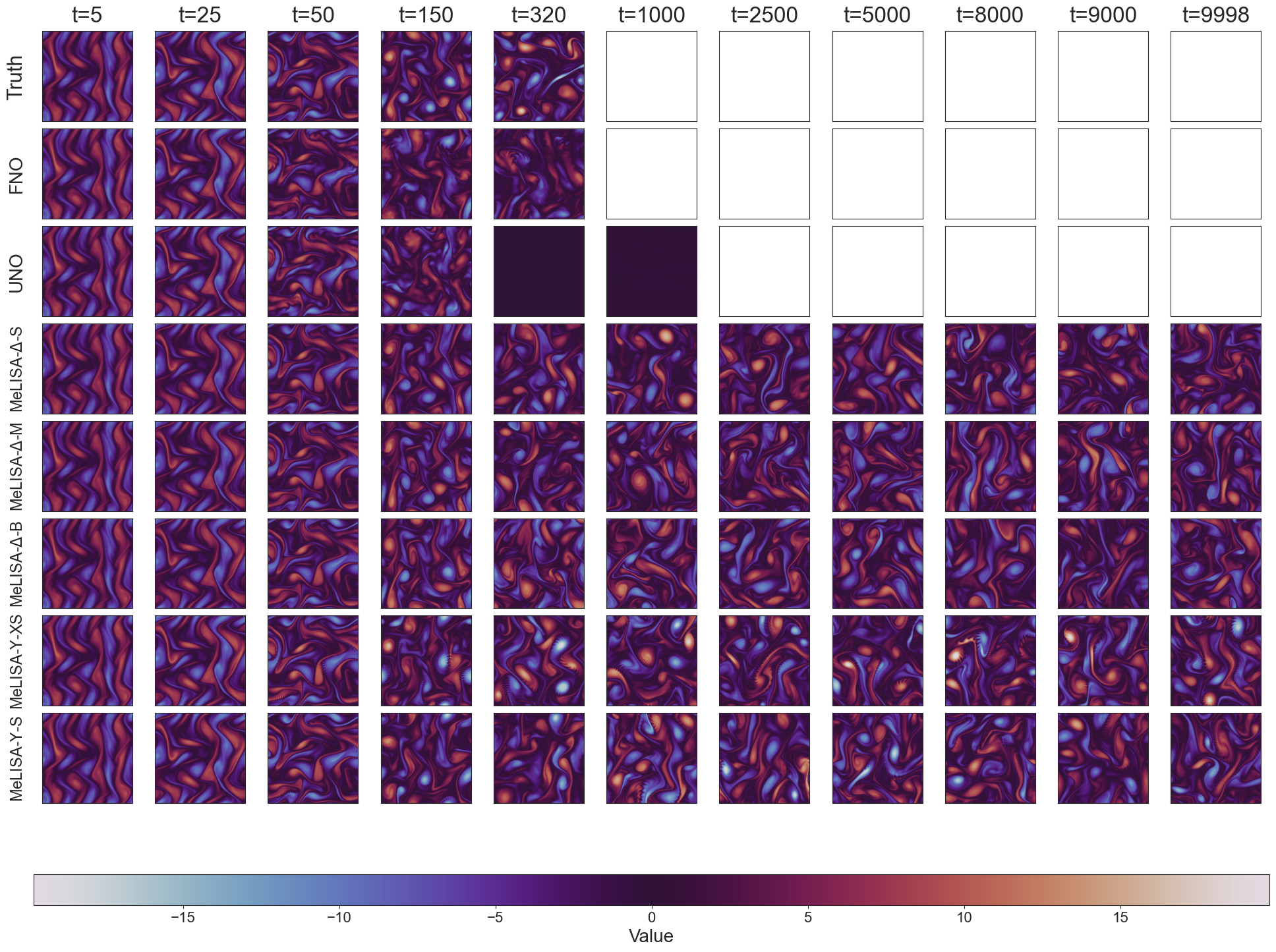}
    \caption{Stress test of maximum rollout length for MeLISA models. White frames indicate either unavailable ground truth or \texttt{NaN} outputs from autoregressive baselines.}
    \label{fig:rollout_kf256}
\end{figure*}

Although the spatial error remains visually reasonable, the spectral error grows much more severely during rollout (Fig.~\ref{fig:psdd_rollout}). Autoregressive neural operators rapidly lose high-frequency content, whereas all MeLISA variants quickly reach a stable spectral-error level. To further assess long-horizon stability under extreme conditions, we roll out all autoregressive models for up to 9998 frames, using two guidance frames for MeLISA. Local-FNO is omitted from this experiment because it cannot stably roll out even to the full in-distribution trajectory length of 320 frames. As shown in Fig.~\ref{fig:rollout_kf256}, all MeLISA variants remain stable up to the final frame and produce plausible results.

\subsection{Probabilistic Forecasting}

\begin{figure*}[h]
    \centering
    \includegraphics[width=1.\linewidth]{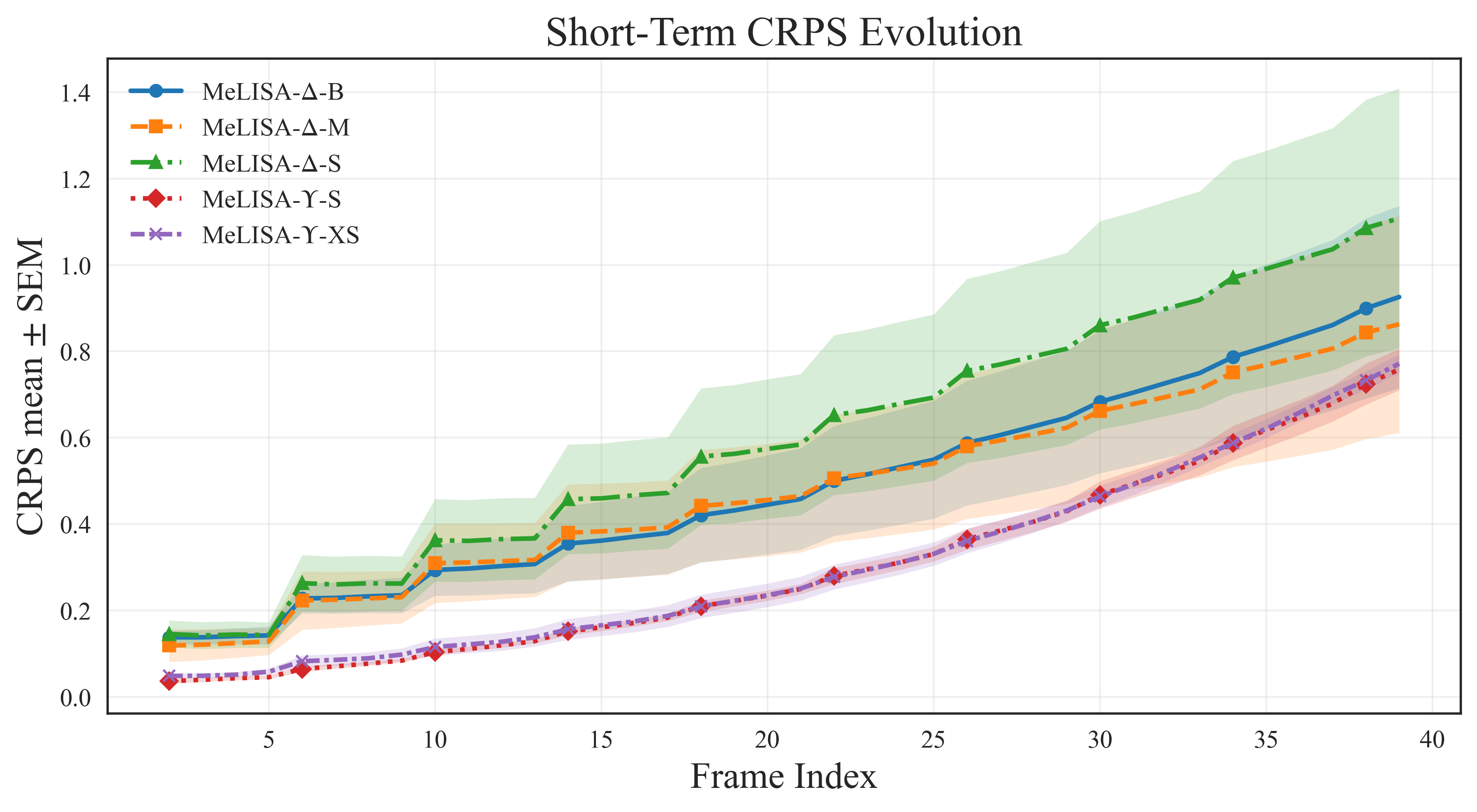}
    \includegraphics[width=1.\linewidth]{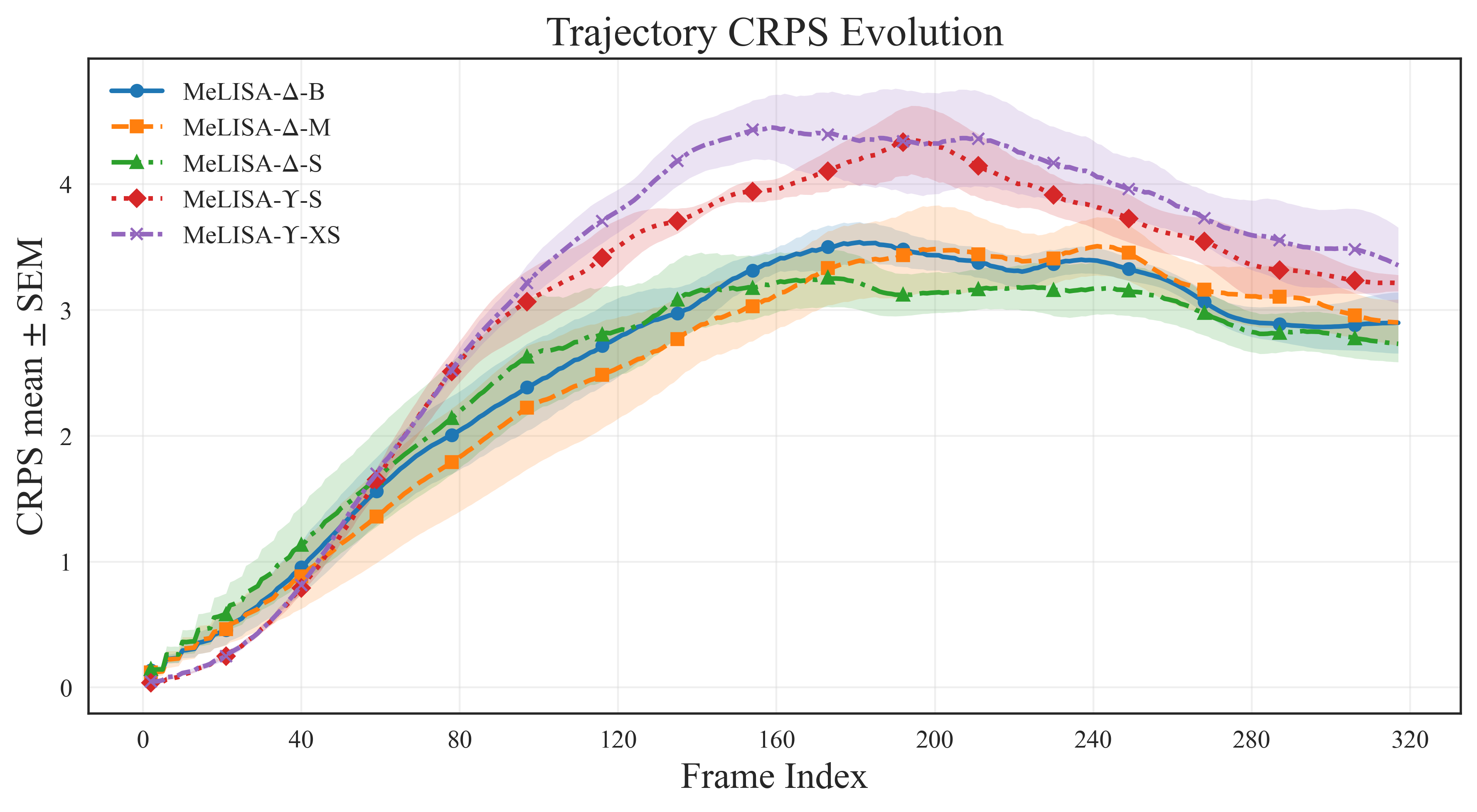}
    \caption{Evolution of CRPS on KF256. The upper panel shows short-term CRPS, and the lower panel shows trajectory-level CRPS over long rollouts.}
    \label{fig:crps_kf256}
\end{figure*}

We emphasize that MeLISA is intrinsically a \textbf{probabilistic forecasting} model. To evaluate this capability, we compute the unscaled continuous ranked probability score (CRPS; see Appendix~\ref{appendix:metrics}) on the KF256 dataset. The absolute CRPS value is not directly comparable across systems with different resolutions or dynamical complexity. For each initial condition, we generate an ensemble of 10 rollout trajectories by varying the random seed, and compute the CRPS with respect to the ground-truth trajectory. The temporal evolution of CRPS is shown in Fig.~\ref{fig:crps_kf256}.

\begin{figure*}[h]
    \centering
    \includegraphics[width=1.\linewidth]{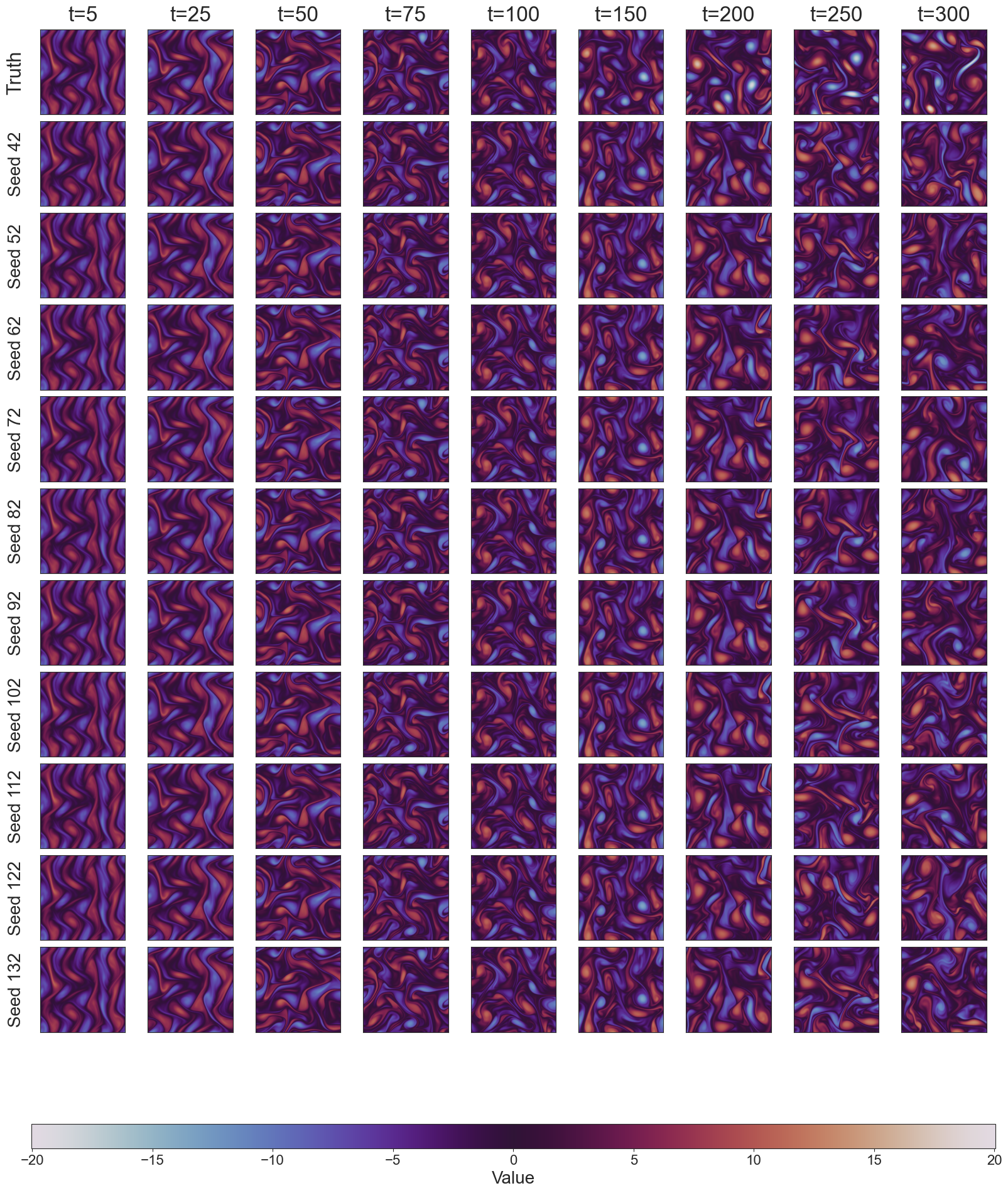}
    \caption{Trajectory ensemble generated by MeLISA-$\Delta$-B for a single initial condition. The random seeds are obtained by adding multiples of 10 to 42.}
    \label{fig:ensem_rollout}
\end{figure*}

In the short-term prediction regime (upper panel of Fig.~\ref{fig:crps_kf256}), the MeLISA-$\Upsilon$ variants achieve lower CRPS. Over longer trajectories (lower panel of Fig.~\ref{fig:crps_kf256}), however, the MeLISA-$\Delta$ variants exhibit greater long-range consistency. For a qualitative illustration, we show the rollout ensemble for a single initial condition in Fig.~\ref{fig:ensem_rollout}.

\subsection{Training Cost \& Inference Costs}\label{appendix:cost_analysis}

We report both training and inference cost in this section. For the neural-operator baselines, all models require roughly 2--3 days of training on a single NVIDIA A100 80GB GPU. All MeLISA models are trained on 32 GH200 nodes (4 GPUs per node) for 4 hours, corresponding to approximately 5 days on a 4 GH200. This training cost is comparable to that reported in prior work~\cite{cachay2025elucidated,molinaro2024generative}. 

For inference, we report \texttt{jax} GFLOP counts per model call and per trajectory in Table~\ref{tab:melisa_flops}. Since neural operators involve Fourier-space operations that are not reliably captured by standard \texttt{PyTorch} profilers, these counts are not directly comparable, and we therefore do not report them here. To assess practical efficiency, we additionally measure wall-clock rollout time per trajectory on a single NVIDIA 4090 GPU in Table~\ref{tab:melisa_time}. Overall, \textbf{MeLISA achieves rollout speeds comparable to those of neural operators}.

\begin{table}[!h]
\centering
\caption{GFLOP counts recorded using \texttt{jax.jit(model).trace(*inputs).lower()} on a single NVIDIA 4090 GPU.}
\setlength{\tabcolsep}{0.3em}
\begin{tabular}{lccccc}
\toprule
\textbf{Model} & MeLISA-$\Upsilon$-XS & MeLISA-$\Upsilon$-S & MeLISA-$\Delta$-S & MeLISA-$\Delta$-M & MeLISA-$\Delta$-B \\
\midrule
TCF192       & 4.20 & 20.5 & 15.9 & - & - \\
KF256        & 11.9 & 63.0 & 64.9 & 234 & 288 \\
\bottomrule
\end{tabular}
\label{tab:melisa_flops}
\end{table}
\clearpage

\begin{table}[!h]
\centering
\caption{Inference time per trajectory in seconds for MeLISA and neural-operator baselines on a single NVIDIA 4090 GPU. Times are reported to two significant figures.}
\setlength{\tabcolsep}{0.3em}
\begin{tabular}{lcccccccc}
\toprule
\textbf{Model} &$\Upsilon$-XS & $\Upsilon$-S & $\Delta$-S & $\Delta$-M & $\Delta$-B & Local-FNO & FNO & UNO \\
\midrule
TCF192 (625 frames) & 12  & 15 & 10  & -   & -    & 19  & 16   & 16   \\
KF256 (320 frames)  & 8.2 & 11 & 6.4 & 9.3 & 10.2 & 9.1 & 7.4 & 8.3 \\
\bottomrule
\end{tabular}
\label{tab:melisa_time}
\end{table}
\clearpage

\section{More Results}

\begin{figure*}[h]
    \centering
    \includegraphics[width=1.\linewidth]{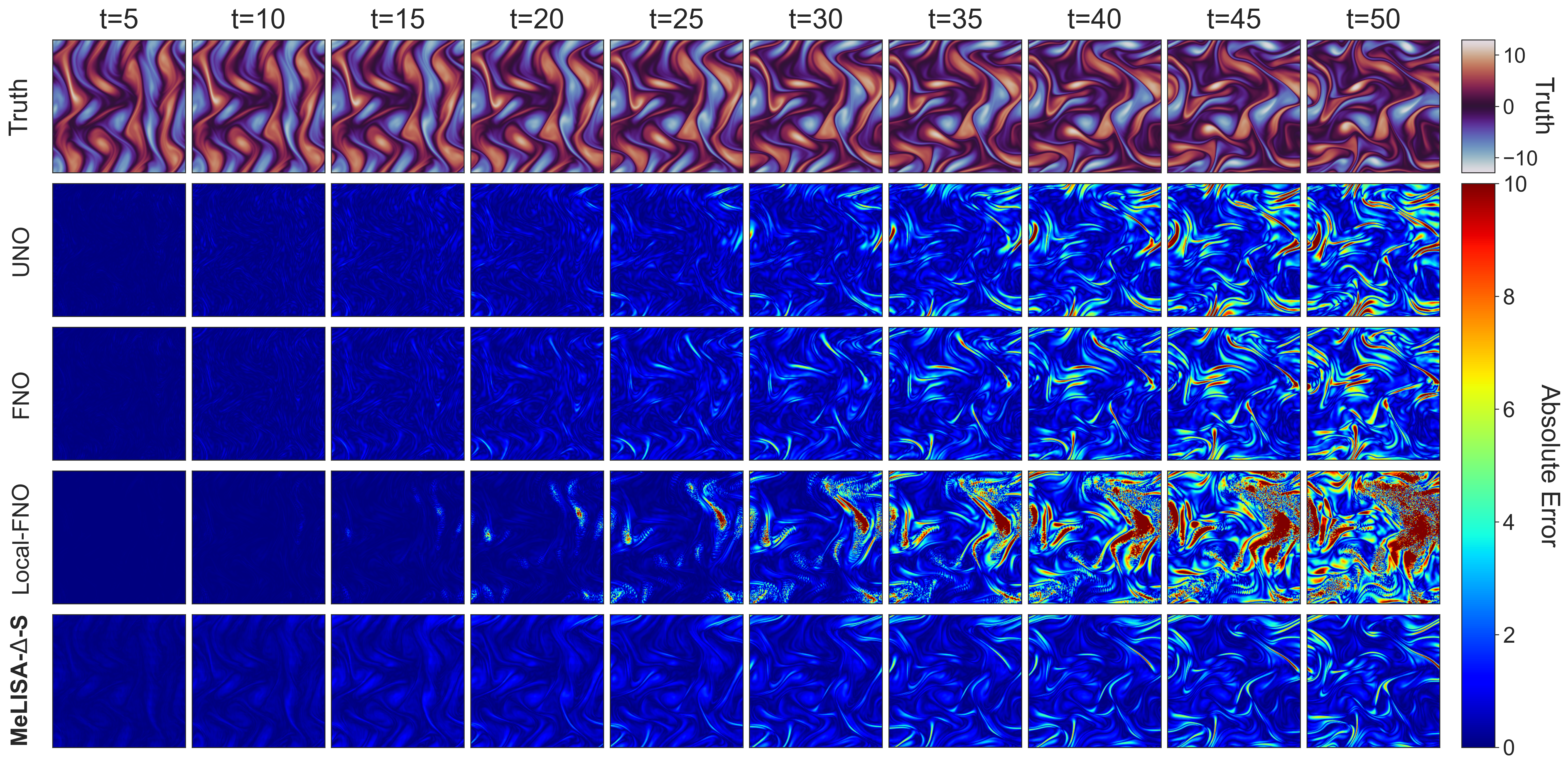}
    \caption{Rollout results on an uncurated test trajectory from \textbf{KF256}, shown as absolute error with respect to the ground truth over the prediction horizon. We compare three autoregressive baselines with one MeLISA variant, MeLISA-$\Delta$-S. Here, $t$ denotes the frame index within the trajectory. MeLISA exhibits substantially improved stability and markedly slower error accumulation.}
    \label{fig:kf256_err_plot}
\end{figure*}

We provide additional qualitative and statistical results for KF256 and TCF192 in this section. Figure~\ref{fig:kf256_err_plot} presents a representative rollout example on KF256. Consistent with the main text, the autoregressive baselines accumulate error rapidly as the rollout proceeds, whereas MeLISA-$\Delta$-S maintains substantially better stability over time. The error maps further show that MeLISA not only reduces the overall error magnitude, but also avoids the localized error amplification that becomes pronounced in the baseline models at later steps.

\begin{figure*}[h]
    \centering
    \includegraphics[width=1.\linewidth]{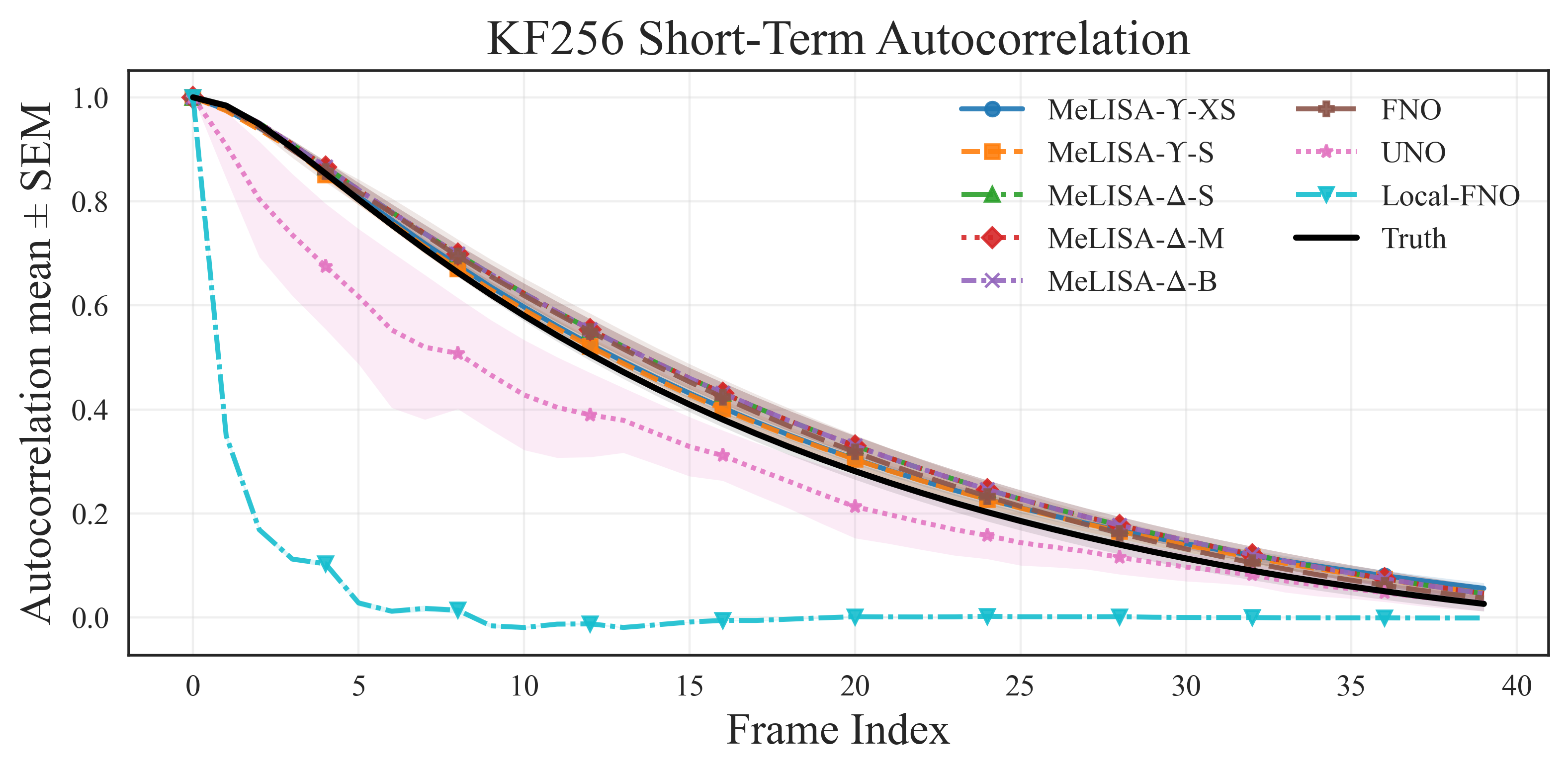}
    \includegraphics[width=1.\linewidth]{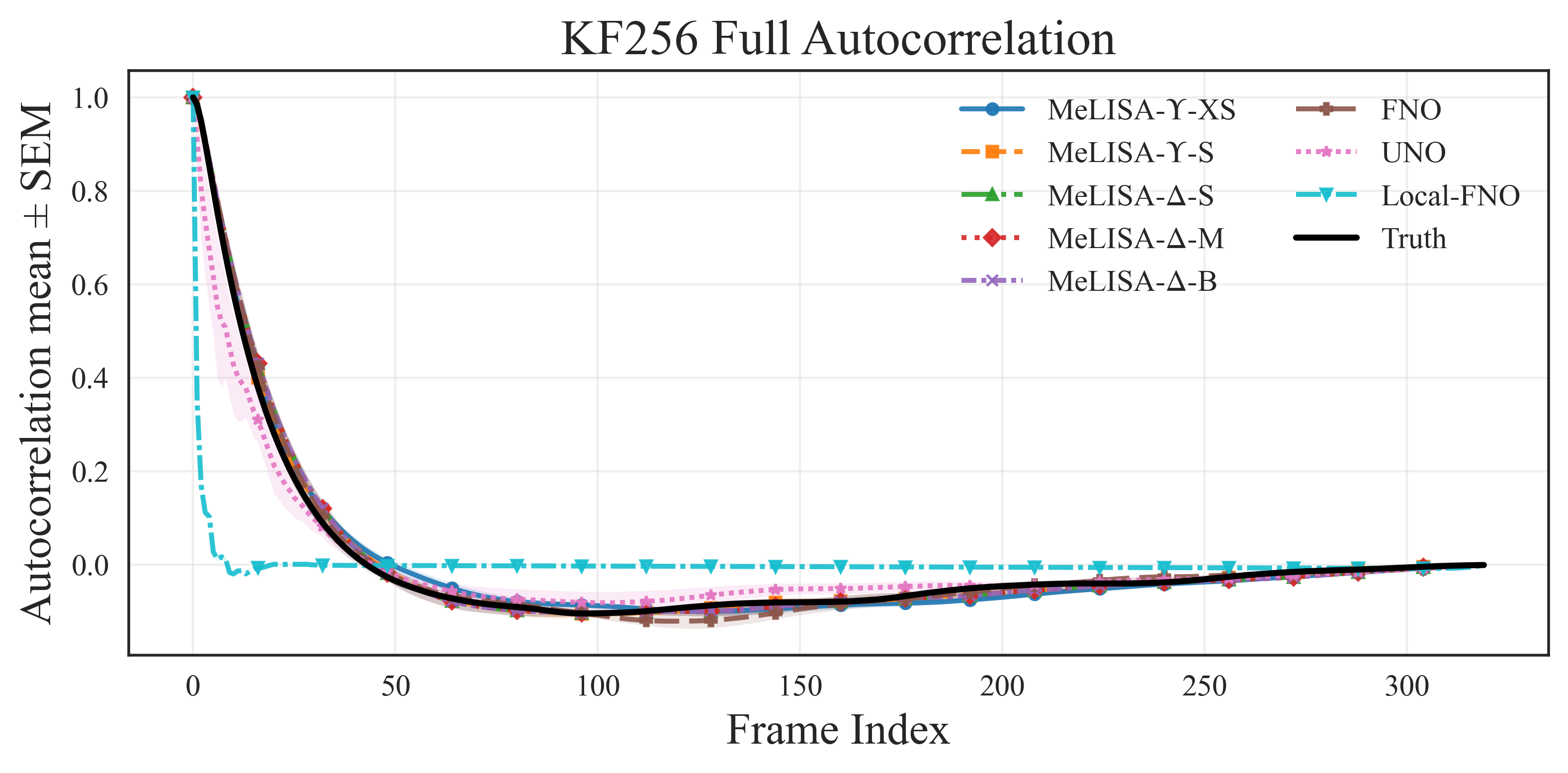}
    \caption{Autocorrelation on the KF256 dataset. The upper panel shows the first 40 frames, while the lower panel shows the full 320 frame trajectory.}
    \label{fig:kf256_autocorr}
\end{figure*}

\begin{figure*}[h]
    \centering
    \includegraphics[width=1.\linewidth]{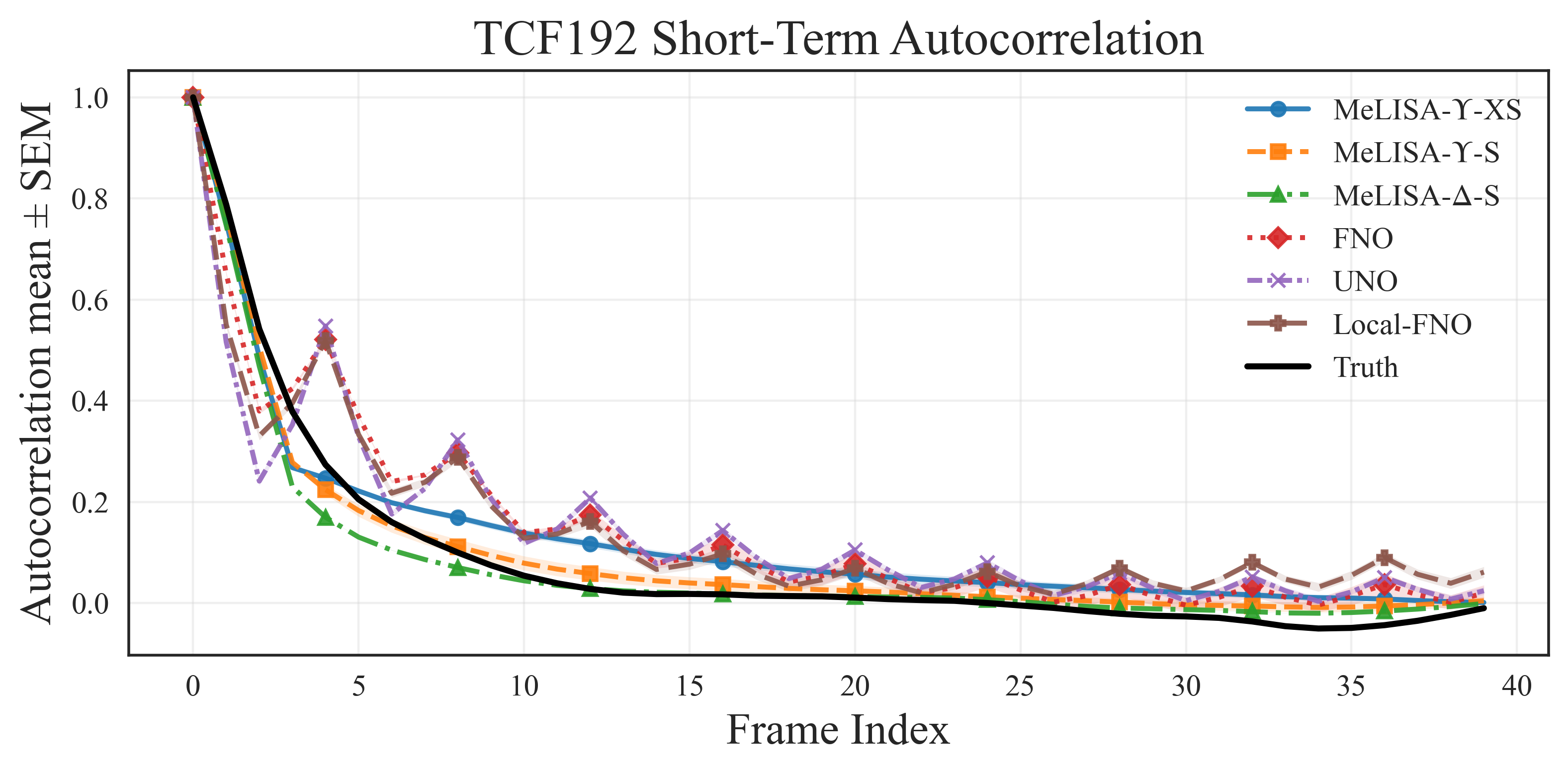}
    \includegraphics[width=1.\linewidth]{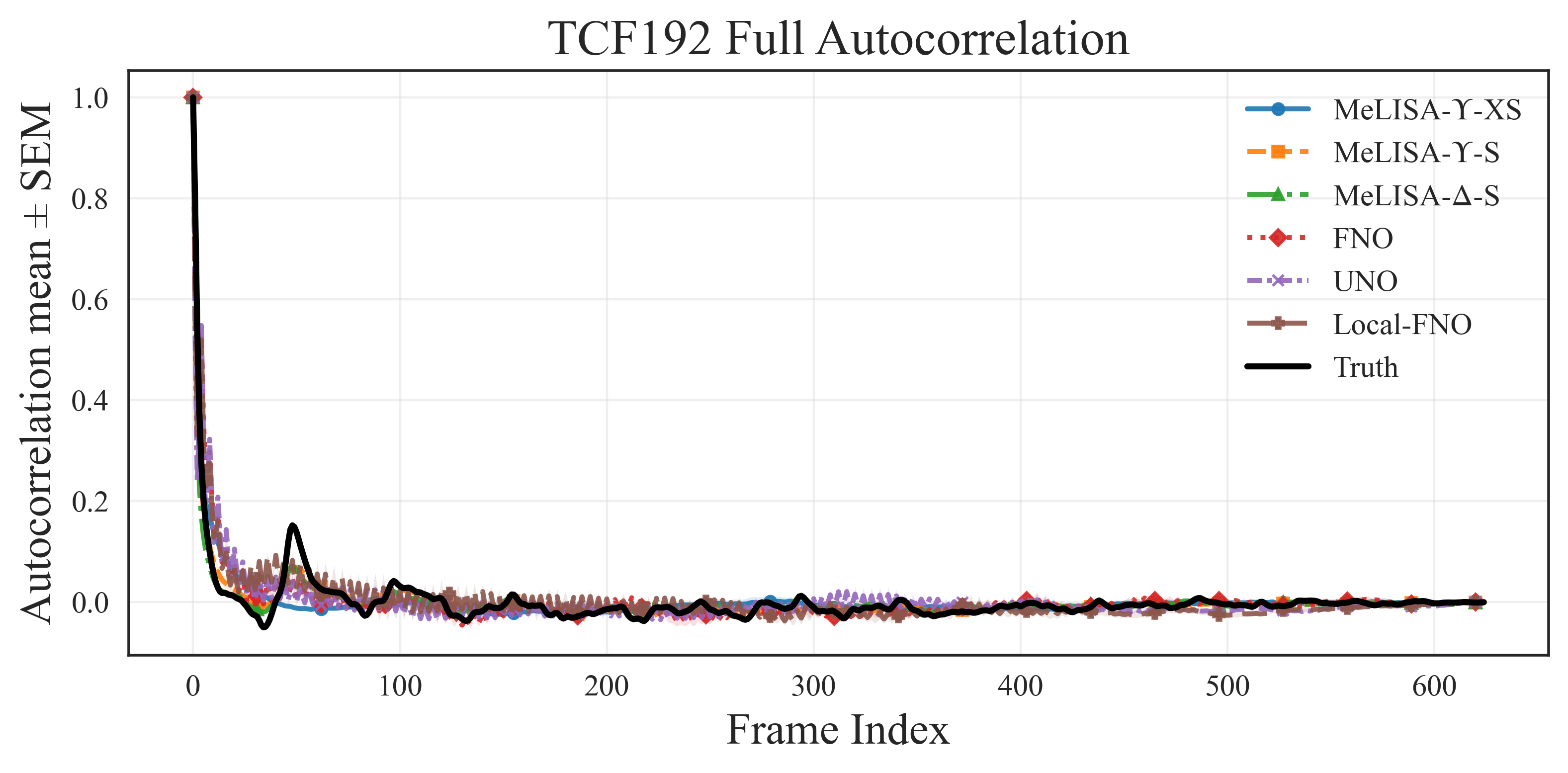}
    \caption{Autocorrelation on the TCF192 dataset. The upper panel shows the first 40 frames, while the lower panel shows the full 625 frame trajectory. Neural operator baselines exhibit pronounced ringing due to the windowed autoregressive rollout setting, whereas the MeLISA curves remain much smoother.}
    \label{fig:tcf192_autocorr}
\end{figure*}

Figures~\ref{fig:kf256_autocorr} and~\ref{fig:tcf192_autocorr} report trajectory autocorrelation statistics for KF256 and TCF192, respectively. These plots provide a complementary view of rollout quality by measuring how well each model preserves temporal dependence over both short and long horizons. On KF256, MeLISA tracks the decay profile of the ground-truth autocorrelation more faithfully across the full trajectory, indicating improved recovery of long-range temporal structure. On TCF192, the difference is even more pronounced: the neural-operator baselines exhibit clear ringing artifacts induced by the windowed autoregressive rollout procedure, while MeLISA produces much smoother curves that better match the physical decay behavior. These results further support the claim that MeLISA preserves long-term temporal statistics more reliably than deterministic autoregressive baselines.

\clearpage

\section{Metrics}\label{appendix:metrics}
We evaluate predicted trajectories against ground-truth trajectories using a combination of short-horizon
field-wise metrics and long-horizon statistical metrics. Let
\[
z,\hat{z} \in \mathbb{R}^{B \times T \times H \times W}
\]
denote the reference and predicted trajectories, respectively.

\subsection{Relative \texorpdfstring{$L_2$}{L2} Error}
\label{sec:metrics_rl2}

To quantify overall prediction error over the short-horizon rollout, we compute a trajectory-level relative
$L^2$ error over the first $T_{\mathrm{eval}}=40$ retained steps. For each trajectory $j$, the predicted and
reference fields are flattened across time and space, and the error is defined as
\begin{equation}
\mathcal{E}_{\mathrm{r}L_2}
=
\frac{1}{B}\sum_{j=1}^{B}
\frac{
\left\lVert \hat{z}^{\,j}_{1:T_{\mathrm{eval}}} - z^{\,j}_{1:T_{\mathrm{eval}}} \right\rVert_2
}{
\left\lVert z^{\,j}_{1:T_{\mathrm{eval}}} \right\rVert_2 + \varepsilon
},
\end{equation}
where $\|\cdot\|_2$ denotes the Euclidean norm after flattening all retained temporal and spatial entries of a
trajectory, and $\varepsilon$ is a small constant for numerical stability.

\subsection{Structural Similarity Index Measure}
\label{sec:metrics_ssim}

To quantify structural agreement between predicted and reference 2D fields, we use the
\textbf{Structural Similarity Index Measure (SSIM)} \cite{wang2004image}. SSIM compares local luminance,
contrast, and structure between a reference field $x \in \mathbb{R}^{H \times W}$ and a prediction
$y \in \mathbb{R}^{H \times W}$.

\paragraph{Gaussian-window local statistics.}
Local moments are estimated using a Gaussian-weighted window of size $w \times w$ with standard deviation
$\sigma$. Denoting Gaussian smoothing by $\mathcal{H}(\cdot)$, the local means are
\begin{equation}
\mu_x = \mathcal{H}(x), \qquad \mu_y = \mathcal{H}(y),
\end{equation}
and the local variances and covariance are
\begin{equation}
\sigma_x^2 = \mathcal{H}(x^2) - \mu_x^2, \qquad
\sigma_y^2 = \mathcal{H}(y^2) - \mu_y^2, \qquad
\sigma_{xy} = \mathcal{H}(xy) - \mu_x \mu_y.
\end{equation}
In our implementation, $\mathcal{H}$ is applied via a separable Gaussian convolution with ``same'' padding,
using a normalized 1D Gaussian kernel along each spatial axis.

\paragraph{SSIM map and aggregation.}
The SSIM map is given by
\begin{equation}
\mathrm{SSIM}(x,y)
=
\frac{(2\mu_x\mu_y + C_1)(2\sigma_{xy} + C_2)}
     {(\mu_x^2 + \mu_y^2 + C_1)(\sigma_x^2 + \sigma_y^2 + C_2)},
\end{equation}
and the scalar SSIM score is the spatial average:
\begin{equation}
\overline{\mathrm{SSIM}}(x,y)
=
\frac{1}{HW}\sum_{i=1}^{H}\sum_{j=1}^{W}\mathrm{SSIM}_{ij}(x,y).
\end{equation}

\paragraph{Stability constants and data range.}
To stabilize the computation, we use
\begin{equation}
C_1 = (K_1 L)^2, \qquad C_2 = (K_2 L)^2,
\end{equation}
where the data range $L$ is computed per sample from the reference field:
\begin{equation}
L = \max(x) - \min(x).
\end{equation}
We use the standard constants $K_1 = 0.01$ and $K_2 = 0.03$. If the reference field is constant
(i.e., $L=0$), the implementation returns $\overline{\mathrm{SSIM}}=1$ when prediction and reference agree
within numerical tolerance, and $0$ otherwise.

\subsection{Mixing Rate Difference}
\label{sec:metrics_mixing}

To assess whether predicted trajectories reproduce the temporal decorrelation behavior of the reference
dynamics, we estimate a \textbf{mixing rate} from the empirical autocovariance and compare it to the
ground-truth value.

\paragraph{Empirical covariance.}
Given a trajectory tensor $x \in \mathbb{R}^{B \times T \times H \times W}$, we first compute the global mean
over all trajectories, times, and spatial locations,
\begin{equation}
\bar{x}
=
\frac{1}{BTHW}
\sum_{b=1}^{B}\sum_{t=1}^{T}\sum_{u=1}^{H}\sum_{v=1}^{W} x_{b,t,u,v},
\end{equation}
and define the centred field $x' = x - \bar{x}$. For a lag $\ell \in \{0,\dots,K\}$, the empirical
covariance is
\begin{equation}
\widehat{C}(\ell)
=
\frac{1}{B\,m_\ell\,H\,W}
\sum_{b=1}^{B}
\sum_{k=1}^{m_\ell}
\sum_{u=1}^{H}\sum_{v=1}^{W}
x'_{b,k,u,v}\,x'_{b,k+\ell,u,v},
\end{equation}
where
\begin{equation}
m_\ell = T - K - \ell.
\end{equation}

\paragraph{Normalized autocorrelation and exponential fit.}
The covariance is normalized by its zero-lag value,
\begin{equation}
\overline{C}(\ell) = \frac{\widehat{C}(\ell)}{\widehat{C}(0)},
\end{equation}
and we estimate the mixing rate $\lambda$ by fitting the exponential decay model
\begin{equation}
\overline{C}(\ell) \approx \exp(-\lambda \ell),
\qquad \ell = 0,\dots,K,
\end{equation}
with the constraint $\lambda \ge 0$.

\paragraph{Reported error.}
Let $\lambda_{\mathrm{gt}}$ be the fitted mixing rate of the ground-truth trajectory and
$\lambda_{\mathrm{pred}}$ that of the prediction. The reported metric is the relative mixing-rate error
\begin{equation}
\mathcal{E}_{\mathrm{mix}}
=
\frac{\left| \lambda_{\mathrm{pred}} - \lambda_{\mathrm{gt}} \right|}
     {\lambda_{\mathrm{gt}}}.
\end{equation}
Lower values indicate better agreement with the temporal mixing behavior of the reference dynamics.

\subsection{Power Spectral Density Discrepancy}

To measure agreement in spatial frequency content, we compute a \textbf{Power Spectral Density
Discrepancy} (PSDD) between predicted and reference fields. This metric compares the radially averaged Fourier
power spectra of the two signals and is particularly useful for assessing whether the model reproduces the
correct distribution of energy across spatial scales.

\paragraph{Frame-wise Fourier representation.}
The PSDD is evaluated on all retained frames after flattening batch and time dimensions, so that each frame is
treated as an independent sample of shape $(H,W,1)$. For each sample $x$, we compute the shifted 2D discrete
Fourier transform
\begin{equation}
\widetilde{F}_x = \mathrm{fftshift}\!\left(\mathcal{F}\{x\}\right),
\end{equation}
and similarly for $y$.

\paragraph{Power spectrum and radial averaging.}
The unnormalized power spectrum is
\begin{equation}
P_x(i,j) = \left| \widetilde{F}_x(i,j) \right|^2.
\end{equation}
For each spatial frequency location $(i,j)$, we define a discrete radial bin index
\begin{equation}
r(i,j)
=
\left\lfloor
\sqrt{\left(i-\left\lfloor H/2 \right\rfloor\right)^2
     +\left(j-\left\lfloor W/2 \right\rfloor\right)^2}
\right\rfloor.
\end{equation}
The radially averaged spectrum is then obtained by averaging $P_x(i,j)$ over all positions belonging to the
same radius bin:
\begin{equation}
R_x(k)
=
\frac{1}{|\Omega_k|}
\sum_{(i,j)\in\Omega_k} P_x(i,j),
\qquad
\Omega_k = \{(i,j): r(i,j)=k\}.
\end{equation}
In our implementation, channels are aggregated before radial averaging.

\paragraph{Cropping and normalization.}
We retain radial bins
\[
k \in \{0,1,\dots,K_r-1\},
\]
where $K_r$ denotes the number of retained radial bins.
The cropped radial spectrum is normalized to sum to one:
\[
\widehat{R}_x(k)
=
\frac{R_x(k)}
{\sum_{k'=0}^{K_r-1} R_x(k') + \varepsilon},
\]
and analogously for $\widehat{R}_y(k)$.

\paragraph{Logarithmic $L_1$ discrepancy.}
The PSDD used in the analysis is the mean absolute difference between the logarithms of the normalized radial
spectra:
\begin{equation}
\mathcal{L}_{\mathrm{PSDD}}(x,y)
=
\frac{1}{K_r}
\sum_{k=0}^{K_r-1}
\left|
\log\!\left(\widehat{R}_x(k)+\varepsilon\right)
-
\log\!\left(\widehat{R}_y(k)+\varepsilon\right)
\right|,
\end{equation}
where $K_r$ (220 for KF256, all bins for TCF192) is the number of retained radial bins. The final score is obtained by averaging this quantity
over all flattened frames.

\subsection{Turbulent Kinetic Energy Difference}
\label{sec:metrics_tked}

To evaluate whether the predicted trajectories reproduce the amplitude of temporal fluctuations at each spatial
location, we compute a \textbf{Turbulent Kinetic Energy Difference} (TKED).

\paragraph{Turbulent fluctuation energy.}
For a trajectory tensor $x \in \mathbb{R}^{B \times T \times H \times W}$, we first compute the temporal mean
for each trajectory and spatial position,
\begin{equation}
\bar{x}_{b,u,v}
=
\frac{1}{T}\sum_{t=1}^{T} x_{b,t,u,v},
\end{equation}
and define the turbulent fluctuation field
\begin{equation}
x'_{b,t,u,v} = x_{b,t,u,v} - \bar{x}_{b,u,v}.
\end{equation}
The corresponding kinetic-energy map is then
\begin{equation}
\mathrm{TKE}_x(u,v)
=
\frac{1}{B}\sum_{b=1}^{B}
\left(
\frac{1}{T}\sum_{t=1}^{T} \left(x'_{b,t,u,v}\right)^2
\right).
\end{equation}
This yields a spatial map describing the average fluctuation energy at each grid location.

\paragraph{Normalized discrepancy.}
Let $\mathrm{TKE}_{\hat{z}}$ and $\mathrm{TKE}_{z}$ denote the predicted and reference temporal kinetic-energy
maps. The reported discrepancy is
\begin{equation}
\mathcal{E}_{\mathrm{TKED}}
=
\frac{
\frac{1}{HW}\sum_{u=1}^{H}\sum_{v=1}^{W}
\left(\mathrm{TKE}_{\hat{z}}(u,v)-\mathrm{TKE}_{z}(u,v)\right)^2
}{
\frac{1}{HW}\sum_{u=1}^{H}\sum_{v=1}^{W} \mathrm{TKE}_{z}(u,v)
}.
\end{equation}
Lower values indicate that the predicted trajectories reproduce the temporal fluctuation energy of the
reference trajectories more faithfully.
\clearpage

\subsection{Continuous Ranked Probability Score}
\label{sec:metrics_crps}

To evaluate probabilistic predictions, we use the \textbf{Continuous Ranked Probability Score} (CRPS), which
measures the agreement between an ensemble forecast distribution and the corresponding observed target.
For a scalar target value $y$ and an ensemble forecast
\[
\{x^{(n)}\}_{n=1}^{M},
\]
the empirical CRPS is defined as
\begin{equation}
\mathrm{CRPS}\!\left(\{x^{(n)}\}_{n=1}^{M}, y\right)
=
\frac{1}{M}\sum_{n=1}^{M} \left|x^{(n)} - y\right|
-
\frac{1}{2M^2}\sum_{n=1}^{M}\sum_{m=1}^{M}\left|x^{(n)} - x^{(m)}\right|.
\end{equation}
The first term measures the mean absolute deviation of the ensemble members from the observation, while the
second term corrects for ensemble spread. Lower values indicate better calibrated and more accurate
probabilistic forecasts.

\paragraph{Field-wise ensemble evaluation.}
For spatial fields, the CRPS is computed independently at each spatial location. Let
\[
\hat{z}_{b,t,u,v}^{(n)}
\]
denote the $n$th ensemble member for batch element $b$ at time step $t$ and spatial location $(u,v)$, and let
\[
z_{b,t,u,v}
\]
be the corresponding ground-truth value. Then the pointwise CRPS is
\begin{equation}
\mathrm{CRPS}_{b,t,u,v}
=
\mathrm{CRPS}\!\left(\{\hat{z}_{b,t,u,v}^{(n)}\}_{n=1}^{M},\; z_{b,t,u,v}\right).
\end{equation}
\clearpage

\section{Dataset Details}\label{appendix:dataset}
\subsection{Turbulent Channel Flow}
\paragraph{Dataset summary.}
The turbulent channel flow dataset used in this work generated via a lattice Boltzmann based in-house numerical solver.

\paragraph{Numerical solver: lattice Boltzmann method.}
Ground-truth trajectories are produced with the large-eddy LBM solver described in Section~S1. LBM is a mesoscopic CFD method that, instead of discretizing the Navier--Stokes equations directly, evolves a discrete set of particle-distribution functions $f_i(\mathbf{x},t)$ on a regular lattice through local collision and streaming steps; macroscopic fields (density, velocity, pressure) are recovered as low-order moments of $f_i$~\cite{latt2008straight}. The collision step is strictly local and the streaming step only moves data to nearest-neighbor lattice sites, which makes LBM massively parallel and numerically well-suited to high-Reynolds-number wall-bounded flows, motivating its use as the reference solver here.

\paragraph{Flow configuration.}
The simulation domain is $L_x \times L_y \times L_z = 768 \times 192 \times 192$ lattice units, with $x$, $y$ and $z$ denoting the streamwise, wall-normal and spanwise directions. The flow is driven by a uniform streamwise body force at a friction Reynolds number $Re_\tau = 180$ (bulk Reynolds number $Re = 3250$), with periodic boundary conditions in $x$ and $z$, no-slip walls at the top and bottom, and a sponge layer near the outlet to suppress spurious reflections. To accelerate transition to a developed turbulent state, a cubic tripping obstacle of size $20 \times 20 \times 100$ is placed at $x = 192$~\cite{xue2022synthetic}; after $10^6$ timesteps the obstacle is removed and the flow is evolved for a further 50 domain-through times before sampling begins on the mid-plane cross-section at $x = 384$.

\paragraph{Computing costs.}
Generating the full dataset required approximately $10^6$ CPU hours on the ARCHER2 supercomputer ($2{,}048$ cores per run, 32 runs).

\subsection{2D Kolmogorov Flow}
\paragraph{Dataset summary.} Based on the governing equations described above, the Kolmogorov flow dataset consists of $70$ independent simulation trajectories of the two-dimensional incompressible Navier-Stokes equations with Kolmogorov forcing at Reynolds number $\mathrm{Re}=1000$. All simulations are performed on a doubly periodic square domain, discretized using a uniform Cartesian grid of size $256 \times 256$. The community code can be found in~\cite{Kochkov2021-ML-CFD}.

Each simulation produces a full spatiotemporal trajectory of the vorticity field consisting of $320$ consecutive temporal frames. From each trajectory, multiple overlapping temporal fragments are extracted to construct supervised learning samples. The validation and test sets are split at trajectory level. 

\paragraph{Governing equation.}
The Kolmogorov flow dataset is generated from numerical simulations of the two-dimensional incompressible Navier-Stokes equations with a spatially periodic body force~\cite{Kochkov2021-ML-CFD}. The governing equations read
\begin{equation}
\begin{aligned}
\frac{\partial \mathbf{u}}{\partial t}
+ \mathbf{u}\cdot\nabla \mathbf{u}
&= -\nabla p
+ \nu \nabla^2 \mathbf{u}
+ \mathbf{f}(\mathbf{x}), \\
\nabla \cdot \mathbf{u} &= 0,
\end{aligned}
\label{eq:kolmogorov_ns}
\end{equation}
where $\mathbf{u}(\mathbf{x},t)=(u,v)$ is the velocity field, $p$ is the pressure, and $\nu$ denotes the kinematic viscosity. The external forcing is given by the Kolmogorov forcing
\begin{equation}
\mathbf{f}(\mathbf{x}) =
\left(
\sin(n y),\; 0
\right),
\end{equation}
with forcing wavenumber $n$, which injects energy at a prescribed spatial scale while preserving periodicity.

The system is evolved on a two-dimensional periodic square domain $\Omega=[0,2\pi]^2$. Under this forcing, the Kolmogorov flow admits laminar solutions at low Reynolds numbers and undergoes a sequence of instabilities and transitions to spatiotemporally chaotic dynamics as the Reynolds number increases, making it a canonical testbed for studies of turbulence, transition, and data-driven modeling
\cite{li2022learning}.

%%%%%%%%%%%%%%%%%%%%%%%%%%%%%%%%%%%%%%%%%%%%%%%%%%%%%%%%%%%%

%\input{checklist}

\end{document}